\documentclass{ieeetj}
\PassOptionsToPackage{table,xcdraw}{xcolor}
\RequirePackage{xcolor}
\usepackage{colortbl}
\usepackage{graphics}           
\usepackage{times}              
\usepackage{amsmath}            
\usepackage{amssymb}            
\usepackage[linesnumbered,ruled,vlined]{algorithm2e}
\usepackage[noend]{algpseudocode}
\usepackage{booktabs}
\usepackage{color}
\usepackage{subcaption}
\usepackage{rotating}
\usepackage{cite}
\usepackage{tcolorbox}
\usepackage{textcomp}
\usepackage{pifont}
\usepackage{array,multirow,graphicx}
\usepackage{hyperref}
\definecolor{instructioncolor}{rgb}{.0,.2,.9}

\def\eqref#1{(\ref{#1})}

\newcommand{\rom}[1]{\uppercase\expandafter{\romannumeral #1\relax}}

\makeatletter
\usepackage{xspace}
\DeclareRobustCommand\onedot{\futurelet\@let@token\@onedot}
\def\@onedot{\ifx\@let@token.\else.\null\fi\xspace}

\def\etal{{\textit{et~al}}\onedot}
\makeatother

\def\etalcite#1{\etal~\cite{#1}}

\usepackage{array}
\newcolumntype{L}[1]{>{\raggedright\let\newline\\\arraybackslash\hspace{0pt}}m{#1}}
\newcolumntype{C}[1]{>{\centering\let\newline\\\arraybackslash\hspace{0pt}}m{#1}}
\newcolumntype{R}[1]{>{\raggedleft\let\newline\\\arraybackslash\hspace{0pt}}m{#1}}

\renewcommand{\b}[1]{\mbox{\boldmath$#1$}}

\newcommand{\rank}[0]{{\mbox{rank}\,}}

\newcommand{\bm}{\b m}

\usepackage{bm}
\usepackage{makecell}
\usepackage{threeparttable}
\usepackage{array}
\usepackage{placeins}
\usepackage{indentfirst}
\graphicspath{{./}}
\hypersetup{hidelinks}

\DeclareCaptionFont{ieeetjblue}{\color{tmlcncolor}}
\DeclareCaptionFont{ieeetjblack}{\color{black}}
\AtBeginDocument{%
	\definecolor{tmlcncolor}{cmyk}{0.93,0.59,0.15,0.02}%
	\definecolor{NavyBlue}{RGB}{0,86,125}%
}

\def\OJlogo{\vspace{-4pt}}
\def\seclogo{\vspace{10pt}}
\def\authorrefmark#1{\ensuremath{^{\textbf{#1}}}}

\definecolor{div}{RGB}{228, 165, 150}
\definecolor{nosupport}{HTML}{D6D6D6}
\definecolor{g11}{RGB}{228, 165, 150}
\definecolor{g10}{RGB}{233, 178, 165}
\definecolor{g9}{RGB}{238, 192, 181}
\definecolor{g8}{RGB}{246, 206, 190}
\definecolor{g7}{RGB}{252, 220, 198}
\definecolor{g6}{RGB}{254, 232, 204}
\definecolor{g5}{RGB}{254, 236, 205}
\definecolor{g4}{RGB}{253, 240, 207}
\definecolor{g3}{RGB}{227, 230, 189}
\definecolor{g2}{RGB}{187, 210, 160}
\definecolor{g1}{RGB}{168, 201, 146}

\newcommand{\dataset}[1]{\scalebox{0.8}[1]{\texttt{#1}}}
\newcommand{\datasettable}[1]{\scalebox{0.85}[1]{\texttt{#1}}}
\SetAlgoSkip{0.5ex}

\begin{document}
\receiveddate{XX Month, XXXX}
\reviseddate{XX Month, XXXX}
\accepteddate{XX Month, XXXX}
\publisheddate{XX Month, XXXX}
\currentdate{XX Month, XXXX}
\doiinfo{XXXX.XXXX.XXXXXXX}

\markboth{GenZ-LIO}{Lee \textit{et al.}}

\title{GenZ-LIO: Generalizable~LiDAR-Inertial~Odometry \\Beyond Confined--Open Boundaries}

\author{Daehan Lee\authorrefmark{1},
	Hyungtae Lim\authorrefmark{2}, \textit{Member, IEEE},
	Seongjun Kim\authorrefmark{1},
	Soonbin Rho\authorrefmark{1},
	Changhyeon Lee\authorrefmark{1},
	Sanghyun Park\authorrefmark{1},
	Junwoo Hong\authorrefmark{1},
	Eunseon Choi\authorrefmark{1},
	Hyunyoung Jo\authorrefmark{1},
	and Soohee Han\authorrefmark{1}, \textit{Senior Member, IEEE}}

\affil{Computational Control Engineering Laboratory (CoCEL), Department of Convergence IT Engineering and Electrical Engineering, \\Pohang University of Science and Technology (POSTECH), Pohang 37673, South Korea}
\affil{Laboratory for Information \& Decision Systems (LIDS), Massachusetts Institute of Technology, Cambridge, MA 02139, USA}
\corresp{Corresponding author: Soohee Han (email: soohee.han@postech.ac.kr).}
\authornote{}

\begin{abstract}
%
%% WHY
% Use 1-2 not too long sentences, which clearly answer the WHY question:
% Why is this relevant, why should I care?
%% WHICH PROBLEM
% One sentence that explain the problem the paper addresses or investigates
% Start with: In this paper, we address the problems of \dots
%% HOW & WHAT
% Around 3 sentences that explain how to approach the problem in general and answers:
% How to solve the problem in general? (1/2 - 1 sentence)
% What makes our approach special? What are we actually doing? What is new?
%% IMPLEMENTATION, EVALUATION, WHAT FOLLOWS

For field robotic missions such as inspection, search-and-rescue, and exploration, light detection and ranging (LiDAR)-inertial odometry (LIO) can serve as a core component of autonomy by providing localization and mapping in GNSS-denied or unstructured environments.
However, transitions between confined and open spaces, which are commonly encountered in field deployments, can induce substantial changes in scan density and local geometric structure, thereby reducing the robustness and computational efficiency of LIO.
To address these issues, we present \textit{GenZ-LIO}, a generalizable LIO framework designed to adapt to variations in spatial scale across confined and open environments.
GenZ-LIO comprises three components: (i) scale-aware adaptive voxelization for regulating scan downsampling across spatial-scale changes, (ii) hybrid-metric state update for combining point-to-plane and point-to-point residuals under varying geometric structure, and (iii)~voxel-pruned correspondence search for efficient point-to-point matching.
We conduct a comprehensive evaluation using 42 sequences from nine public datasets and our newly collected \textit{NarrowWide} dataset to analyze LIO performance under spatial-scale variations across diverse field scenarios.
Across the evaluated sequences, GenZ-LIO maintains stable odometry estimation without divergence, indicating practical robustness under the tested field conditions.
Our code is available~at \textcolor{magenta}{\href{https://github.com/cocel-postech/genz-lio}{\texttt{\textbf{https://github.com/cocel-postech/genz-lio}}}}.
\end{abstract}


\maketitle

%%%%%%%%%%%%%%%%%%%%%%%%%%%%%%%%%%%%%%%%%%%%%%%%%%%%%%%%%%%%%%%%%%%%%%%%%%%%%%%%
\section{INTRODUCTION}
\label{sec:introduction}

\IEEEPARstart{F}{ield} robotic systems are increasingly expected to support inspection, search-and-rescue, and exploration across diverse environments.
In these missions, robots often operate in GNSS-denied or unstructured environments, where reliable localization and mapping are essential for sustained autonomy.
For robots equipped with light detection and ranging (LiDAR), LiDAR-inertial odometry (LIO) addresses this need by fusing geometric measurements from
LiDAR scans with high-frequency inertial measurements from an inertial measurement unit (IMU)~\cite{cadena2016tro, lee2024isr}.

Despite substantial progress, LIO performance can degrade when robots move between confined spaces and open areas commonly encountered in field deployments.
During such transitions, the typical distance from the LiDAR sensor to the observed surfaces changes substantially.
In confined spaces, most returns come from nearby walls and objects, whereas in open areas, they are distributed over longer ranges.
In this paper, we refer to this scene-level extent as \textit{spatial scale}.
Changes in spatial scale can induce substantial variations in scan density and local geometric structure, affecting both voxel-based scan downsampling and the reliability of scan-to-map residuals.

Regarding scan density, changes in spatial scale directly affect how densely LiDAR points are distributed in the observed scene. However, LIO systems typically use a fixed voxel downsampling resolution (i.e., voxel size) pre-tuned for an expected operating scale. Therefore, a mismatch between the current spatial scale and the fixed voxel size can reduce pose estimation stability or computational efficiency~\cite{reinke2022ral, lim2023ur, cheng2025ral}.
For instance, a large voxel size pre-tuned for open spaces becomes excessively coarse when the robot enters confined scenes, oversimplifying geometric structure and degrading estimation accuracy~\cite{lim2023ur, cheng2025ral}.
Conversely, a small voxel size pre-tuned for narrow scenes becomes overly fine in wide areas, resulting in an excessively large number of voxelized points and increased computational load~\cite{reinke2022ral}.

Regarding local geometric structure, the appropriate error metric for scan-to-map alignment depends on whether the environment supports planar modeling~\cite{xu2022tro}.
In confined scenes, LiDAR points tend to be densely distributed and often form locally planar structures, for which the point-to-plane error metric~\cite{rusinkiewicz2001IntConfThreeDDigitalImagingAndModeling} can provide strong constraints when surface normals are reliable~\cite{lee2024ral}.
In contrast, in open areas where points are farther apart and planar structure is weaker, surface normals tend to be less accurate, in which case the point-to-point error metric~\cite{besl1992tpami} can offer comparatively more reliable geometric constraints~\cite{vizzo2023ral, lee2024ral}.
Since these local geometric structures can change as the robot moves, relying on a single residual type may reduce robustness across diverse scenes, as further discussed with experimental evidence in Sec.~\ref{subsec: exp_hybrid_metric}.

\begin{figure}[!t]
	\centering
	\includegraphics[width=\columnwidth]{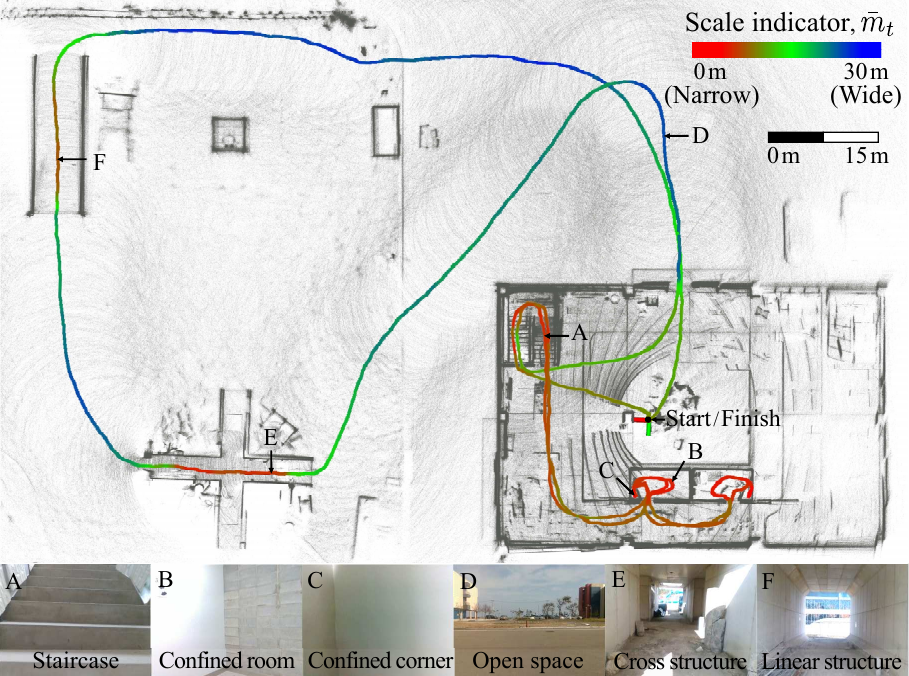}
	\vspace{-0.1cm}
	\caption{Estimated trajectory and mapping result of our \textit{GenZ-LIO} on the \dataset{Handheld-A-01} sequence of our \dataset{NarrowWide} dataset, introduced in this work, where the platform traverses environments with substantially different spatial scales. The trajectory is color-coded by the proposed scale indicator $\bar{m}_t$, which reflects the spatial extent of the surrounding scene. By incorporating this indicator, GenZ-LIO adapts to varying spatial scales, enabling consistent odometry estimation while maintaining computational efficiency across both confined and open areas.}
	\label{fig:Figure1}
	\vspace{-0.7cm}
\end{figure}

For these reasons, we seek a LIO framework that operates robustly and efficiently not only in narrow and wide environments but also during transitions between them.
To this end, as presented in Fig.~\ref{fig:Figure1}, we propose a generalizable LIO framework, called \textit{GenZ-LIO},
designed around two guiding principles: (i) \textit{robustness} to varying spatial scales and shifts in the dominant local geometric structure,
and (ii)~\textit{computational efficiency} in correspondence search.

To achieve robustness, GenZ-LIO incorporates two scale- and geometry-aware mechanisms.
First, it performs scale-aware adaptive voxelization, in which the voxel size is regulated via a feedback controller inspired by the principle of the proportional-integral-derivative (PID) controller, a widely used feedback control mechanism in control theory.
The controller adjusts the voxel size so that the number of voxelized points tracks a target determined from the estimated spatial scale of the current
scan.
Its gains are scheduled based on the estimated spatial scale, the point-count tracking error, and its derivative to support stable and responsive voxel size adjustment during rapid spatial-scale changes.
Second, GenZ-LIO employs a hybrid-metric state update within an error-state iterated Kalman filter (ESIKF), adaptively combining point-to-plane~\cite{rusinkiewicz2001IntConfThreeDDigitalImagingAndModeling} and point-to-point~\cite{besl1992tpami} residuals through reliability-based weighting informed by measurement uncertainty~\cite{yuan2022ral} and voxel discretization error~\cite{wu2024icra}.

To improve computational efficiency, we introduce a voxel-pruned correspondence search strategy to reduce the computational burden of point-to-point matching.
In a voxel hash map, identifying the closest correspondence point typically requires examining up to 27 voxels around the query point~\cite{vizzo2023ral, wu2024icra}, which can incur considerable overhead.
To avoid this brute-force traversal, our method first selects a compact subset of candidate voxels based on the query point’s location within the root voxel.
During the subsequent search, we further prune candidate voxels by comparing the distance from the query point to each voxel with the current closest distance, skipping voxels that cannot yield a closer point.
This two-stage pruning strategy prevents unnecessary voxel access and substantially improves correspondence search efficiency.

%% 6. Contributions
The main contribution of this paper is \textit{GenZ-LIO}, a field-oriented LIO system designed for robust and efficient operation across diverse spatial scales.
The key insight underlying GenZ-LIO is that spatial-scale variation affects not only the number of geometric measurements retained after downsampling but also the reliability of the local geometric constraints used for state estimation. Accordingly, GenZ-LIO jointly adapts the downsampling and state-update stages to address these coupled effects, while the voxel-pruned search enables computationally efficient correspondence matching.

We further conduct a comprehensive evaluation using 42 sequences from nine public datasets and our newly collected \dataset{NarrowWide} dataset to analyze LIO performance under spatial-scale variations across diverse field scenarios.
The \dataset{NarrowWide} dataset was collected using custom hardware platforms in disaster-response and rough-terrain testbeds, including confined spaces, open terrain, and diverse artificial obstacles, to examine repeated confined--open transitions.

%% 7. Key claims
In sum, we make the following five claims:
(i)~our system consistently achieves competitive odometry estimation performance across confined spaces, open areas, and transitions between them;
(ii)~our scale-aware adaptive voxelization enables robust state estimation while efficiently regulating computational resources by accounting for varying spatial scales;
(iii)~our sensitivity-informed gain scheduling improves transient response, resulting in faster convergence and reduced overshoot and oscillations during voxel size control;
(iv)~our hybrid-metric state update improves odometry robustness in scenes where only a few reliable planar regions are observed by using point-to-point residuals as complementary constraints to point-to-plane residuals;
(v)~our voxel-pruned correspondence search substantially reduces computation time by pruning redundant traversal of neighboring voxels.
These claims are backed up by the following sections and our experimental evaluation.
%%%%%%%%%%%%%%%%%%%%%%%%%%%%%%%%%%%%%%%%%%%%%%%%%%%%%%%%%%%%%%%%%%%%%%%%%%%%%%%%
\section{RELATED WORK}
\label{sec:related work}

In this section, we discuss two essential components of LiDAR-based odometry: (\lowercase\expandafter{\romannumeral1}) point cloud downsampling for efficient scan processing and (\lowercase\expandafter{\romannumeral2}) error metrics for residual formulation in scan-to-map alignment.

% 1. Downsampling
Point cloud downsampling is an essential preprocessing step in LiDAR-based odometry to alleviate the computational and memory burden caused by the large volume of LiDAR data, which can reach hundreds of thousands to millions of points per second~\cite{xu2022tro}.
A common choice is uniform voxel grid sampling~\cite{schnabel2006vgtc,rusu2011icra}, which discretizes space into fixed-size cells (i.e., voxels) and retains a small number of points per voxel.
This approach is simple to implement, computationally efficient, and integrates easily into existing front-end modules.
For these reasons, it remains a common choice in state-of-the-art (SOTA) LiDAR(-inertial) odometry systems~\cite{dellenbach2022icra, chen2022ral, vizzo2023ral, lee2024ral, xu2022tro, he2023ais, chen2023icra, chen2024ral}.
As discussed in Sec.~\ref{sec:introduction}, however, its reliance on a fixed voxel size makes it susceptible to substantial variations in spatial scale.

To address the limitation of uniform voxel grid sampling, various point sampling approaches have been explored in LiDAR-based odometry to reduce redundancy or leverage informative measurements during optimization~\cite{zhou2020icra,jiao2021icra,li2021icra,petracek2024ral,tuna2024tro,tuna2025tfr}.
However, such approaches typically incur extra computational cost due to additional processing such as surface normal estimation.
Moreover, they are commonly applied after an initial fixed-resolution voxelization of the raw scan, which may still expose them to the limitations of voxel grid sampling with a fixed voxel size.

\begin{figure}[t]
	\centering
	\begin{subfigure}{1.0\columnwidth}
		\centering
		\includegraphics[width=\linewidth]{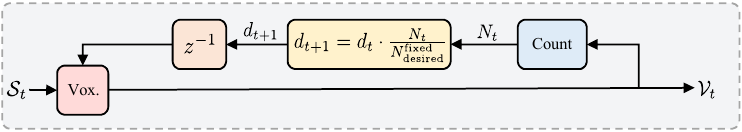}
		\captionsetup{font=scriptsize}
		\caption{LOCUS\,2.0~\cite{reinke2022ral}}
		\label{subfig:locus}
	\end{subfigure}
	
	\vspace{-0.1cm}
	
	\begin{subfigure}{1.0\columnwidth}
		\centering
		\includegraphics[width=\linewidth]{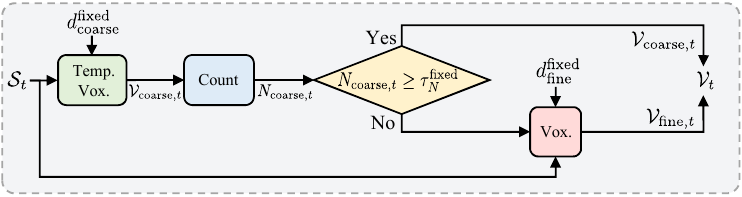}
		\captionsetup{font=scriptsize}
		\caption{AdaLIO~\cite{lim2023ur}}
		\label{subfig:adalio}
	\end{subfigure}
	
	\vspace{-0.1cm}
	
	\begin{subfigure}{1.0\columnwidth}
		\centering
		\includegraphics[width=\linewidth]{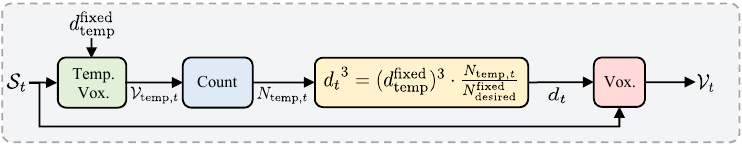}
		\captionsetup{font=scriptsize}
		\caption{LIVOX-CAM~\cite{cheng2025ral}}
		\label{subfig:livox-cam}
	\end{subfigure}
	
	\vspace{-0.1cm}
	
	\begin{subfigure}{1.0\columnwidth}
		\centering
		\includegraphics[width=\linewidth]{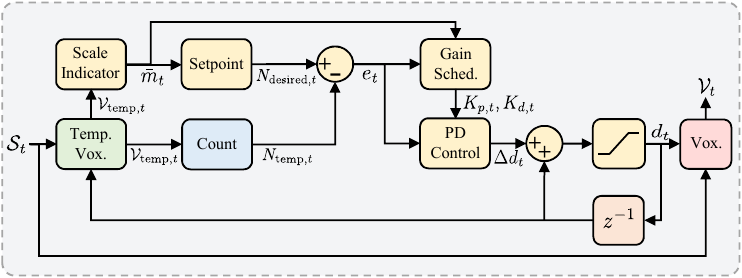}
		\captionsetup{font=scriptsize}
		\caption{Ours}
		\label{subfig:ours}
	\end{subfigure}\par % Apply the following vertical space outside horizontal mode.
	\vspace{-0.2cm}
	\caption{Comparison of adaptive voxelization strategies. (a)~LOCUS\,2.0~\cite{reinke2022ral} updates the voxel size for the next scan, denoted by $d_{t+1}$, using the ratio of the current voxelized point count $N_t$ to the fixed desired point count $N_\mathrm{desired}^\mathrm{fixed}$. (b)~AdaLIO~\cite{lim2023ur} adaptively selects between the predefined coarse voxel size $d_\mathrm{coarse}^\mathrm{fixed}$ and the fine voxel size~$d_\mathrm{fine}^\mathrm{fixed}$ by checking whether the coarse voxelization yields fewer points than a fixed threshold~$\tau_N^\mathrm{fixed}$. (c)~LIVOX-CAM~\cite{cheng2025ral} adjusts the voxel size $d_t$ by first performing a temporary voxelization with a fixed voxel size~$d_\mathrm{temp}^\mathrm{fixed}$ and then updating it based on the ratio between the temporary point count $N_{\mathrm{temp},t}$ and the desired point count $N_\mathrm{desired}^\mathrm{fixed}$. This update follows a volume-based scaling strategy rather than the linear scaling used in LOCUS\,2.0~\cite{reinke2022ral}. (d)~Our method adaptively computes the desired point count~$N_{\mathrm{desired},t}$ based on the scale indicator $\bar m_t$ and determines the corresponding voxel size $d_t$ via proportional-derivative (PD) control with sensitivity-informed gain scheduling. This process corresponds to Fig.~\ref{fig:genz-lio_flowchart}(b).}
	\label{fig:adap_vox_diagram_comparison}
	\vspace{-0.7cm}
\end{figure}

In light of these considerations, adaptive voxel grid sampling schemes have been proposed to handle variations in spatial scale while preserving real-time efficiency.
Reinke~\etalcite{reinke2022ral} introduced an adaptive scheme in LOCUS\,2.0 that updates the voxel size to maintain a constant desired number of voxelized points $N_\mathrm{desired}^\mathrm{fixed}$, using a linear scaling strategy, as illustrated in Fig.~\ref{fig:adap_vox_diagram_comparison}(a).
While this approach effectively regulates the voxelized point count to a user-defined level, it leaves room for further exploration on how the desired number of voxelized points should adapt in response to continuous variations in spatial scale.
Lim~\etalcite{lim2023ur} proposed AdaLIO, which switches between two predefined voxel sizes, as shown in Fig.~\ref{fig:adap_vox_diagram_comparison}(b).
It uses a coarse voxel size $d_\mathrm{coarse}^\mathrm{fixed}$ in general operation and switches to a fine voxel size $d_\mathrm{fine}^\mathrm{fixed}$ when a confined space is detected based on the number of voxelized points and the spatial distribution of occupied voxels.
This approach can be effective when entering confined scenes, but its use of only two discrete, user-defined voxel sizes constrains continuous adaptation across a wide range of spatial scales.
More recently, Cheng~\etalcite{cheng2025ral} proposed LIVOX-CAM, which adjusts the voxel size by first performing a temporary voxelization with a fixed voxel size $d_\mathrm{temp}^\mathrm{fixed}$ and then updating it based on the ratio between the resulting and desired point counts; see Fig.~\ref{fig:adap_vox_diagram_comparison}(c).
It updates the voxel size through a volume-based scaling strategy, unlike LOCUS\,2.0~\cite{reinke2022ral}, which relies on a linear adjustment.
Similar to LOCUS\,2.0, LIVOX-CAM aims to maintain a fixed desired number of points~$N_\mathrm{desired}^\mathrm{fixed}$, leaving open the question of how this target should adapt to continuous changes in scene~scale.

\begin{figure*}[!t]
	\centering
	\includegraphics[width=\textwidth]{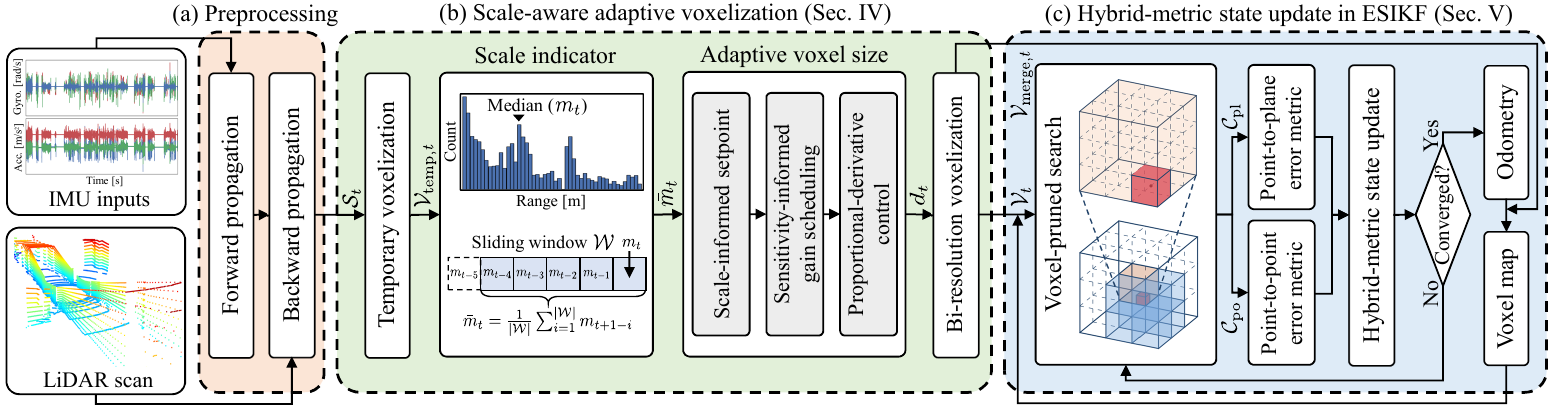}
	\vspace{-0.3cm}
	\caption{System overview of \textit{GenZ-LIO}. (a) In the preprocessing stage, forward propagation uses IMU measurements to propagate the state and covariance, and backward propagation removes motion distortion from the LiDAR scan, yielding the deskewed scan $\mathcal{S}_t$. (b) The scan $\mathcal{S}_t$ is voxelized using the voxel size from the previous timestep to produce the temporary voxelized scan $\mathcal{V}_{\mathrm{temp},t}$. The median range $m_t$ of points in $\mathcal{V}_{\mathrm{temp},t}$ is inserted into a sliding window to compute the spatial-scale indicator $\bar m_t$. Based on $\bar m_t$, a target number of voxelized points is set as a scale-informed control setpoint, and the voxel size~$d_t$ is adaptively adjusted via a PD controller with gain scheduling. The updated $d_t$ is then used for bi-resolution voxelization of $\mathcal{S}_t$, yielding $\mathcal{V}_{\mathrm{merge},t}$ with~$d_t/2$ for map integration with reduced discretization error and $\mathcal{V}_t$ with $d_t$ for state update. (c) The voxelized scan $\mathcal{V}_t$ is aligned with the voxel map for a voxel-pruned correspondence search, which avoids unnecessary traversal of neighboring voxels. This process produces the point-to-plane correspondence set~$\mathcal{C}_\mathrm{pl}$ and point-to-point correspondence set~$\mathcal{C}_\mathrm{po}$, which are used in the hybrid-metric state update. Finally, the transformed $\mathcal{V}_{\mathrm{merge},t}$ is integrated into the voxel map. For clarity, the control flow of the scale-aware adaptive voxelization is further illustrated in Fig.~\ref{fig:adap_vox_diagram_comparison}(d), where the bi-resolution voxelization step is omitted for comparison with other adaptive voxelization strategies.}
	\label{fig:genz-lio_flowchart}
	\vspace{-0.43cm}
\end{figure*}

In this study, we build on the computational efficiency and implementation simplicity of adaptive voxel grid sampling, while addressing key limitations of existing methods, including fixed target point counts~\cite{reinke2022ral, cheng2025ral} and discrete voxel size switching~\cite{lim2023ur}.
In contrast to these approaches, our method explicitly estimates the spatial scale using the scale indicator~$\bar m_t$ and determines a scale-informed target number of voxelized points~$N_{\mathrm{desired},t}$.
The voxel size is then regulated via a feedback controller with gain scheduling to drive the voxelized point count toward this target, enabling stable and responsive setpoint tracking under varying spatial scales; see~Fig.~\ref{fig:adap_vox_diagram_comparison}(d).

% 2. Error metrics
After downsampling the incoming LiDAR scan, the subsequent state estimation stage computes residuals from correspondences established between the current downsampled scan and a reference (e.g., another scan or a local submap) using various error metrics.
A representative feature-based line of work originates from LOAM~\cite{zhang2014rss}, which estimates ego-motion by extracting edge and planar features, and registering them to a sparse feature map.
This design has influenced several subsequent systems, including LeGO-LOAM~\cite{shan2018iros}, which incorporates ground constraints, and F-LOAM~\cite{wang2021iros}, which improves computational efficiency through a revised optimization strategy.
Although LOAM-style methods have demonstrated strong performance, their parameter settings often need to be adjusted according to LiDAR resolution, point density, and scene structure~\cite{vizzo2023ral}.

Recently, many modern LIO systems have adopted direct scan-to-map residual formulations, which commonly use point-to-point~\cite{besl1992tpami, wu2024icra}, point-to-plane~\cite{rusinkiewicz2001IntConfThreeDDigitalImagingAndModeling, xu2021ral, xu2022tro, bai2022ral, he2023ais}, or generalized-ICP (G-ICP)~\cite{segal2009rss, chen2023icra, chen2024ral}-based error metrics.
For instance, Xu~\etalcite{xu2022tro} employed the point-to-plane metric in FAST-LIO2, an enhanced version of FAST-LIO~\cite{xu2021ral} featuring direct matching, an ikd-Tree, and a novel Kalman gain formulation.
Bai~\etalcite{bai2022ral} also utilized the point-to-plane metric in Faster-LIO, introducing parallel sparse incremental voxel updates for lightweight operation. 
Chen~\etalcite{chen2023icra} applied the G-ICP-based metric~\cite{koide2021icra} in DLIO with continuous-time motion correction to mitigate distortion.
He~\etalcite{he2023ais} leveraged the point-to-plane metric in Point-LIO to perform state updates at every LiDAR point measurement.
Wu~\etal~\cite{wu2024icra} adopted the point-to-point metric in LIO-EKF, employing adaptive thresholding for more robust data association.
Chen~\etalcite{chen2024ral} integrated G-ICP-based constraints with inertial data in iG-LIO. 

While these methods have demonstrated their effectiveness across various scenarios, relying solely on a single error metric can limit estimation accuracy depending on the geometric characteristics of the surroundings~\cite{rusinkiewicz2001IntConfThreeDDigitalImagingAndModeling, vizzo2023ral, lee2024ral}.
For example, the point-to-point error metric~\cite{besl1992tpami} does not exploit planar information, and thus tends to be less accurate than the point-to-plane metric in structured environments~\cite{rusinkiewicz2001IntConfThreeDDigitalImagingAndModeling, lee2024ral}.
Conversely, the point-to-plane error metric~\cite{rusinkiewicz2001IntConfThreeDDigitalImagingAndModeling} relies on estimated surface normals, which may become less reliable when LiDAR points are sparse or unstructured, potentially affecting alignment accuracy~\cite{vizzo2023ral, lim2024ijrr, lee2024ral}.
Furthermore, rejecting correspondences associated with less reliable normals can reduce geometric constraints in certain directions, which may lead to pose drift, particularly in degenerate environments such as long corridors or open spaces~\cite{lim2024ijrr, tuna2024tro, lee2024ral}.
Similarly, G-ICP~\cite{segal2009rss} operates as a plane-to-plane error metric by assuming that all the surroundings are local planes.
Under this formulation, however, it is also subject to the limitations of the point-to-plane metric, particularly when surface normals are unreliable or the local planar assumption does not hold, which may influence registration accuracy in certain scenarios~\cite{lee2024ral}.

To address the inherent limitations of individual error metrics, Lee~\etalcite{lee2024ral} revisited both the point-to-plane~\cite{rusinkiewicz2001IntConfThreeDDigitalImagingAndModeling} and point-to-point~\cite{besl1992tpami} metrics, leveraging their complementary strengths in GenZ-ICP.
Specifically, the point-to-plane error metric is applied to correspondences on structured surfaces where reliable surface normals can be estimated, while the point-to-point error metric is applied to correspondences in unstructured or sparse regions to avoid using unreliable normals.
This adaptive metric selection improves robustness to geometric variation and mitigates optimization degradation by generating constraints along a broader range of directions.

However, GenZ-ICP assigns a uniform weight to all correspondences within each residual type, without accounting for the varying uncertainty of individual measurements.
To overcome this limitation, we extend the uncertainty-based weighting strategy proposed by Yuan~\etalcite{yuan2022ral}, originally formulated for point-to-plane~\cite{rusinkiewicz2001IntConfThreeDDigitalImagingAndModeling} correspondences, to a hybrid residual formulation.
Specifically, we formulate a covariance model for point-to-point~\cite{besl1992tpami} residuals and augment it with a discretization-error variance to account for uncertainty induced by voxel-based map discretization.
This formulation allows the optimization to prioritize more reliable observations while reducing the influence of noisy or uncertain measurements.
We refer to this reliability-aware integration of point-to-plane and point-to-point residuals as the hybrid-metric state update.
For a more comprehensive comparison, Table~\ref{tab:methodological_comparison_paper} in Sec.~\ref{sec:methodological_comparison} explicitly compares GenZ-LIO with closely related methods in terms of adaptive voxelization and state update.

%%%%%%%%%%%%%%%%%%%%%%%%%%%%%%%%%%%%%%%%%%%%%%%%%%%%%%%%%%%%%%%%%%%%%%%%%%%%%%%%
\section{SYSTEM OVERVIEW}
\label{sec:system overview}
As illustrated in Fig.~\ref{fig:genz-lio_flowchart}, GenZ-LIO is designed to maintain robust and efficient odometry estimation across a wide range of spatial scales, spanning both confined and open environments.
To this end, the scale-aware adaptive voxelization module estimates the spatial scale and determines the voxel size via a feedback controller with gain scheduling, driving the voxelized point count toward a scale-informed setpoint~(Fig.~\ref{fig:genz-lio_flowchart}(b)).
The resulting voxelized scan is then processed by a reliability-aware hybrid-metric state update, in which point-to-plane~\cite{rusinkiewicz2001IntConfThreeDDigitalImagingAndModeling} and point-to-point~\cite{besl1992tpami} residuals are weighted according to their estimated reliabilities~(Fig.~\ref{fig:genz-lio_flowchart}(c)).
In contrast to point-to-plane-only approaches~\cite{xu2022tro,bai2022ral,lim2023ur,he2023ais,hviktortsoi2023pvlio}, which can be sensitive to environments with only a few reliable planar regions, the proposed update jointly uses planar and non-planar geometric constraints to improve robustness across diverse local structures.
However, the point-to-point correspondence search introduced by this update can increase computational cost; thus, the voxel-pruned search strategy, described in Sec.~\ref{subsec: voxel-pruned correspondence search}, is employed to avoid redundant computations without degrading odometry accuracy.

\section{SCALE-AWARE ADAPTIVE VOXELIZATION}
\label{sec:scale-aware adaptive voxelization}
The purpose of scale-aware adaptive voxelization is to adapt the scan downsampling resolution to the spatial scale of the current LiDAR scan.
To this end, we use a lightweight range-based scale indicator to determine a target number of voxelized points and regulate the voxel size through feedback control so that the voxelized point count tracks this target.
This formulation is designed for real-time field deployment, where adapting the scan resolution to confined--open scene changes can improve robustness and computational efficiency without requiring expensive scene analysis.

\subsection{SCALE INDICATOR}
We begin by voxelizing the deskewed scan $\mathcal{S}_t$ using the voxel size $d_{t-1}$ from the previous timestep, resulting in a temporary voxelized scan $\mathcal{V}_{\mathrm{temp},t}$, where $d_{0}$ denotes the user-defined initial voxel size.
To represent the spatial scale of the current scene, we first compute the range of each point in $\mathcal{V}_{\mathrm{temp},t}$ with respect to the LiDAR frame as $\sqrt{x^2 + y^2 + z^2}$ and define the median of these ranges as $m_t$.
The rationale for using the median range as a lightweight spatial-scale statistic is discussed in Sec.~\ref{sec:median_scale_indicator}.

However, directly using the median range~$m_t$ as a scale indicator can exhibit high-frequency variations due to scan-level fluctuations, making the subsequent voxel size adjustment overly sensitive.
To mitigate this, we compute the smoothed median range $\bar m_t = \frac{1}{|\mathcal{W}|}\sum_{i=1}^{|\mathcal{W}|} m_{t+1-i}$ by applying a moving average, where $\mathcal{W}$ is a sliding window that stores up to $N_w$ recent median ranges, and $|\mathcal{W}|$ denotes the number of stored medians~($|\mathcal{W}|\leq N_w$).
The resulting $\bar m_t$ serves as a lightweight proxy for the spatial extent represented by the current LiDAR scan rather than a sensor-invariant property of the physical environment.

\begin{figure}[!t]
	\centering
	\includegraphics[width=\columnwidth]{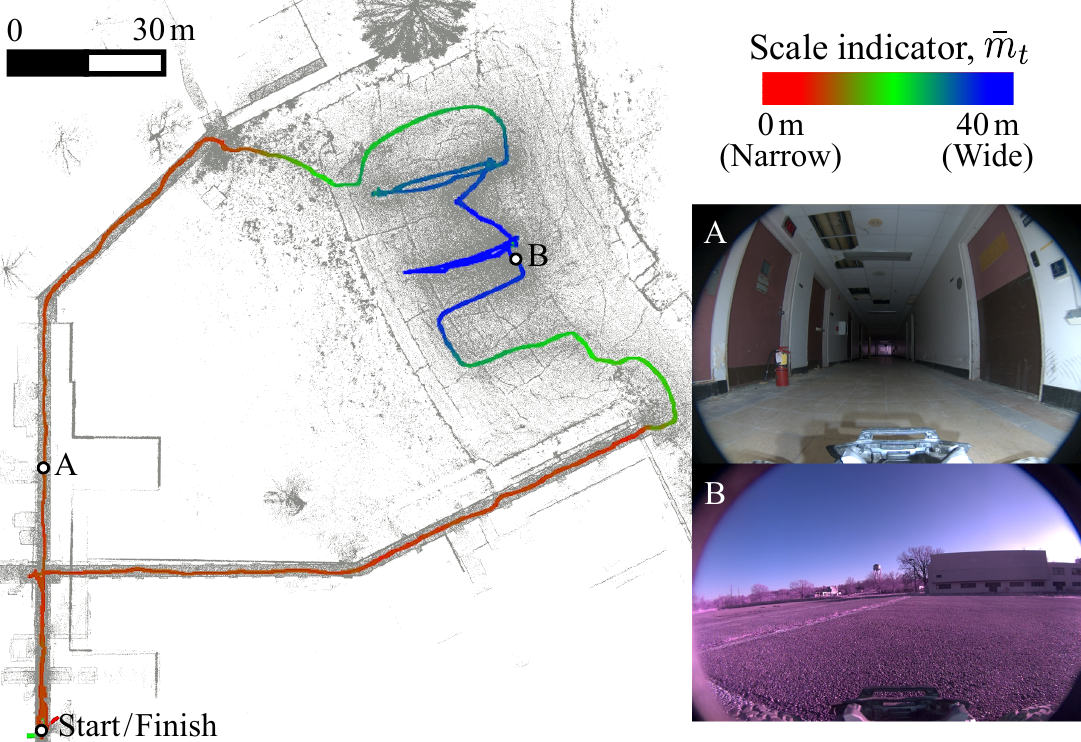}
	\vspace{-0.15cm}
	\caption{Our scale indicator $\bar{m}_t$ mapped onto the estimated trajectory for the \dataset{Corridor\,02} sequence of the \dataset{SuperLoc}~\cite{zhao2025icra} dataset, demonstrating its variation with the scene’s spatial scale.}
	\label{fig:median_traj}
	\vspace{-0.5cm}
\end{figure}

As illustrated in Fig.~\ref{fig:median_traj}, the scale indicator $\bar m_t$ is proportional to the spatial scale, yielding smaller values in confined scenes (region~A in Fig.~\ref{fig:median_traj}) and larger values in open areas (region~B in~Fig.~\ref{fig:median_traj}).
This property is used to determine the desired number of voxelized points in Sec.~\ref{subsec:scale-informed setpoint}.
Moreover, the scale indicator $\bar m_t$, together with tracking error-related terms, is used for gain scheduling in Sec.~\ref{subsec:sensitivity-informed gain scheduling}.

\subsection{MOTIVATION FOR PID CONTROL IN ADAPTIVE VOXELIZATION}
The scale indicator $\bar m_t$ is used to determine the desired number of voxelized points, $N_{\mathrm{desired},t}$, rather than an information-optimal point count.
Larger values of $\bar m_t$ increase this target to retain sufficient geometric support in open scenes, whereas smaller values decrease it to avoid unnecessary computation in confined scenes.
An empirical analysis supporting this scale-dependent setpoint design is provided in Sec.~\ref{sec:scale_dependent_setpoint}.

The voxel size is then adjusted to drive the resulting voxelized point count toward $N_{\mathrm{desired},t}$.
However, the relationship between voxel size and the voxelized point count is scene-dependent and nonlinear; it evolves continuously as the scene changes and is difficult to model reliably in a closed form.
This motivates the use of feedback control, and a PID controller provides a natural starting point for regulating the voxel size so that the voxelized point count tracks $N_{\mathrm{desired},t}$ through the proportional, integral, and derivative responses to tracking error.

In our case, however, the setpoint $N_{\mathrm{desired},t}$ is not fixed but varies over time with the spatial scale (see Sec.~\ref{subsec:scale-informed setpoint}).
In such a time-varying setpoint tracking problem, the integral term can accumulate transient errors induced by setpoint changes and retain stale corrective actions, potentially leading to overshoot, oscillation, or delayed settling.
Mitigating these effects while retaining the integral term would require additional mechanisms, such as conditional integration, anti-windup, or decay/reset strategies, together with extra user-defined parameters and increased tuning effort.
Therefore, in view of these considerations, we adopt proportional-derivative (PD) control in this work, retaining the proportional and derivative components while excluding the integral term.
Based on this design choice, the components required for PD control are addressed in the following subsections: the control setpoint in Sec.~\ref{subsec:scale-informed setpoint}, the tracking error and its derivative in Sec.~\ref{subsec: tracking error formulation}, and gain scheduling in Sec.~\ref{subsec:sensitivity-informed gain scheduling}.

\subsection{SCALE-INFORMED SETPOINT}
\label{subsec:scale-informed setpoint}
The scale-informed setpoint $N_{\mathrm{desired},t}$, which serves as the control target, is defined as follows:
\begin{equation} \label{eq: scale-informed setpoint}
N_{\mathrm{desired},t}
= N_{\min} + (N_{\max}-N_{\min})\,\rho(\bar m_t),
\end{equation}
where $N_{\min}$ and $N_{\max}$ denote the user-defined lower and upper bounds of the desired number of voxelized points, respectively.
%Here, $N_{\min}$ prevents the voxelized scan from becoming overly sparse, while $N_{\max}$ bounds the computational cost arising from processing an excessively large number of points.
The interpolation function $\rho(\bar m_t)$ is defined using a saturating power function as follows:
\begin{equation}\label{eq:rho}
\rho(\bar m_t)
= 1 - \left(1 - \frac{\min(\bar m_t,\tau_m)}{\tau_m}\right)^{p},
\qquad p > 1,
\end{equation}
where $\tau_m$ denotes the saturation threshold of the scale indicator beyond which $N_{\mathrm{desired},t}$ saturates at $N_{\max}$.
The function satisfies $\rho(0)=0$ and $\rho(\tau_m)=1$, and its first derivative satisfies $\rho'(\tau_m)=0$.
By enforcing a zero slope at the boundary $\bar m_t=\tau_m$, the setpoint $N_{\mathrm{desired},t}$ varies smoothly at the saturation point, avoiding abrupt derivative changes that could degrade the transient response of the proposed controller, such as unintended oscillations.
The constant exponent $p$ controls the growth rate of the setpoint, allowing the responsiveness of the setpoint evolution to be adjusted via a single parameter.
%Overall, the proposed closed-form setpoint design offers smooth setpoint evolution with lightweight computational overhead, making it suitable for real-time operation.

%This design is motivated by two practical considerations.
%First, from a control perspective, enforcing a zero slope at the boundary $\bar m_t=\tau_m$ allows $N_\mathrm{desired}$ to vary smoothly near the saturation point, avoiding abrupt changes in the setpoint rate that could lead to oscillations in the PD controller.
%Second, from an information gain perspective, the setpoint is designed to increase more rapidly at smaller scales, where expanding the observable range is more likely to introduce new geometric structures (e.g., planes or edges) and thus provide relatively higher marginal information gain.
%As the number of points increases, this gain gradually diminishes, since a larger point set increasingly contains redundant observations of already captured structures.
%The resulting saturating form of $\rho(\bar m_t)$ reflects this tendency by allowing $N_\mathrm{desired}$ to grow quickly at first and then gradually level off toward $N_{\max}$.
%The exponent $p$ modulates the growth rate of the setpoint, enabling the responsiveness to be tuned via a single parameter while preserving a lightweight closed-form expression suitable for real-time operation.

\subsection{TRACKING ERROR FORMULATION}
\label{subsec: tracking error formulation}
To track the scale-informed setpoint $N_{\mathrm{desired},t}$ in~(\ref{eq: scale-informed setpoint}), we define the tracking error $e_t$ and its derivative $\Delta e_t$ for the PD controller as follows:
\begin{equation} \label{eq: tracking_error}
e_t \; = \; N_{\mathrm{desired},t} - N_{\mathrm{temp},t}, \quad
\Delta e_t = \frac{e_t - e_{t-1}}{\Delta t_\mathrm{scan}},
\end{equation}
where $N_{\mathrm{temp},t}$ is the number of points in the temporarily voxelized scan $\mathcal{V}_{\mathrm{temp},t}$, and $\Delta t_\mathrm{scan}$ denotes the time interval between successive scans.

\subsection{SENSITIVITY-INFORMED GAIN SCHEDULING}
\label{subsec:sensitivity-informed gain scheduling}
Subsequently, $e_t$ and $\Delta e_t$ in (\ref{eq: tracking_error}) are input to the PD controller along with proportional and derivative gains, $K_{p,t}$ and $K_{d,t}$, respectively.
For linear systems, a standard PD controller with fixed gains can be sufficient.
As discussed earlier, however, the relationship between voxel size and voxelized point count is nonlinear, meaning that fixed gains can lead to oscillations or delayed setpoint tracking; see Sec.~\ref{subsec: exp_ablation_gain_scheduling}.

For instance, in narrow and highly enclosed environments (i.e., low $\bar m_t$), scan points are concentrated within a limited space, leading to high point density.
Under such conditions, even small variations in voxel size can produce large changes in the voxelized point count.
As a result, high gains tend to amplify overshoot and oscillations, making lower gains more suitable.
Conversely, in wide and open environments (i.e., high $\bar m_t$), scan points are distributed over a larger space, resulting in sparser point distributions.
In this case, the voxelized point count is less sensitive to changes in voxel size, and higher gains are often required for more aggressive compensation.
Therefore, the scale indicator $\bar m_t$ can be used to infer the sensitivity of the voxelized point count to voxel size changes in the current scene and to adjust the control gains accordingly.

In addition to the scale indicator $\bar m_t$, the magnitudes of the tracking error $\lvert e_t \rvert$ and its derivative $\lvert \Delta e_t \rvert$ can also be determinants of gain selection.
If the voxelized point count is already close to $N_{\mathrm{desired},t}$ (i.e., low $\lvert e_t \rvert$), a smaller proportional gain promotes smoother tracking with reduced oscillation.
When the tracking error is large (i.e., high $\lvert e_t \rvert$), a larger proportional gain enables stronger compensation and faster setpoint tracking.
Likewise, if the derivative magnitude $\lvert \Delta e_t \rvert$ is small (i.e., low $\lvert \Delta e_t \rvert$), a smaller derivative gain is sufficient because the tracking error is changing slowly.
In contrast, when $\lvert \Delta e_t \rvert$ is large (i.e., high $\lvert \Delta e_t \rvert$), a larger derivative gain helps damp rapid error variation, thereby reducing overshoot and oscillation.

Motivated by these observations, we propose a sensitivity-informed gain scheduling strategy in which the scale indicator~$\bar m_t$ is jointly considered with the tracking error magnitude~$\lvert e_t \rvert$ for the proportional gain~$K_{p,t}$, and with the magnitude of the error derivative~$\lvert \Delta e_t \rvert$ for the derivative gain~$K_{d,t}$.
To this end, the spatial-scale indicator $\bar m_t$ is normalized to $\phi_t = \frac{\min\!\left(\bar m_t,\, \tau_m\right)}{\tau_m}$.
Similarly, the magnitudes of the tracking error $\lvert e_t \rvert$ and its derivative $\lvert \Delta e_t \rvert$ are normalized to $\psi_{p,t}$ and $\psi_{d,t}$, respectively, as follows:
\begin{equation}
\begin{aligned}
\psi_{p,t} &= \frac{\min\!\left(\lvert e_t \rvert,\, \lambda_p N_{\mathrm{desired},t}\right)}
{\lambda_p N_{\mathrm{desired},t}},\\
\psi_{d,t} &= \frac{\min\!\left(\lvert \Delta e_t \rvert,\, \lambda_d N_{\mathrm{desired},t}/\Delta t_\mathrm{scan}\right)}
{\lambda_d N_{\mathrm{desired},t}/\Delta t_\mathrm{scan}},
\end{aligned}
\end{equation}
where $\lambda_p \in (0,1]$ and $\lambda_d \in (0,1]$ are empirically chosen scaling factors that compensate for the scale mismatch between the setpoint~$N_{\mathrm{desired},t}$ in~(\ref{eq: scale-informed setpoint}) and the magnitudes of the tracking error and its derivative in~(\ref{eq: tracking_error}), respectively.

The interpolation factors $\Gamma_{p,t}$ and $\Gamma_{d,t}$ for the proportional and derivative gains, respectively, are then computed using the geometric mean of the corresponding normalized values as
\begin{equation}
\Gamma_{p,t} = \sqrt{\, \phi_t \cdot \psi_{p,t} \,},
\qquad
\Gamma_{d,t} = \sqrt{\, \phi_t \cdot \psi_{d,t} \,}.
\end{equation}
These interpolation factors are used to determine the proportional and derivative gains as follows:
\begin{equation} \label{eq: computed gains}
\begin{aligned}
K_{p,t} &= K_{p,\min} + \left(K_{p,\max} - K_{p,\min}\right)\Gamma_{p,t}, \\
K_{d,t} &= K_{d,\min} + \left(K_{d,\max} - K_{d,\min}\right)\Gamma_{d,t},
\end{aligned}
\end{equation}
where $K_{p,\min}$, $K_{p,\max}$, $K_{d,\min}$, and $K_{d,\max}$ denote user-defined gain bounds.

Through the geometric mean, the proposed gain scheduling incorporates both the scale indicator and the corresponding error-related magnitude into each gain.
For the proportional gain, consider a narrow scene (i.e., low $\phi_t$) with large tracking error (i.e., high $\psi_{p,t}$).
If only the spatial scale were considered, the resulting gain would remain low, leading to slow compensation.
Conversely, if only the tracking error magnitude were considered, the resulting gain could become excessively high, potentially inducing oscillations in narrow scenes where the voxelized point count is highly sensitive to small changes in voxel size.
By accounting for both $\phi_t$ and $\psi_{p,t}$, $\Gamma_{p,t}$ yields a proportional gain that is higher than that based on spatial scale alone, promoting faster compensation, yet lower than that based on tracking error magnitude alone, mitigating oscillations.

A similar rationale applies to the derivative gain.
In narrow scenes, using only the spatial scale would keep the derivative gain low even when the error derivative is large, thereby weakening damping against rapid error variation.
Conversely, using only the derivative magnitude could produce an excessively large derivative gain, which may overreact to transient fluctuations.
By jointly accounting for $\phi_t$ and $\psi_{d,t}$, $\Gamma_{d,t}$ adjusts the derivative gain in a balanced manner, strengthening damping when the error changes rapidly while avoiding unnecessarily aggressive responses in highly sensitive scenes.
A comparison of the geometric mean with alternative averaging strategies is provided in Sec.~\ref{sec:geometric_mean_gain_interpolation}.

%While other averaging strategies can be considered for gain interpolation, we adopt the geometric mean for its balance between responsiveness and conservativeness.
%Unlike the arithmetic mean, which can be disproportionately affected by a single large input and may result in excessive gains even when the other input is small, potentially inducing oscillations, the geometric mean mitigates such effects by increasing only when both inputs are large.
%Although the harmonic mean also suppresses dominant values, it places excessive emphasis on the smaller input, often resulting in overly conservative gains and slower convergence.
%In contrast, the geometric mean provides a more balanced modulation between aggressive and cautious control responses, enabling stable and faster convergence across varying conditions (see~\ref{}).

% Algorithm 1
\begin{algorithm}[!t]
	\caption{Scale-aware adaptive voxelization}
	\label{alg:adaptive_voxelization}
	\SetKwInOut{Input}{Input}
	\SetKwInOut{Output}{Output}
	
	\Input{Deskewed scan $\mathcal{S}_t$; previous voxel size $d_{t-1}$; previous tracking error $e_{t-1}$; scan interval $\Delta t_\mathrm{scan}$; voxel size bounds $[d_{\min},d_{\max}]$; window size $N_w$; point count bounds $[N_{\min}, N_{\max}]$; spatial-scale threshold~$\tau_m$; scaling factors $\lambda_p$, $\lambda_d$; exponent $p$; gain bounds $[K_{p,\min},K_{p,\max}]$, $[K_{d,\min},K_{d,\max}]$.}
	\Output{Voxelized scans $\mathcal{V}_t,\, \mathcal{V}_{\mathrm{merge},t}$.}
	
	\textcolor{gray}{\tcp{Scale indicator}}
	$\mathcal{V}_{\mathrm{temp},t} \gets \mathrm{Voxelize}(\mathcal{S}_t,\, d_{t-1})$\,;\\
	$N_{\mathrm{temp},t} \gets \mathrm{Count}(\mathcal{V}_{\mathrm{temp},t})$\,;\\
	Let $R = \{\lVert \mathbf{p}_i\rVert \mid \mathbf{p}_i \in \mathcal{V}_{\mathrm{temp},t}\}$\,;\\
	$m_t \gets \mathrm{median}(R)$\,;\\
	Append $m_t$ to window $\mathcal{W}$; \If{$|\mathcal{W}| > N_w$}{remove oldest element from $\mathcal{W}$\,;}%
	$\bar m_t \gets \frac{1}{|\mathcal{W}|}\sum_{i=1}^{|\mathcal{W}|} m_{t+1-i}$\,;
	
	\textcolor{gray}{\tcp{Scale-informed setpoint}}
	$\rho(\bar m_t) \gets 1 - \left(1 - \frac{\min(\bar m_t,\tau_m)}{\tau_m}\right)^{p}$\,;\\
	$N_{\mathrm{desired},t} \gets N_{\min} + (N_{\max} - N_{\min})\, \rho(\bar m_t)$\,;
	
	\textcolor{gray}{\tcp{Tracking error formulation}}
	$e_t \gets N_{\mathrm{desired},t} - N_{\mathrm{temp},t}$\,;\\
	$\Delta e_t \gets (e_t - e_{t-1}) / \Delta t_\mathrm{scan}$\,;
	
	\textcolor{gray}{\tcp{Sensitivity-informed gain~scheduling}}
	$\phi_t \gets \dfrac{\min(\bar m_t,\, \tau_m)}{\tau_m}$\,;\\[0.5ex]
	$\psi_{p,t} \gets \dfrac{\min(\lvert e_t\rvert,\, \lambda_p N_{\mathrm{desired},t})}{\lambda_p N_{\mathrm{desired},t}}$\,;\\[0.5ex]
	$\psi_{d,t} \gets \dfrac{\min\!\left(\lvert \Delta e_t\rvert,\, \lambda_d N_{\mathrm{desired},t}/\Delta t_\mathrm{scan}\right)}{\lambda_d N_{\mathrm{desired},t}/\Delta t_\mathrm{scan}}$\,;\\[0.5ex]
	$\Gamma_{p,t} \gets \sqrt{\phi_t \cdot \psi_{p,t}}$\,;\\
	$\Gamma_{d,t} \gets \sqrt{\phi_t \cdot \psi_{d,t}}$\,;\\
	$K_{p,t} \gets K_{p,\min} + (K_{p,\max} - K_{p,\min})\,\Gamma_{p,t}$\,;\\
	$K_{d,t} \gets K_{d,\min} + (K_{d,\max} - K_{d,\min})\,\Gamma_{d,t}$\,;
	
	\textcolor{gray}{\tcp{Voxel size update}}
	$\Delta d_t \gets - K_{p,t}\, e_t - K_{d,t}\, \Delta e_t$\,;\\
	$d_t \gets \mathrm{clamp}\!\left(d_{t-1} + \Delta d_t,\; d_{\min},\; d_{\max}\right)$\,;\\
	$e_{t-1} \gets e_t$\,;
	
	\textcolor{gray}{\tcp{Bi-resolution voxelization}}
	$\mathcal{V}_{\mathrm{merge},t} \gets \mathrm{Voxelize}(\mathcal{S}_t,\, d_t/2)$\,;\\
	$\mathcal{V}_t \gets \mathrm{Voxelize}(\mathcal{V}_{\mathrm{merge},t},\, d_t)$\,;\\
	
	\Return{$\mathcal{V}_t,\, \mathcal{V}_{\mathrm{merge},t}$.}
\end{algorithm}

Following the PD control law, the voxel size adjustment~$\Delta d_t$ is computed using $K_{p,t}$ and $K_{d,t}$ in~(\ref{eq: computed gains}) as follows:
\begin{equation}
\Delta d_t= -K_{p,t} e_t - K_{d,t} \Delta e_t.
\end{equation}
The negative signs in both terms reflect the inverse relationship between voxelized point count and voxel size.
When the error is positive, which indicates that $N_{\mathrm{temp},t}$ is lower than $N_{\mathrm{desired},t}$, the voxel size needs to be decreased to increase the point count.
The voxel size is then updated as follows:
\begin{equation}\label{eq: updated voxel size}
d_t = \mathrm{clamp}\bigl(d_{t-1} + \Delta d_t, d_{\min}, d_{\max}\bigr),
\end{equation}
where $\mathrm{clamp}(\cdot,\cdot,\cdot)$ constrains $d_t$ to the user-defined interval $[d_{\min}, d_{\max}]$ to prevent the voxelized scan from becoming excessively sparse or dense.

\subsection{BI-RESOLUTION VOXELIZATION} \label{subsec: bi-resolution voxelization}
Next, $d_t$ in~(\ref{eq: updated voxel size}) is used in a bi-resolution voxelization, which is widely adopted in existing LiDAR-based odometry approaches~\cite{dellenbach2022icra, vizzo2023ral} to reduce discretization errors by retaining higher density in map update scans.
First, the current LiDAR scan $\mathcal{S}_t$ is voxelized with $d_t/2$ to produce $\mathcal{V}_{\mathrm{merge},t}$ for map update.
Then, $\mathcal{V}_{\mathrm{merge},t}$ is re-voxelized using $d_t$ to obtain $\mathcal{V}_t$, which is used for state update, as illustrated in Fig.~\ref{fig:genz-lio_flowchart}.
This bi-resolution design makes the scan used for map update denser than that used for state update, reducing discretization errors in the voxel map and thereby improving the reliability of subsequent scan-to-map correspondence matching.

In summary, the proposed scale-aware adaptive voxelization adjusts the voxel size according to the spatial scale, improving the robustness and computational efficiency of LiDAR-based odometry across a wide range of spatial scales; see Secs.~\ref{subsec: exp_benchmark} and~\ref{subsec: exp_adap_vox_comparison}.
Moreover, the proposed sensitivity-informed gain scheduling strategy mitigates oscillations in voxel size control and enables improved stability and responsiveness in tracking the scale-informed setpoint; see Sec.~\ref{subsec: exp_ablation_gain_scheduling}.
Algorithm~\ref{alg:adaptive_voxelization} details the proposed method, and Fig.~\ref{fig:adap_vox_diagram_comparison} further contrasts it with other adaptive voxelization strategies~\cite{reinke2022ral, lim2023ur, cheng2025ral} through control flow diagrams.

\section{HYBRID-METRIC STATE UPDATE IN ERROR-STATE ITERATED KALMAN FILTER}
\label{sec:hybrid-metric state update}
The voxelized scan $\mathcal{V}_t$ is used for correspondence search, residual computation, and state update, as illustrated in Fig.~\ref{fig:genz-lio_flowchart}(c).
Before detailing the state update procedure, the notations and assumptions are summarized in Sec.~\ref{subsec: assumptions and notations}.
The voxel-pruned correspondence search is then described in Sec.~\ref{subsec: voxel-pruned correspondence search}, the residual formulation and associated Jacobians and covariances are presented in Secs.~\ref{subsec:Setting Point-to-Plane Residual, Jacobian, and Covariance} and~\ref{subsec:Setting Point-to-Point Residual, Jacobian, and Covariance}, and the hybrid-metric state update is detailed in Sec.~\ref{subsec:hybrid-metric state update}.

\begin{figure*}[t!]
	\centering
	\includegraphics[width=1.0\linewidth]{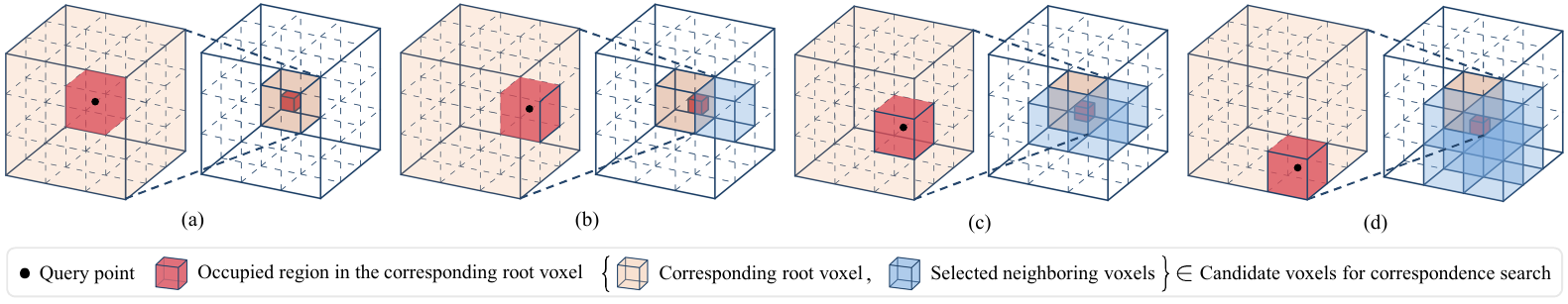}
	\vspace{-0.25cm}
	\caption{Candidate voxel selection based on the region occupied by a query point within its corresponding root voxel.
		The root voxel is divided into 27 regions, and the occupied region falls into one of four cases: (a) center case, selecting no neighboring voxel; (b) surface case, selecting one surface-sharing neighboring voxel; (c) edge case, selecting three edge-sharing neighboring voxels; and (d) corner case, selecting seven corner-sharing neighboring voxels.
		The selected neighboring voxels, together with the root voxel, are considered candidate voxels for correspondence search.
		In~Algorithm~\ref{alg: pruning strategy}, the function $\texttt{GetCandidateVoxels}$ selects the candidate voxels based on these sharing relations.}
	\label{fig:candidate voxel selection}
	\vspace{-0.3cm}
\end{figure*}

\subsection{ASSUMPTIONS AND NOTATIONS}\label{subsec: assumptions and notations}
The system employs a tightly coupled LiDAR--IMU configuration with a known temporal offset obtained through prior calibration or synchronization.
The IMU frame, denoted as~$I$, is adopted as the body frame, and the initial body frame is aligned with the world frame~$W$.
The LiDAR and IMU are rigidly mounted, with their extrinsic transformation pre-calibrated.
The notations used in the ESIKF are summarized in Table~\ref{table:notations_ESIKF}.
Although this section primarily focuses on the hybrid-metric state update (i.e., correction step), we briefly review the prediction step of the ESIKF~\cite{xu2022tro, zheng2025tro} to introduce the state variables and notations used in the subsequent derivations.

The discrete state transition model associated with the $i$-th IMU measurement is given by:
\begin{equation}\label{eq: state_transition}
\mathbf{x}_{i+1} = \mathbf{x}_{i} \boxplus \left(\Delta t_\mathrm{imu} \, f \left(\mathbf{x}_i, \mathbf{u}_i, \mathbf{w}_i \right)\right),
\end{equation}
where $\Delta t_\mathrm{imu}$ is the IMU sampling period.
The state manifold~$\mathcal{M}$, state $\mathbf{x}$, IMU input $\mathbf{u}$, process noise $\mathbf{w}$, and system function~$f(\cdot,\cdot,\cdot)$ are defined as follows:
\begin{equation} \label{eq: state definition}
\begin{aligned}
	&\mathcal{M} \triangleq SO(3) \times \mathbb{R}^{15},\ \text{dim}(\mathcal{M}) = 18,\\
	&\mathbf{x} \triangleq
	\begin{bmatrix}
		^W\mathbf{t}_{I}^\intercal & ^W\mathbf{R}_{I}^\intercal & ^W\mathbf{v}_{I}^\intercal & \mathbf{b}_{{g}}^\intercal & \mathbf{b}_{{a}}^\intercal & ^W\mathbf{g}^\intercal
	\end{bmatrix}^\intercal  \in \mathcal{M},\\
	&\mathbf{u} \triangleq
	\begin{bmatrix}
		\boldsymbol{\omega}_{m}^\intercal  & \mathbf{a}_{m}^\intercal
	\end{bmatrix}^\intercal, \hspace{0.2cm}
	\mathbf{w} \triangleq
	\begin{bmatrix}
		\bm{\delta}_{{g}}^\intercal  & \bm{\delta}_{{a}}^\intercal &
		\bm{\delta}_{\mathbf{b}_{g}}^\intercal  & \bm{\delta}_{\mathbf{b}_{a}}^\intercal
	\end{bmatrix}^\intercal,\\
	&f(\mathbf{x}, \mathbf{u}, \mathbf{w} ) \!=\!\!
	\begin{bmatrix}
		^W\mathbf{v}_I+\frac{1}{2}(^W\mathbf{R}_{I}\left( \mathbf{a}_{m} - \mathbf{b}_{{a}} - \bm{\delta}_{{a}}\right)+{^{W}\mathbf{g}} )\Delta t_\mathrm{imu} \vspace{0.1cm} \\
		\boldsymbol{\omega}_{m} - \mathbf{b}_{{g}} - \bm{\delta}_{{g}}  \\
		^W\mathbf{R}_{I}\left( \mathbf{a}_{m} - \mathbf{b}_{{a}} - \bm{\delta}_{{a}}\right) + {^{W}\mathbf{g}}  \\
		\bm{\delta}_{\mathbf{b}_{{g}}}\\
		\bm{\delta}_{\mathbf{b}_{{a}}} \\
		\mathbf{0}_{3\times 1}
	\end{bmatrix},
\end{aligned}
\end{equation}
where $^W\mathbf{t}_I$, $^W\mathbf{R}_I$, and $^W\mathbf{v}_I$ represent the IMU position, attitude, and velocity in the world frame, respectively, ${^W\mathbf{g}}$~is the gravity vector in the world frame, $\boldsymbol{\omega}_m$ and $\mathbf{a}_m$ are the measured angular velocity and linear acceleration with associated noises $\bm{\delta}_{g}$ and $\bm{\delta}_{a}$, respectively, and $\mathbf{b}_{g}$ and $\mathbf{b}_{a}$ are IMU biases modeled as random walks driven by Gaussian noise $\bm{\delta}_{\mathbf{b}_g}$ and $\bm{\delta}_{\mathbf{b}_a}$, respectively.

The state and covariance are propagated over the duration of a LiDAR scan using the available IMU measurements.
During this interval, forward propagation predicts the state at each IMU input~$\mathbf{u}_i$ by setting the process noise~$\mathbf{w}_i$ in~(\ref{eq: state_transition}) to zero.
Let $\widehat{\mathbf{x}}$ and $\widehat{\mathbf{P}}$ denote the propagated state and covariance, respectively, which are used as the prior for the subsequent update step as follows:
\begin{equation} \label{eq: prior distribution}
\widetilde{\mathbf{x}}=\mathbf{x}\boxminus\widehat{\mathbf{x}}\sim\mathcal{N}(\mathbf{0},\widehat{\mathbf{{P}}}),
\end{equation}
where the error state $\widetilde{\mathbf{x}}$ corresponds to the state $\mathbf{x}$ in (\ref{eq: state definition}) and is expressed as follows:
\begin{equation}
\widetilde{\mathbf{x}} = 
\begin{bmatrix}
{\bm{\delta}_{^{W}\mathbf{t}_{I}}}^\intercal,
{\bm{\delta}_{^{W}\mathbf{r}_{I}}}^\intercal,
{\bm{\delta}_{^{W}\mathbf{v}_{I}}}^\intercal,
{\bm{\delta}_{\mathbf{b}_{g}}}^\intercal,
{\bm{\delta}_{\mathbf{b}_{a}}}^\intercal,
{\bm{\delta}_{^{W}\mathbf{g}}}^\intercal
\end{bmatrix}^\intercal\in\mathbb{R}^{18\times1}.
\end{equation}

\begin{table}[!t]
	\caption{Some important notations in an error-state iterated Kalman filter.}
	\label{table:notations_ESIKF}
	\centering
	\renewcommand\arraystretch{0.8}
	\begin{tabular}{@{}>{\centering\arraybackslash}p{0.14\columnwidth}>{\raggedright\arraybackslash}p{0.81\columnwidth}@{}}
		\toprule[0.8pt]
		Notation  & Explanation \\
		\midrule
		$\boxplus$ / $\boxminus$ & The encapsulated ``boxplus'' and ``boxminus'' operations on the state manifold \\
		${^W(\cdot)}$ & A vector ${(\cdot)}$ in the world frame \\
		${^L(\cdot)}$ & A vector ${(\cdot)}$ in the LiDAR frame\\
		$^I\mathbf{T}_L$ & The extrinsic of the LiDAR frame w.r.t. the IMU frame \\
		$^W\mathbf T_{I}$ & The pose of the IMU frame w.r.t. the world frame \\
		$\mathbf{x}, \widehat{\mathbf{x}}$, $\bar{\mathbf{x}}$ & The true state, propagated estimate, and updated estimate\\
		$\widehat{\mathbf{x}}^{\ell}$ & The $\ell$-th update of $\mathbf{x}$ \\
		$\widetilde{\mathbf{x}}$ & The error state between ground truth $\mathbf{x}$ and its estimate $\widehat{\mathbf{x}}$ \\
		\bottomrule[0.8pt]
	\end{tabular}
	% \vspace{-5pt}
	\vspace{-0.0cm}
\end{table}

\subsection{VOXEL-PRUNED CORRESPONDENCE SEARCH} \label{subsec: voxel-pruned correspondence search}
The voxelized scan $\mathcal{V}_t$ obtained from Sec.~\ref{subsec: bi-resolution voxelization} is first transformed into the world frame using the pose prior for scan-to-map matching.
At this stage, to establish point-to-plane and point-to-point correspondences in a computationally efficient manner, we propose a voxel-pruned correspondence search that prunes unnecessary traversal of neighboring voxels.

\subsubsection{CANDIDATE VOXEL SELECTION}
Given a query point $^{W}\mathbf{p}$ from the transformed $\mathcal{V}_t$, the voxel map~\cite{yuan2022ral} is queried to identify the corresponding root voxel that contains $^{W}\mathbf{p}$.
Subsequently, correspondence search is performed over the root voxel and its neighboring voxels to find a point-to-plane or point-to-point correspondence.
In this process, we first determine a set of candidate voxels to visit for correspondence search.
That is, instead of performing a brute-force search over all 26 neighboring voxels of the root voxel~\cite{vizzo2023ral, wu2024icra}, we adaptively select only those neighboring voxels that can plausibly contain a valid correspondence, based on the position of the query point within the root voxel.

To this end, as illustrated in Fig.~\ref{fig:candidate voxel selection}, the root voxel, shown as a zoomed-in light orange voxel, is conceptually partitioned into 27 regions to determine the relative location of the query point within the voxel.
The occupied region of the query point is categorized into one of four cases—center, surface, edge, or corner—each corresponding to a distinct sharing relation with neighboring voxels.
Based on this classification, the root voxel and only a subset of neighboring voxels that share a face, edge, or corner with the occupied region are selected as candidate voxels for correspondence search.
For example, when the query point lies in a corner region of the root voxel, as shown in Fig.~\ref{fig:candidate voxel selection}(d), the root voxel and seven corner-sharing neighboring voxels are selected, resulting in eight candidate voxels in total.

%\subsubsection{Outlier rejection}
%A correspondence search is then performed over the selected candidate voxels.
%For outlier rejection during the correspondence search, we apply uncertainty-based filtering~\cite{yuan2022ral} for point-to-plane~\cite{rusinkiewicz2001IntConfThreeDDigitalImagingAndModeling} correspondences, and a variant of Euclidean distance-based adaptive filtering~\cite{vizzo2023ral} for point-to-point~\cite{besl1992tpami} correspondences.
%Although uncertainty-based filtering is theoretically applicable to point-to-point correspondences, it is not employed in our formulation due to its substantial computational cost.
%Unlike the scalar point-to-plane covariance, applying uncertainty-based filtering to point-to-point correspondences requires inverting a $3\times3$ covariance matrix for each candidate match.
%Given the large number of potential correspondences, these matrix inversions impose considerable overhead, making the approach impractical for real-time operation.
%Accordingly, we instead employ Euclidean distance–based filtering for point-to-point correspondences.
%We omit the detailed explanation of outlier rejection to focus on the main contributions of our method.

\begin{figure}[t!]
	\centering
	\includegraphics[width=0.9\linewidth]{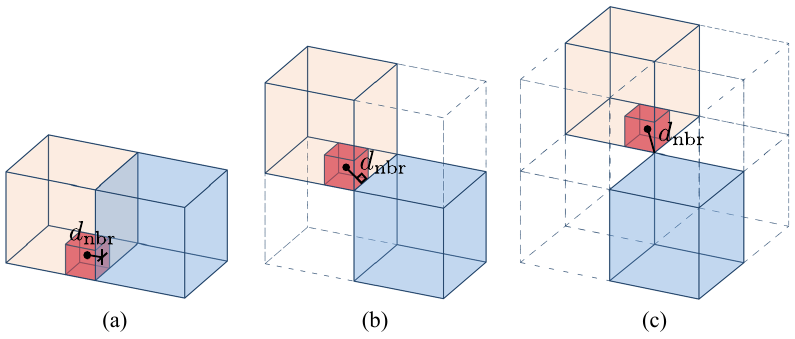}
	\vspace{0.2cm}
	\caption{Definition of the distance $d_\mathrm{nbr}$ between a query point and selected neighboring voxels.
		In Fig.~\ref{fig:candidate voxel selection}, the distance from the query point to a selected neighboring voxel falls into one of three categories:
		(a)~point-to-surface distance (surface-sharing relation);
		(b) point-to-edge distance (edge-sharing relation); and
		(c) point-to-corner distance (corner-sharing relation).
		In Algorithm~\ref{alg: pruning strategy}, the function $\texttt{ComputeDistanceToVoxel}$ determines the distance type based on these sharing relations between the query point’s occupied region and the selected neighboring voxel.}
	\label{fig:distance to candidate neighboring voxel}
	\vspace{-0.7cm}
\end{figure}

\subsubsection{POINT-TO-PLANE CORRESPONDENCE SEARCH}
Correspondence search is then performed over the candidate voxels selected above.
Each voxel stores an estimated normal vector~$^{W}\mathbf{n}$, a centroid~$^{W}\mathbf{q}$, and a set of accumulated points $\mathcal{P} = \left\{^{W}\acute{\mathbf{p}}_j \right\}_{j=1}^{N_\mathrm{stored}}$ from past observations, where $^{W}\acute{\mathbf{p}}_j$ denotes a stored past observation point expressed in the world frame, and $N_\mathrm{stored}$ is the number of stored points per voxel, upper-bounded by a user-defined maximum.
Point-to-plane matching directly leverages the stored normal and centroid and can therefore be evaluated at low computational cost.
Accordingly, we first attempt to establish a point-to-plane correspondence.

Using the stored normal and centroid, candidate point-to-plane correspondences are evaluated over the selected voxels with a statistical gating test based on the 3$\sigma$ criterion~\cite{yuan2022ral}.
If multiple candidate planes satisfy the criterion, the plane with the highest matching probability is selected as the final point-to-plane correspondence~\cite{yuan2022ral}.
This correspondence is then inserted into the point-to-plane correspondence set $\mathcal{C}_\mathrm{pl}$.

\subsubsection{POINT-TO-POINT CORRESPONDENCE SEARCH WITH ADDITIONAL VOXEL PRUNING}
When no valid point-to-plane correspondence is found, the algorithm falls back to point-to-point correspondence search rather than discarding the query point.
Compared with using only point-to-plane correspondences, this fallback helps form geometric constraints in more diverse directions, thereby mitigating degeneracy in the optimization~\cite{lee2024ral}; see Sec.~\ref{subsec: exp_hybrid_metric}.

However, unlike point-to-plane matching, point-to-point matching requires computing Euclidean distances to all stored points within candidate voxels to identify the closest correspondence, which can incur substantial computational overhead.
For this reason, we propose an additional pruning strategy, which skips candidate neighboring voxels that cannot yield a closer point than the current best one.

We first define the minimum distance from the query point to a candidate neighboring voxel, denoted by $d_\mathrm{nbr}$.
As illustrated in Fig.~\ref{fig:distance to candidate neighboring voxel}, $d_\mathrm{nbr}$ is determined by the sharing relation between the query point’s occupied region within the root voxel and the neighboring voxel.
Specifically, it corresponds to a point-to-surface, point-to-edge, or point-to-corner distance when the neighboring voxel shares a surface, edge, or corner with the occupied region, respectively.

% Algorithm 2
\begin{algorithm}[!t]
	\caption{Pruning strategy for point-to-point correspondence search}
	\label{alg: pruning strategy}
	\SetKwInOut{Input}{Input}
	\SetKwInOut{Output}{Output}

	\Input{Transformed query point $^W\mathbf{p}$; fixed root voxel size~$d_\mathrm{root}$; outlier rejection threshold~$\tau_\mathrm{closest}$.}
	\Output{Nearest point $^W\acute{\mathbf{p}}$ and its distance $d_\mathrm{closest}$.}

	$(\mathcal{V}_\mathrm{root}, \mathcal{V}_\mathrm{nbrs}) \!\gets\! \texttt{GetCandidateVoxels}(^W\mathbf{p}, d_\mathrm{root})$\;
	$d_\mathrm{closest} \gets \tau_\mathrm{closest}$\;
	$(^W\acute{\mathbf{p}}_\mathrm{temp},\, d_\mathrm{temp}) \gets \texttt{NN-Search}(\mathcal{V}_\mathrm{root}, {^W\mathbf{p}})$\;
	\If{$d_\mathrm{temp} < d_\mathrm{closest}$}{
		$(^W\acute{\mathbf{p}},\, d_\mathrm{closest}) \gets (^W\acute{\mathbf{p}}_\mathrm{temp},\, d_\mathrm{temp})$\;
	}

	\For{$\mathcal{V}_\mathrm{nbr} \in \mathcal{V}_\mathrm{nbrs}$}{
		$d_\mathrm{nbr} \!\gets\! \texttt{ComputeDistanceToVoxel}(^W\mathbf{p}, \mathcal{V}_\mathrm{nbr})$\;
		\If{$d_\mathrm{nbr} < d_\mathrm{closest}$}{
			$(^W\acute{\mathbf{p}}_\mathrm{temp},\, d_\mathrm{temp}) \gets \texttt{NN-Search}(\mathcal{V}_\mathrm{nbr}, {^W\mathbf{p}})$\;
			\If{$d_\mathrm{temp} < d_\mathrm{closest}$}{
				$(^W\acute{\mathbf{p}},\, d_\mathrm{closest}) \gets (^W\acute{\mathbf{p}}_\mathrm{temp},\, d_\mathrm{temp})$\;
			}
		}
	}

	\If{$d_\mathrm{closest} < \tau_\mathrm{closest}$}{
		\Return{$(^W\acute{\mathbf{p}},\, d_\mathrm{closest})$}\;
	}
	\Else{
		\Return{$\emptyset$}.
	}
\end{algorithm}

With this definition, the search first evaluates the root voxel and obtains the distance to the closest point found so far, denoted by $d_\mathrm{closest}$.
For each candidate neighboring voxel, a coarse-to-fine check is applied: $d_\mathrm{nbr}$ is first computed as the minimum distance from the query point to the voxel itself, without accessing any stored points.
Since no stored point inside that voxel can be closer than $d_\mathrm{nbr}$, the voxel is skipped if $d_\mathrm{nbr} \ge d_\mathrm{closest}$; otherwise, a point-level nearest-neighbor search is performed within that voxel.
Note that $d_\mathrm{closest}$ is initialized with an outlier rejection threshold $\tau_\mathrm{closest}$ to discard correspondences that are excessively distant from the query point.
The detailed pruning procedure is summarized in Algorithm~\ref{alg: pruning strategy}.

If a valid closest point is found, it is inserted into the point-to-point correspondence set $\mathcal{C}_\mathrm{po}$.
If neither a valid point-to-plane nor a valid point-to-point correspondence is found, the query point is regarded as a noise-induced outlier and excluded from further processing.
The above procedures are performed independently for each query point in the transformed~$\mathcal{V}_t$, resulting in two correspondence sets: the point-to-plane set~$\mathcal{C}_\mathrm{pl}$ and the point-to-point set~$\mathcal{C}_\mathrm{po}$.
The correspondences in~$\mathcal{C}_\mathrm{pl}$ are used to compute point-to-plane residuals, Jacobians, and covariances in Sec.~\ref{subsec:Setting Point-to-Plane Residual, Jacobian, and Covariance},
while those in~$\mathcal{C}_\mathrm{po}$ are used to compute point-to-point residuals, Jacobians, and covariances in Sec.~\ref{subsec:Setting Point-to-Point Residual, Jacobian, and Covariance}.

\subsection{SETTING POINT-TO-PLANE RESIDUAL, JACOBIAN, AND COVARIANCE}
\label{subsec:Setting Point-to-Plane Residual, Jacobian, and Covariance} 
For each correspondence in the point-to-plane set $\mathcal{C}_\mathrm{pl}$, the point-to-plane error metric~\cite{rusinkiewicz2001IntConfThreeDDigitalImagingAndModeling} is applied.
The normal ${^{W}\mathbf{n}}$, center~${^{W}\mathbf{q}}$, and query point ${^{L}\mathbf{p}}$ are accompanied by noise terms ${\bm{\delta}_{^{W}\mathbf{n}}}$, ${\bm{\delta}_{^{W}\mathbf{q}}}$, and ${\bm{\delta}_{^{L}\mathbf{p}}}$, respectively.
Considering these noise terms jointly, the point-to-plane measurement model $\mathbf{h}_\mathrm{pl}(\mathbf{x},\mathbf{v}_\mathrm{pl})$ can be expressed as follows:
\begin{equation} \label{eq: point-to-plane measurement model}
\begin{aligned}
0&=\mathbf{h}_\mathrm{pl}(\mathbf{x},\mathbf{v}_\mathrm{pl})\\
&\triangleq ({^{W}\mathbf{n}}\boxplus {\bm{\delta}_{^{W}\mathbf{n}})}^\intercal({^{W}\mathbf{T}_{I}}{^{I}\mathbf{T}_{L}}({^{L}\mathbf{p}}+{\bm{\delta}_{^{L}\mathbf{p}}})-{^{W}\mathbf{q}}-{\bm{\delta}_{^{W}\mathbf{q}}}),
\end{aligned}
\end{equation}
where the point-to-plane noise vector $\mathbf{v}_\mathrm{pl}$ is defined as follows:
\begin{equation}
{\mathbf{v}_\mathrm{pl}}=
\begin{bmatrix}
{\bm{\delta}_{^{W}\mathbf{n}}}^\intercal,
{\bm{\delta}_{^{W}\mathbf{q}}}^\intercal,
{\bm{\delta}_{^{L}\mathbf{p}}}^\intercal
\end{bmatrix}^\intercal\in \mathbb{R}^{9\times1}.
\end{equation}

Note that the state estimate at the $\ell$-th iteration is denoted by $\widehat{\mathbf{x}}^\ell$, where $\widehat{\mathbf{x}}^0 = \widehat{\mathbf{x}}$.
To compute the residual, Jacobian, and covariance, the point-to-plane measurement model~(\ref{eq: point-to-plane measurement model}) is linearized at $\widehat{\mathbf{x}}^\ell$ using a first-order Taylor expansion, as follows:
\begin{equation} \label{eq: linearized point-to-plane measurement model for state update}
0 \simeq
z_\mathrm{pl}^\ell
+\mathbf{H}_{\widetilde{\mathbf{x}},\mathrm{pl}}^\ell \widetilde{\mathbf{x}}^\ell
+\mathbf{H}_{\mathbf{v},\mathrm{pl}}^\ell\mathbf{v}_\mathrm{pl},
\end{equation}
where $z_\mathrm{pl}^\ell$ $(=\mathbf{h}_\mathrm{pl}(\widehat{\mathbf{x}}^{\ell}, \mathbf{0}))$ is the point-to-plane residual, $\widetilde{\mathbf{x}}^\ell=\mathbf{x}\boxminus\widehat{\mathbf{x}}^\ell$, $\mathbf{H}_{\mathbf{v},\mathrm{pl}}^\ell\mathbf{v}_\mathrm{pl}\sim\mathcal{N}(0,R_\mathrm{pl}^\ell)$ is the lumped point-to-plane measurement noise, $\mathbf{H}_{\widetilde{\mathbf{x}},\mathrm{pl}}^\ell$ and $\mathbf{H}_{\mathbf{v},\mathrm{pl}}^\ell$ are the Jacobian matrices of $\mathbf{h}_\mathrm{pl}(\widehat{\mathbf{x}}^\ell\boxplus\widetilde{\mathbf{x}}^\ell,\mathbf{v}_\mathrm{pl})$ with respect to $\widetilde{\mathbf{x}}^\ell$ and $\mathbf{v}_\mathrm{pl}$, evaluated at zero, respectively.
$z_\mathrm{pl}^\ell \in \mathbb{R}$, $\mathbf{H}_{\widetilde{\mathbf{x}},\mathrm{pl}}^\ell \in \mathbb{R}^{1\times18}$ and $\mathbf{H}_{\mathbf{v},\mathrm{pl}}^\ell \in \mathbb{R}^{1\times9}$ are expressed as follows:
\begin{equation}
\begin{aligned}
&z_\mathrm{pl}^\ell = {^{W}\mathbf{n}}^\intercal({^{W}\mathbf{T}_I}^\ell{^{I}\mathbf{T}_L}{^{L}\mathbf{p}}-{^{W}\mathbf{q}}),\\
&\mathbf{H}_{\widetilde{\mathbf{x}},\mathrm{pl}}^\ell=
\begin{bmatrix}
{^{W}\mathbf{n}}^\intercal,
-{^{W}\mathbf{n}}^\intercal{^{W}\mathbf{R}_{I}}^\ell{^{I}\mathbf{R}_{L}}{[^{L}\mathbf{p}]_\times},
\mathbf{0}_{1 \times 12}
\end{bmatrix},\\
&\mathbf{H}_{\mathbf{v},\mathrm{pl}}^\ell=
\begin{bmatrix}
({^{W}\mathbf{T}_{I}}^\ell{^{I}\mathbf{T}_{L}}{^{L}\mathbf{p}}-{^{W}\mathbf{q}})^\intercal,
-{^{W}\mathbf{n}}^\intercal,
{^{W}\mathbf{n}}^\intercal{^{W}\mathbf{R}_I}^\ell{^{I}\mathbf{R}_L}
\end{bmatrix},
\end{aligned}
\end{equation}
where $[\cdot]_\times$ is an operator that converts the 3D vector into a skew-symmetric matrix.

Thus, (\ref{eq: linearized point-to-plane measurement model for state update}) can be reformulated as follows:
\begin{equation} \label{eq: point-to-plane measurement model distribution}
z_\mathrm{pl}^\ell
+\mathbf{H}_{\widetilde{\mathbf{x}},\mathrm{pl}}^\ell \widetilde{\mathbf{x}}^\ell
\simeq-\mathbf{H}_{\mathbf{v},\mathrm{pl}}^\ell \mathbf{v}_\mathrm{pl}
\sim \mathcal{N}(0,R_\mathrm{pl}^\ell),
\end{equation}
where the point-to-plane covariance $R_\mathrm{pl}^\ell$ is defined as
\begin{equation} \label{eq:R_pl}
R_\mathrm{pl}^\ell=
\mathbf{H}_{\mathbf{v},\mathrm{pl}}^\ell
\,
\mathbf{\Sigma}_{{^{W}\mathbf{n}},{^{W}\mathbf{q}},{^{L}\mathbf{p}}}
\,
{\mathbf{H}_{\mathbf{v},\mathrm{pl}}^\ell}^\intercal
\in \mathbb{R}.
\end{equation}
Here, the noise covariance $\mathbf{\Sigma}_{{^{W}\mathbf{n}},{^{W}\mathbf{q}},{^{L}\mathbf{p}}}$ is given by
\begin{equation}
\mathbf{\Sigma}_{{^{W}\mathbf{n}},{^{W}\mathbf{q}},{^{L}\mathbf{p}}}=
\begin{bmatrix}
\mathbf{\Sigma}_{{^{W}\mathbf{n}},{^{W}\mathbf{q}}} & \mathbf{0}_{6\times3} \\
\mathbf{0}_{3\times6} & \mathbf{\Sigma}_{^{L}\mathbf{p}}
\end{bmatrix}\in\mathbb{R}^{9\times9},
\end{equation}
where $\mathbf{\Sigma}_{{^{W}\mathbf{n}},{^{W}\mathbf{q}}}$ denotes the covariance of the normal ${^{W}\mathbf{n}}$ and center ${^{W}\mathbf{q}}$, and $\mathbf{\Sigma}_{^{L}\mathbf{p}}$ denotes the covariance of the query point~${^{L}\mathbf{p}}$ in the LiDAR frame. Detailed derivations of these terms are provided in Yuan~\etal~\cite{yuan2022ral}.

This procedure is applied to all elements of $\mathcal{C}_\mathrm{pl}$ (i.e., point-to-plane correspondences) to obtain the point-to-plane residuals $z_\mathrm{pl}^\ell$, Jacobians $\mathbf{H}_{\widetilde{\mathbf{x}},\mathrm{pl}}^\ell$, and covariances $R_\mathrm{pl}^\ell$ for each correspondence.
These terms are then used in the hybrid-metric state update described in Sec.~\ref{subsec:hybrid-metric state update}.

\subsection{SETTING POINT-TO-POINT RESIDUAL, JACOBIAN, AND COVARIANCE}
\label{subsec:Setting Point-to-Point Residual, Jacobian, and Covariance}
Next, for each correspondence in the point-to-point set $\mathcal{C}_\mathrm{po}$, the point-to-point error metric~\cite{besl1992tpami} is applied.
As with~(\ref{eq: point-to-plane measurement model}), the point-to-point measurement model $\mathbf{h}_\mathrm{po}(\mathbf{x},\mathbf{v}_\mathrm{po})$ incorporating the noise terms is given by:
\begin{equation} \label{eq: point-to-point measurement model}
\mathbf{0}=\mathbf{h}_\mathrm{po}(\mathbf{x},\mathbf{v}_\mathrm{po}) \triangleq {^{W}\mathbf{T}_{I}}{^{I}\mathbf{T}_{L}}({^{L}\mathbf{p}}+{\bm{\delta}_{^{L}\mathbf{p}}})-{^{W}\acute{\mathbf{p}}}-{\bm{\delta}_{^{W}\acute{\mathbf{p}}}},
\end{equation}
where the point-to-point noise vector $\mathbf{v}_\mathrm{po}$ is defined as follows:
\begin{equation}
{\mathbf{v}_\mathrm{po}}=
\begin{bmatrix}
{\bm{\delta}_{^{L}\mathbf{p}}}^\intercal,
{\bm{\delta}_{^{W}\acute{\mathbf{p}}}}^\intercal
\end{bmatrix}^\intercal\in \mathbb{R}^{6\times1},
\end{equation}
where $\bm{\delta}_{^{W}\acute{\mathbf{p}}}$ is the noise term of the closest target point $^{W}\acute{\mathbf{p}}$.

In this work, the point-to-point residual is reformulated as an $L_2$ norm scalar residual, which measures the Euclidean distance between corresponding points, rather than using a full vector residual~\cite{besl1992tpami}. 
This choice reduces the dimensionality of the Jacobians and covariances, leading to lower computational overhead during state updates. 
It also provides a scalar residual that is consistent with the point-to-plane formulation, enabling balanced residual fusion within the hybrid-metric state update.
While vector residuals may provide richer geometric constraints in structured environments, such cases are adequately addressed by the point-to-plane residuals~\cite{rusinkiewicz2001IntConfThreeDDigitalImagingAndModeling} in our system.

Accordingly, the point-to-point $L_2$ norm measurement model $\mathbf{h}_\mathrm{po}^\mathrm{norm}(\mathbf{x},\mathbf{v}_\mathrm{po})$ can be defined as follows:
\begin{equation} \label{eq: point-to-point L2 norm measurement model}
\begin{aligned}
0&=\mathbf{h}_\mathrm{po}^\mathrm{norm}(\mathbf{x},\mathbf{v}_\mathrm{po}) \triangleq \lVert \mathbf{h}_\mathrm{po}(\mathbf{x},\mathbf{v}_\mathrm{po}) \rVert \\ &=\lVert{^{W}\mathbf{T}_{I}}{^{I}\mathbf{T}_{L}}({^{L}\mathbf{p}}+{\bm{\delta}_{^{L}\mathbf{p}}})-{^{W}\acute{\mathbf{p}}}-{\bm{\delta}_{^{W}\acute{\mathbf{p}}}}\rVert.
\end{aligned}
\end{equation}

To compute the residual, Jacobian, and covariance, (\ref{eq: point-to-point L2 norm measurement model}) is linearized at $\widehat{\mathbf{x}}^\ell$ using the chain rule and a first-order Taylor expansion, as follows:
\begin{equation} \label{eq: linearized point-to-point L2 norm measurement model for state update}
0 \simeq
\lVert \mathbf{z}_\mathrm{po}^\ell \rVert
+\frac{{\mathbf{z}_\mathrm{po}^\ell}^\intercal}{\lVert \mathbf{z}_\mathrm{po}^\ell \rVert} \mathbf{H}_{\widetilde{\mathbf{x}},\mathrm{po}}^\ell \widetilde{\mathbf{x}}^\ell
+\frac{{\mathbf{z}_\mathrm{po}^\ell}^\intercal}{\lVert \mathbf{z}_\mathrm{po}^\ell \rVert} \mathbf{H}_{\mathbf{v},\mathrm{po}}^\ell \mathbf{v}_\mathrm{po},
\end{equation}
where ${\mathbf{z}}_\mathrm{po}^\ell$ $(=\mathbf{h}_\mathrm{po}(\widehat{\mathbf{x}}^{\ell}, \mathbf{0}))$ is the point-to-point residual, $\mathbf{H}_{\mathbf{v},\mathrm{po}}^\ell \mathbf{v}_\mathrm{po} \sim \mathcal{N}(\mathbf{0},\mathbf{R}_\mathrm{po}^\ell)$ is the lumped point-to-point measurement noise, $\mathbf{H}_{\widetilde{\mathbf{x}},\mathrm{po}}^\ell$ and $\mathbf{H}_{\mathbf{v},\mathrm{po}}^\ell$ are the Jacobian matrices of $\mathbf{h}_\mathrm{po}(\widehat{\mathbf{x}}^\ell\boxplus\widetilde{\mathbf{x}}^\ell,\mathbf{v}_\mathrm{po})$ with respect to $\widetilde{\mathbf{x}}^\ell$ and $\mathbf{v}_\mathrm{po}$, evaluated at zero, respectively.
${\mathbf{z}}_\mathrm{po}^\ell$, $\mathbf{H}_{\widetilde{\mathbf{x}},\mathrm{po}}^\ell$ and $\mathbf{H}_{\mathbf{v},\mathrm{po}}^\ell$ are expressed as
\begin{equation}
\begin{aligned}
&{\mathbf{z}}_\mathrm{po}^\ell = {^{W}\mathbf{T}_I}^\ell{^{I}\mathbf{T}_L}{^{L}\mathbf{p}}-{^{W}\acute{\mathbf{p}}}\in \mathbb{R}^{3 \times 1},\\
&\mathbf{H}_{\widetilde{\mathbf{x}},\mathrm{po}}^\ell=
\begin{bmatrix}
\mathbf{I}_3,
-{^{W}\mathbf{R}_{I}}^\ell{^{I}\mathbf{R}_{L}}{[^{L}\mathbf{p}]_\times},
\mathbf{0}_{3 \times 12}
\end{bmatrix} \in \mathbb{R}^{3\times18},\\
&\mathbf{H}_{\mathbf{v},\mathrm{po}}^\ell=
\begin{bmatrix}
{^{W}\mathbf{R}_I}^\ell{^{I}\mathbf{R}_L},-\mathbf{I}_3
\end{bmatrix} \in \mathbb{R}^{3\times6}.
\end{aligned}
\end{equation}

The normalization by $\lVert \mathbf{z}_\mathrm{po}^\ell \rVert$ in (\ref{eq: linearized point-to-point L2 norm measurement model for state update}) yields the unit direction of the point-to-point residual.
Although this term is well defined for nonzero residuals, it becomes undefined at zero residual and can be numerically unstable when the residual norm is very small.
Therefore, in implementation, when $\lVert \mathbf{z}_\mathrm{po}^\ell \rVert < \epsilon_\mathrm{po}$, where $\epsilon_\mathrm{po}$ is a user-defined near-zero threshold, we omit the corresponding scalar point-to-point residual term from the update.
Such near-zero residuals indicate that the matched points are already nearly aligned; hence, they provide negligible corrective information, while their unit direction may impose an arbitrary numerical constraint.

For point-to-point residuals satisfying $\lVert \mathbf{z}_\mathrm{po}^\ell \rVert \ge \epsilon_\mathrm{po}$, (\ref{eq: linearized point-to-point L2 norm measurement model for state update})~can be written as follows:
\begin{equation} \label{eq: point-to-point L2 norm measurement model distribution}
z_\mathrm{po}^{\mathrm{norm},\ell}
+\mathbf{H}_{\widetilde{\mathbf{x}},\mathrm{po}}^{\mathrm{norm},\ell} \widetilde{\mathbf{x}}^\ell
\simeq -\frac{{\mathbf{z}_\mathrm{po}^\ell}^\intercal}{\lVert \mathbf{z}_\mathrm{po}^\ell \rVert} \mathbf{H}_{\mathbf{v},\mathrm{po}}^\ell \mathbf{v}_\mathrm{po}
\sim \mathcal{N}(0,R_\mathrm{po}^{\mathrm{norm},\ell}),
\end{equation}
where $z_\mathrm{po}^{\mathrm{norm},\ell}=\lVert \mathbf{z}_\mathrm{po}^{\ell} \rVert$, and the Jacobian $\mathbf{H}_{\widetilde{\mathbf{x}},\mathrm{po}}^{\mathrm{norm},\ell}$ and covariance $R_\mathrm{po}^{\mathrm{norm},\ell}$ are defined as
\begin{equation} \label{eq:R_po}
\begin{aligned}
&\mathbf{H}_{\widetilde{\mathbf{x}},\mathrm{po}}^{\mathrm{norm},\ell}=\frac{{\mathbf{z}_\mathrm{po}^\ell}^\intercal}{\lVert \mathbf{z}_\mathrm{po}^\ell \rVert} \mathbf{H}_{\widetilde{\mathbf{x}},\mathrm{po}}^\ell \in \mathbb{R}^{1\times18},\\
&R_\mathrm{po}^{\mathrm{norm},\ell}=\frac{{\mathbf{z}_\mathrm{po}^\ell}^\intercal}{\lVert \mathbf{z}_\mathrm{po}^\ell \rVert} \mathbf{R}_{\mathrm{po}}^\ell \frac{\mathbf{z}_\mathrm{po}^\ell}{\lVert \mathbf{z}_\mathrm{po}^\ell \rVert}\in \mathbb{R}.
\end{aligned}
\end{equation}
Here, $\mathbf{R}_\mathrm{po}^{\ell}$ is given by
\begin{equation}
\mathbf{R}_\mathrm{po}^\ell=
\mathbf{H}_{\mathbf{v},\mathrm{po}}^\ell\,
\mathbf{\Sigma}_{{^{L}\mathbf{p}},{^{W}\acute{\mathbf{p}}}}\,
{\mathbf{H}_{\mathbf{v},\mathrm{po}}^\ell}^\intercal
\in \mathbb{R}^{3\times3},
\end{equation}
where the noise covariance $\mathbf{\Sigma}_{{^{L}\mathbf{p}},{^{W}\acute{\mathbf{p}}}}$ is defined as follows:
\begin{equation}
\mathbf{\Sigma}_{{^{L}\mathbf{p}},{^{W}\acute{\mathbf{p}}}}=
\begin{bmatrix}
\mathbf{\Sigma}_{^{L}\mathbf{p}} & \mathbf{0}_{3 \times 3} \\
\mathbf{0}_{3 \times 3} & \mathbf{\Sigma}_{^{W}\acute{\mathbf{p}}}
\end{bmatrix}\in\mathbb{R}^{6\times6},
\end{equation}
where $\mathbf{\Sigma}_{^{W}\acute{\mathbf{p}}}$ denotes the covariance of the closest target point ${^{W}\acute{\mathbf{p}}} \in \mathcal{P}$ (see Yuan~\etalcite{yuan2022ral} for derivations).

In addition to the uncertainty-based point-to-point $L_2$ norm covariance $R_\mathrm{po}^{\mathrm{norm},\ell}$, we also account for the discretization error of point-to-point correspondences.
As described in Sec.~\ref{subsec: voxel-pruned correspondence search}, a set of past observation points $\mathcal{P} = \left\{^{W}\acute{\mathbf{p}}_j \right\}_{j=1}^{N_\mathrm{stored}}$ is stored in each root voxel of the map.
While point-to-plane correspondences are matched to planes estimated from these accumulated points, point-to-point correspondences are matched directly to one of the stored past observation points, making them more susceptible to discretization errors~\cite{wu2024icra}.

Accordingly, inspired by the map discretization error of Wu~\etal~\cite{wu2024icra}, we model the variance term of the discretization error using quantities available from the voxel-pruned correspondence search described in Sec.~\ref{subsec: voxel-pruned correspondence search}, as follows:
\begin{equation} \label{eq: discretization error}
R_\mathrm{disc} = \frac{N_\mathrm{vox}^\mathrm{acc} \cdot{d_\mathrm{root}}^2}{N_\mathrm{pt}^\mathrm{eval}},
\end{equation}
where $d_\mathrm{root}$ is the fixed root voxel size, and $N_\mathrm{vox}^\mathrm{acc}$ and $N_\mathrm{pt}^\mathrm{eval}$ denote the numbers of candidate voxels accessed and points evaluated during the voxel-pruned point-to-point correspondence search for the query point $^{W}\mathbf{p}$, respectively.
Here, $N_\mathrm{vox}^\mathrm{acc} \cdot {d_\mathrm{root}}^2$ approximates the squared spatial extent of the accessed candidate voxel region, while $N_\mathrm{pt}^\mathrm{eval}$ normalizes this extent by the number of points actually evaluated during the search.
Thus, the discretization variance $R_\mathrm{disc}$ becomes larger when the search region is broader and the evaluated points are more sparsely distributed.

The discretization variance $R_\mathrm{disc}$ is then incorporated with the uncertainty-based covariance $R_\mathrm{po}^{\mathrm{norm},\ell}$ to obtain the combined point-to-point covariance $R_\mathrm{po}^{\mathrm{comb},\ell}$ as follows:
\begin{equation} \label{eq: final point-to-point covariance}
R_\mathrm{po}^{\mathrm{comb},\ell}=\lambda_\mathrm{po}\,(R_\mathrm{po}^{\mathrm{norm},\ell}+R_\mathrm{disc}) \in \mathbb{R},
\end{equation}
where $\lambda_\mathrm{po}$ is a user-defined covariance scaling factor to account for the different numerical scales of the point-to-plane and point-to-point residual covariances.
The need for this scaling arises because $R_\mathrm{pl}^{\ell}$ in (\ref{eq:R_pl}) is estimated from a local plane constructed using tens to hundreds of points~\cite{yuan2022ral}, whereas $R_\mathrm{po}^{\mathrm{norm},\ell}$ in (\ref{eq:R_po}) is derived from a single query-target point correspondence.
These different geometric supports can lead to different raw covariance scales in the stacked state update.
Without weight scaling, point-to-plane residuals tend to dominate the state update, causing point-to-point $L_2$ norm residuals to be underweighted.

For instance, in open or unstructured environments where point-to-plane correspondences are sparse, point-to-point $L_2$ norm residuals can play a crucial role in providing geometric constraints; see Sec.~\ref{subsec: exp_hybrid_metric}.
When their weights are excessively low, however, they cannot contribute effectively to the state update, leading to degraded estimation performance.
Thus, by introducing the scaling factor $\lambda_\mathrm{po}$, the proposed formulation enables balanced residual fusion, leading to more consistent state estimation across diverse environments.
A detailed analysis of this combined point-to-point covariance design, including a sensitivity analysis of $\lambda_\mathrm{po}$ and an ablation study of $R_\mathrm{disc}$ across confined and open environments, is provided in Sec.~\ref{sec:ptpo_covariance_design}.

This process is applied to all elements of $\mathcal{C}_\mathrm{po}$ (i.e., point-to-point correspondences), yielding the point-to-point $L_2$ norm residuals $z_\mathrm{po}^{\mathrm{norm},\ell}$, Jacobians $\mathbf{H}_{\widetilde{\mathbf{x}},\mathrm{po}}^{\mathrm{norm},\ell}$, and combined point-to-point covariances $R_\mathrm{po}^{\mathrm{comb},\ell}$ for each correspondence.
These terms are then utilized in the hybrid-metric state update described in Sec.~\ref{subsec:hybrid-metric state update}.

\subsection{HYBRID-METRIC STATE UPDATE}
\label{subsec:hybrid-metric state update}
By combining the prior distribution in (\ref{eq: prior distribution}) with the point-to-plane and point-to-point $L_2$ norm measurement models in~(\ref{eq: point-to-plane measurement model distribution}) and~(\ref{eq: point-to-point L2 norm measurement model distribution}), respectively, we obtain the posterior distribution of the state $\mathbf{x}$.
This posterior can be equivalently represented in terms of the error-state $\widetilde{\mathbf{x}}^\ell$, whose maximum a posteriori (MAP) estimate is given by:
\begin{equation} \label{eq: hybrid-metric state update MAP}
\min_{\widetilde{\mathbf{x}}^\ell} \left(\lVert \mathbf{x} \boxminus \widehat{\mathbf{x}} \rVert^2_{\widehat{\mathbf{P}}}
+ \sum_{i=1}^{N} \lVert \mathbf{z}_{i}^\ell + \mathbf{H}_{\widetilde{\mathbf{x}},i}^\ell \widetilde{\mathbf{x}}^\ell \rVert^2_{\mathbf{R}_{i}^\ell} 
\right),
\end{equation}
where $\lVert \mathbf{a} \rVert_{\mathbf{M}}^2 \triangleq \mathbf{a}^\intercal \mathbf{M}^{-1} \mathbf{a}$, and $\sum\limits_{i=1}^{N} \lVert \mathbf{z}_{i}^\ell + \mathbf{H}_{\widetilde{\mathbf{x}},i}^\ell \widetilde{\mathbf{x}}^\ell \rVert^2_{\mathbf{R}_{i}^\ell}$ can be expressed as follows:
\begin{equation}
\begin{aligned}
\sum_{i=1}^{N} \lVert \mathbf{z}_{i}^\ell + \mathbf{H}_{\widetilde{\mathbf{x}},i}^\ell \widetilde{\mathbf{x}}^\ell \rVert^2_{\mathbf{R}_{i}^\ell} 
&=\sum_{j=1}^{N_\mathrm{pl}} \lVert z_{\mathrm{pl},j}^\ell + \mathbf{H}_{\widetilde{\mathbf{x}},\mathrm{pl},j}^\ell \widetilde{\mathbf{x}}^\ell \rVert^2_{R_{\mathrm{pl},j}^\ell}\\
&+\sum_{k=1}^{N_\mathrm{po}} \lVert z_{\mathrm{po},k}^{\mathrm{norm},\ell} + \mathbf{H}_{\widetilde{\mathbf{x}},\mathrm{po},k}^{\mathrm{norm},\ell} \widetilde{\mathbf{x}}^\ell \rVert^2_{R_{\mathrm{po},k}^{\mathrm{comb},\ell}},
\end{aligned}
\end{equation}
where $N_\mathrm{pl}$ and $N_\mathrm{po}$ denote the number of point-to-plane and point-to-point correspondences, respectively, with $N = N_\mathrm{pl} + N_\mathrm{po}$.
Furthermore, the overall residual vector $\mathbf{z}^\ell$, Jacobian matrix $\mathbf{H}_{\widetilde{\mathbf{x}}}^\ell$, and covariance matrix $\mathbf{R}^\ell$ are constructed by stacking the terms from point-to-plane and point-to-point correspondences.
That is, $z_\mathrm{pl}^\ell$ and $z_\mathrm{po}^{\mathrm{norm},\ell}$ form the residual~$\mathbf{z}^\ell$,
$\mathbf{H}_{\widetilde{\mathbf{x}},\mathrm{pl}}^\ell$ and $\mathbf{H}_{\widetilde{\mathbf{x}},\mathrm{po}}^{\mathrm{norm},\ell}$ form the Jacobian matrix $\mathbf{H}_{\widetilde{\mathbf{x}}}^\ell$,
and $R_{\mathrm{pl}}^\ell$ and $R_{\mathrm{po}}^{\mathrm{comb},\ell}$ constitute the block-diagonal covariance matrix~$\mathbf{R}^\ell$.

% Algorithm 3
\begin{algorithm}[!t]
	\caption{Hybrid-metric state update in ESIKF}
	\label{alg: hybrid-metric state update}
	\SetKwInOut{Input}{Input}
	\SetKwInOut{Output}{Output}
	
	\Input{Propagated state $\widehat{\mathbf{x}}$; propagated covariance $\widehat{\mathbf{P}}$; voxelized scan $\mathcal{V}_t$; voxel map $\mathcal{G}$; convergence threshold $\tau_\mathrm{converge}$.}
	\Output{Updated state $\bar{\mathbf{x}}$; updated covariance $\bar{\mathbf{P}}$.}
	
	$\widehat{\mathbf{x}}^{0} \gets \widehat{\mathbf{x}}$; \quad $\ell \gets -1$;
	
	\Repeat{$\|\widehat{\mathbf{x}}^{\ell+1} \boxminus \widehat{\mathbf{x}}^{\ell}\| < \tau_\mathrm{converge}$}{
		$\ell \gets \ell + 1$;

		$\mathcal{C}_\mathrm{pl},\, \mathcal{C}_\mathrm{po} \gets \texttt{GetCorrespondences}(\mathcal{V}_t, \mathcal{G}, \widehat{\mathbf{x}}^\ell)$;

		$\{z_{\mathrm{pl},j}^\ell,\, \mathbf{H}_{\widetilde{\mathbf{x}},\mathrm{pl},j}^\ell,\, R_{\mathrm{pl},j}^\ell\}_{j=1}^{N_\mathrm{pl}} 
		\gets \texttt{SetPointToPlaneTerms}(\mathcal{C}_\mathrm{pl}, \widehat{\mathbf{x}}^\ell)$;

		$\{z_{\mathrm{po},k}^{\mathrm{norm},\ell},\, \mathbf{H}_{\widetilde{\mathbf{x}},\mathrm{po},k}^{\mathrm{norm},\ell},\, R_{\mathrm{po},k}^{\mathrm{comb},\ell}\}_{k=1}^{N_\mathrm{po}} 
		\gets \texttt{SetPointToPointTerms}(\mathcal{C}_\mathrm{po}, \widehat{\mathbf{x}}^\ell)$;

		Construct overall residual $\mathbf{z}^\ell$, Jacobian $\mathbf{H}_{\widetilde{\mathbf{x}}}^\ell$, and covariance $\mathbf{R}^\ell$;
		
		Compute prior covariance $\mathbf{P}^\ell = (\mathbf{J}^\ell)^{-1} \, \widehat{\mathbf{P}} \, (\mathbf{J}^\ell)^{-\intercal}$;
		
		$\mathbf{K}^\ell \!\gets\! \left({\mathbf{H}_{\widetilde{\mathbf{x}}}^\ell}^\intercal (\mathbf{R}^\ell)^{-1} \mathbf{H}_{\widetilde{\mathbf{x}}}^\ell + (\mathbf{P}^\ell)^{-1} \right)^{-1} {\mathbf{H}_{\widetilde{\mathbf{x}}}^\ell}^\intercal (\mathbf{R}^\ell)^{-1}$;
		
		$\widehat{\mathbf{x}}^{\ell+1} \!\gets\! \widehat{\mathbf{x}}^\ell \boxplus \left(\!-\mathbf{K}^\ell \mathbf{z}^\ell \!\!-\!(\mathbf{I} \!-\! \mathbf{K}^\ell \mathbf{H}_{\widetilde{\mathbf{x}}}^\ell)(\mathbf{J}^\ell)^{\!-1}\!(\widehat{\mathbf{x}}^\ell \!\boxminus\! \widehat{\mathbf{x}}) \right)$\!;
	}
	
	$\bar{\mathbf{x}} \gets \widehat{\mathbf{x}}^{\ell+1}$;\quad $\bar{\mathbf{P}} \gets (\mathbf{I} - \mathbf{K}^\ell \mathbf{H}_{\widetilde{\mathbf{x}}}^\ell) \, \mathbf{P}^\ell$;
	
	\Return{$\bar{\mathbf{x}}, \bar{\mathbf{P}}$}.
	\vspace{-0.1cm}
\end{algorithm}

This MAP problem can be solved by an error-state iterated Kalman filter~\cite{xu2022tro} as follows:
\begin{equation} \label{eq: error-state iterated Kalman filter}
\begin{aligned}
&\mathbf{K}^\ell=({\mathbf{H}_{\widetilde{\mathbf{x}}}^\ell}^\intercal (\mathbf{R}^\ell)^{-1} {\mathbf{H}_{\widetilde{\mathbf{x}}}^\ell}+(\mathbf{P}^\ell)^{-1})^{-1}
{\mathbf{H}_{\widetilde{\mathbf{x}}}^\ell}^\intercal (\mathbf{R}^\ell)^{-1},\\
&\widehat{\mathbf{x}}^{\ell+1}=\widehat{\mathbf{x}}^\ell \boxplus (-\mathbf{K}^\ell \mathbf{z}^\ell - (\mathbf{I}-\mathbf{K}^\ell \mathbf{H}_{\widetilde{\mathbf{x}}}^\ell)(\mathbf{J}^\ell)^{-1}(\widehat{\mathbf{x}}^\ell \boxminus \widehat{\mathbf{x}})),
\end{aligned}
\end{equation}
where $\mathbf{z}^\ell\in\mathbb{R}^{N\times1}$, $\mathbf{H}_{\widetilde{\mathbf{x}}}^\ell\in\mathbb{R}^{N\times18}$, $\mathbf{R}^\ell\in\mathbb{R}^{N\times N}$, and $\mathbf{P}^\ell\in\mathbb{R}^{18\times18}$ are defined as follows:
\begin{equation}
\label{eq: stacked lidar residual jacobian covariance}
\begin{aligned}
&\mathbf{z}^\ell=
\begin{bmatrix}
{z_{\mathrm{pl},1}^\ell},\dots,{z_{\mathrm{pl},N_\mathrm{pl}}^\ell}, {z_{\mathrm{po},1}^{\mathrm{norm},\ell}},\dots,{z_{\mathrm{po},N_\mathrm{po}}^{\mathrm{norm},\ell}}
\end{bmatrix}^\intercal,\\
&\mathbf{H}_{\widetilde{\mathbf{x}}}^\ell=
\begin{bmatrix}
{\mathbf{H}_{\widetilde{\mathbf{x}},\mathrm{pl},1}^\ell}^\intercal,...,{\mathbf{H}_{\widetilde{\mathbf{x}},\mathrm{pl},N_\mathrm{pl}}^\ell}^\intercal, {\mathbf{H}_{\widetilde{\mathbf{x}},\mathrm{po},1}^{\mathrm{norm},\ell}}^\intercal,...,{\mathbf{H}_{\widetilde{\mathbf{x}},\mathrm{po},N_\mathrm{po}}^{\mathrm{norm},\ell}}^\intercal
\end{bmatrix}^\intercal,\\
&\mathbf{R}^\ell = \mathrm{diag}(
R_{\mathrm{pl},1}^\ell, \dots, R_{\mathrm{pl},N_\mathrm{pl}}^\ell,\;
R_{\mathrm{po},1}^{\mathrm{comb},\ell}, \dots, R_{\mathrm{po},N_\mathrm{po}}^{\mathrm{comb},\ell}),\\
&\mathbf{P}^\ell=(\mathbf{J}^\ell)^{-1}\widehat{\mathbf{P}}(\mathbf{J}^\ell)^{-\intercal},
\end{aligned}
\end{equation}
where $\mathbf{J}^\ell$ is the partial derivative of $(\widehat{\mathbf{x}}^\ell \boxplus \widetilde{\mathbf{x}}^\ell)\boxminus \widehat{\mathbf{x}}$ with respect to $\widetilde{\mathbf{x}}^\ell$ evaluated at zero~\cite{xu2022tro}.
Note that computing the Kalman gain~$\mathbf{K}^\ell$ requires inverting a matrix of the state dimension instead of the measurement dimension~\cite{xu2021ral}.

The previous process in Sec.~\ref{sec:hybrid-metric state update} repeats until convergence (i.e., $\lVert \widehat{\mathbf{x}}^{\ell+1} \boxminus \widehat{\mathbf{x}}^\ell \rVert < \tau_\mathrm{converge}$).
After convergence, the optimal state and covariance estimates are updated as follows:
\begin{equation}
\bar{\mathbf{x}}=\widehat{\mathbf{x}}^{\ell+1},\ 
\bar{\mathbf{P}}=(\mathbf{I}-\mathbf{K}^\ell \mathbf{H}_{\widetilde{\mathbf{x}}}^\ell)\mathbf{P}^\ell.
\end{equation}

With the updated state $\bar{\mathbf{x}}$, each LiDAR point $^{L}\mathbf{p}_i$ in $\mathcal{V}_{\mathrm{merge},t}$, which is obtained in Sec.~\ref{subsec: bi-resolution voxelization}, is transformed to the world frame as follows:
\begin{equation}
{^{W}\bar{\mathbf{p}}_i}={^{W}\bar{\mathbf{T}}_I}{^{I}{\mathbf{T}}_L}{^{L}\mathbf{p}_i}
;\,i=1,\dots,N_{\mathrm{merge},t},
\end{equation}
where $N_{\mathrm{merge},t}$ is the number of points in $\mathcal{V}_{\mathrm{merge},t}$.
Then, the transformed LiDAR points are merged into the voxel map~\cite{yuan2022ral}.
Our hybrid-metric state update is summarized in Algorithm~\ref{alg: hybrid-metric state update}.

%%%%%%%%%%%%%%%%%%%%%%%%%%%%%%%%%%%%%%%%%%%%%%%%%%%%%%%%%%%%%%%%%%%%%%%%%%%%%%%%

\section{EXPERIMENTAL EVALUATION}
\label{sec:experimental evaluation}
The main focus of this work is to develop a LIO framework that remains robust and computationally efficient across a wide range of spatial scales, including confined spaces, open areas, and transitions between them.
We present a series of experiments to demonstrate the capabilities of our method and to support our key claims.
%We present our experiments to show the capabilities of our method and to support the following key claims.
%First, our approach consistently achieves competitive odometry estimation performance across indoor, outdoor, and transitional environments.
%Second, it remains robust and computationally efficient when transitioning between environments with substantially different spatial scales.
%Third, our sensitivity-informed gain scheduling strategy mitigates oscillations during voxel size control and enables faster, more stable tracking of the scale-informed setpoint for voxelized point count.
%Fourth, our voxel-pruned correspondence search algorithm significantly reduces processing time by pruning redundant neighboring voxel traversal.
%Fifth, our scale-aware adaptive voxelization is not restricted to a specific LIO algorithm but can be easily applied to a wide range of LiDAR-based odometry systems, thereby improving robustness and computational efficiency in scale-varying environments.

\subsection{EXPERIMENTAL SETUP}
\subsubsection{PUBLIC DATASETS}
To evaluate the performance of GenZ-LIO across a wide range of spatial scales, we conduct experiments on nine public datasets: \dataset{SubT-MRS}~\cite{zhao2024cvpr}, \dataset{SuperLoc}~\cite{zhao2025icra}, \dataset{2021 HILTI}~\cite{helmberger2022ral}, \dataset{2022 HILTI}~\cite{zhang2022ral}, \dataset{GEODE}~\cite{chen2026ijrr}, \dataset{M3DGR}~\cite{zhang2025arxiv}, \dataset{NTU-VIRAL}~\cite{nguyen2022ijrr}, \dataset{ENWIDE}~\cite{pfreundschuh2024icra}, and \dataset{Oxford Spires}~\cite{tao2025ijrr}.

The selected sequences, their scene characteristics, and the associated LiDAR sensors are described as follows.
From each dataset, we select sequences that exhibit challenging spatial-scale conditions, including extremely confined spaces, wide open areas, and transitions between them.
For \dataset{SubT-MRS}~\cite{zhao2024cvpr}, we use \dataset{Long Corridor} (degenerate corridor), \dataset{Laurel Cavern} (unstructured cave), and \dataset{Multi Floor} (from confined stairs to open spaces), all recorded with a Velodyne VLP-16.
For \dataset{SuperLoc}~\cite{zhao2025icra}, we use \dataset{Cave\,01}, \dataset{Cave\,02}, and \dataset{Cave\,04} (unstructured caves), together with \dataset{Corridor\,02} (from corridor to open space), all under the VLP-16 setup.
For \dataset{2021 HILTI}~\cite{helmberger2022ral}, we use \dataset{Basement\,04} and \dataset{Drone Arena}, which represent confined environments and are recorded with a Livox MID-70.
For \dataset{2022 HILTI}~\cite{zhang2022ral}, we use \dataset{Exp\,10}, \dataset{Exp\,16}, and \dataset{Exp\,18}, which include confined staircases and open hall segments, recorded with a Hesai PandarXT-32.
For \dataset{GEODE}~\cite{chen2026ijrr}, we use \dataset{Stairs} (confined staircase), \dataset{Waterways-Short}, \dataset{Waterways-Medium}, and \dataset{Waterways-Long} (open waterways), as well as \dataset{Offroad-02}, \dataset{Offroad-04}, and \dataset{Offroad-07} (open off-road spaces), all recorded with a VLP-16.
For \dataset{M3DGR}~\cite{zhang2025arxiv}, we use \dataset{Corridor\,01} and \dataset{Corridor\,02} (degenerate corridors), and \dataset{GNSS-denial\,01} and \dataset{GNSS-denial\,02} (open outdoor areas), recorded with a Livox AVIA.
For \dataset{NTU-VIRAL}~\cite{nguyen2022ijrr}, we use \dataset{SPMS\,01}, \dataset{SPMS\,02}, and \dataset{SPMS\,03}, captured in high-altitude open environments with agile drone motion, using an Ouster OS1-16.
For \dataset{ENWIDE}~\cite{pfreundschuh2024icra}, we use \dataset{Katzensee\,S} and \dataset{Katzensee\,D} (walking and running in unstructured open scenes), and \dataset{Intersection\,S} and \dataset{Intersection\,D} (walking and running in structured open spaces), all recorded with an Ouster OS0-128.
For \dataset{Oxford Spires}~\cite{tao2025ijrr}, we use \dataset{christ-church-01}, \dataset{christ-church-02}, and \dataset{christ-church-05}, as well as \dataset{blenheim-palace-01}, \dataset{blenheim-palace-02}, and \dataset{blenheim-palace-05}, which feature confined--open transitions and are recorded with a Hesai QT64.

\subsubsection{NARROWWIDE DATASET}
\begin{figure*}[t!]
	\centering
	
	\begin{subfigure}[t]{0.272\textwidth}
		\centering
		\includegraphics[width=\linewidth]{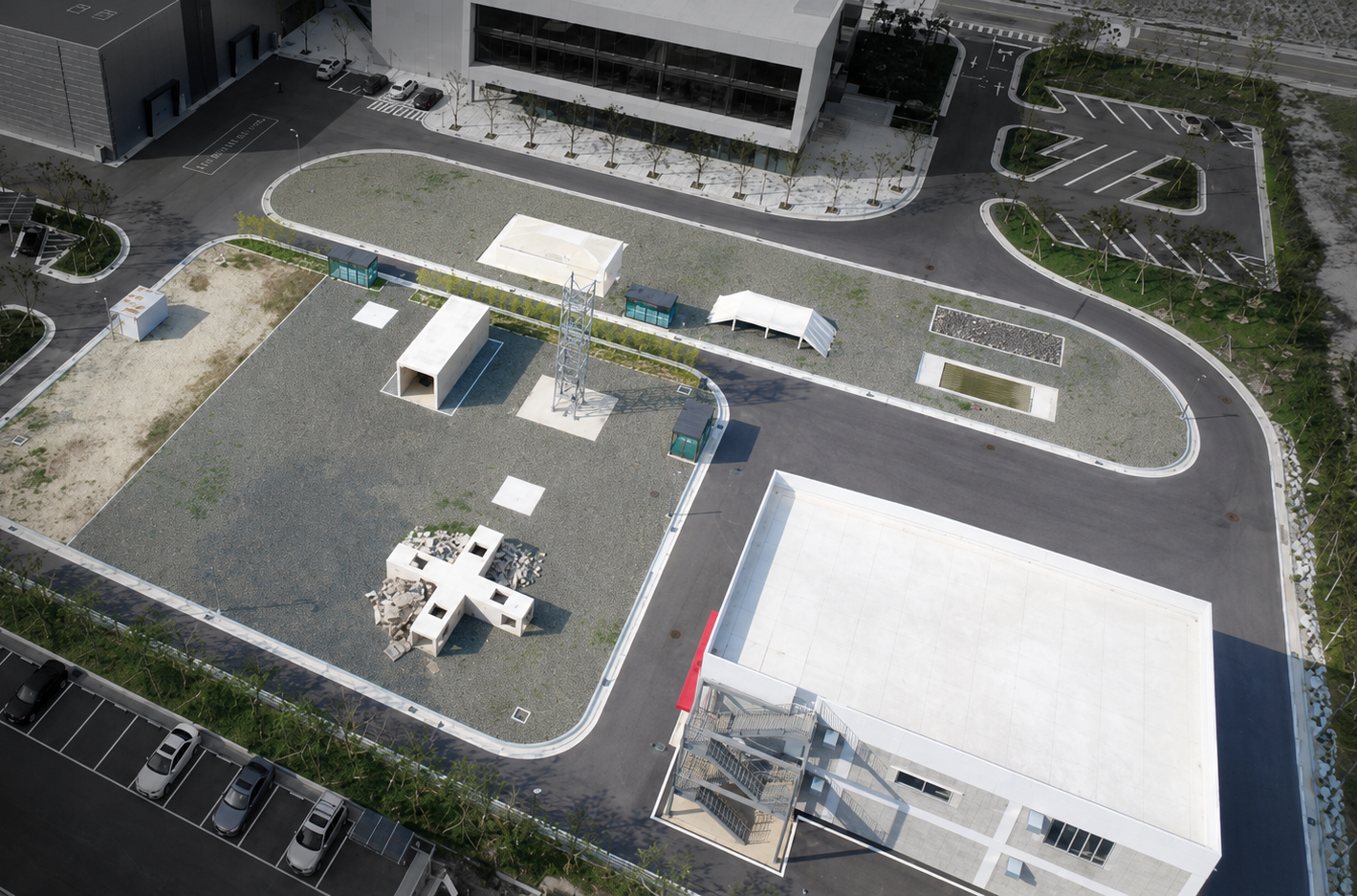}
		\caption{}
		\label{fig:narrowwide_overview}
	\end{subfigure}
	\hfill
	\begin{subfigure}[t]{0.72\textwidth}
		\centering
		\includegraphics[width=\linewidth]{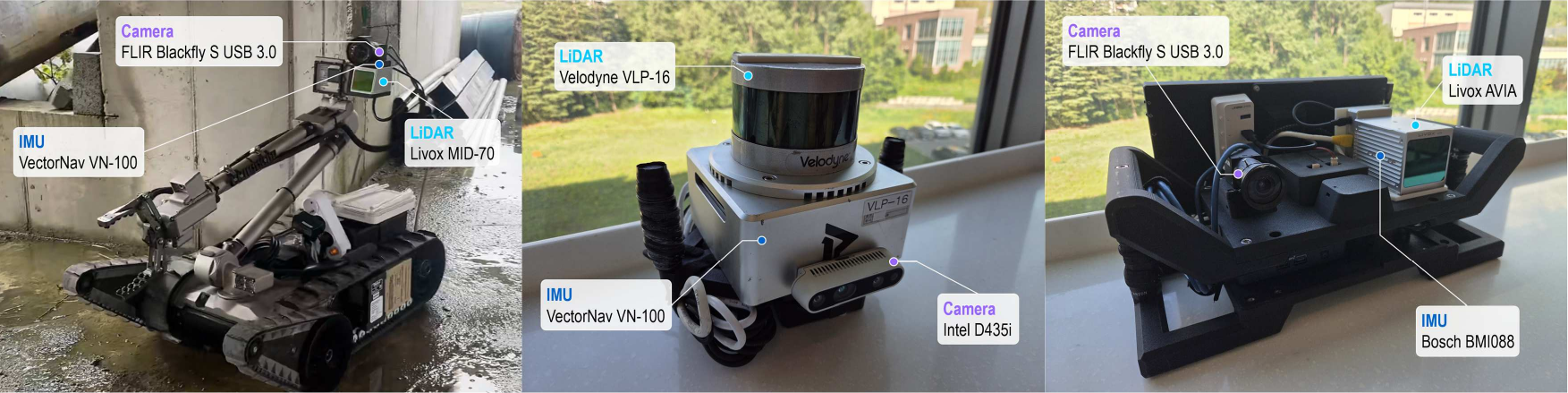}
		\caption{}
		\label{fig:narrowwide_platforms}
	\end{subfigure}
	
	\vspace{-0.2cm}
	\begin{subfigure}[t]{1.0\textwidth}
		\centering
		\includegraphics[width=\linewidth]{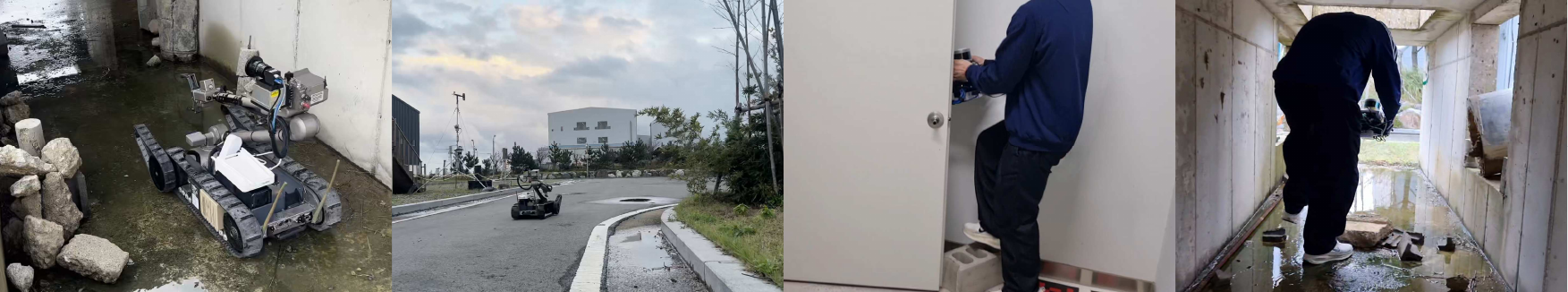}
		\caption{}
		\label{fig:narrowwide_field_experiments}
	\end{subfigure}
	\vspace{-0.3cm}
	
	\caption{
		Overview of the \dataset{NarrowWide} dataset.
		(a)~Data acquisition environment designed to evaluate robotic systems for disaster-response scenarios.
		(b)~Platforms used for data collection: a tracked robot and two handheld devices.
		These platforms employ different LiDAR sensors to cover diverse sensing configurations, as detailed in Table~\ref{tab:narrowwide_platform}.
		(c)~Field experiments across different spatial scales using a tracked robot and handheld sensor platforms.
		The handheld platforms complement the tracked robot experiments by covering confined spaces that are difficult for the tracked robot to access due to its physical size.
		Details and traversal routes of the individual sequences are provided in Table~\ref{tab:narrowwide_sequence} and Fig.~\ref{fig:narrowwide_mapping}, respectively.
	}
	\label{fig:narrow_wide_dataset}
	\vspace{-0.6cm}
\end{figure*}

To further evaluate performance under confined--open transition scenarios with more extreme variations in spatial scale, we acquire a new dataset named \dataset{NarrowWide}.
Although several existing datasets include confined--open transitions, they often provide limited opportunities to evaluate repeated transitions across substantially different spatial scales. To complement these benchmarks, \dataset{NarrowWide} is collected to capture frequent transitions between open areas and confined structures with varying spatial scales, as illustrated in Fig.~\ref{fig:narrow_wide_dataset}.

\begin{figure*}[t!]
	\captionsetup{}
	\captionsetup[subfigure]{skip=1pt}
	\centering
	\begin{subfigure}[t]{0.325\textwidth}
		\centering
		\includegraphics[width=\linewidth]{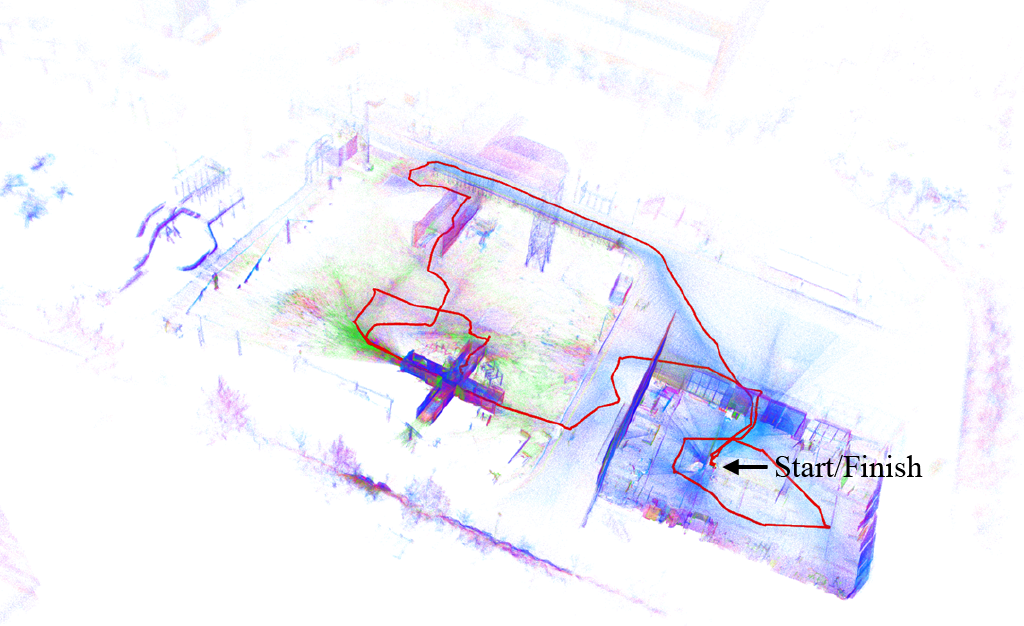}
		\caption{\dataset{Tracked-01}}
	\end{subfigure}
	\hfill
	\begin{subfigure}[t]{0.325\textwidth}
		\centering
		\includegraphics[width=\linewidth]{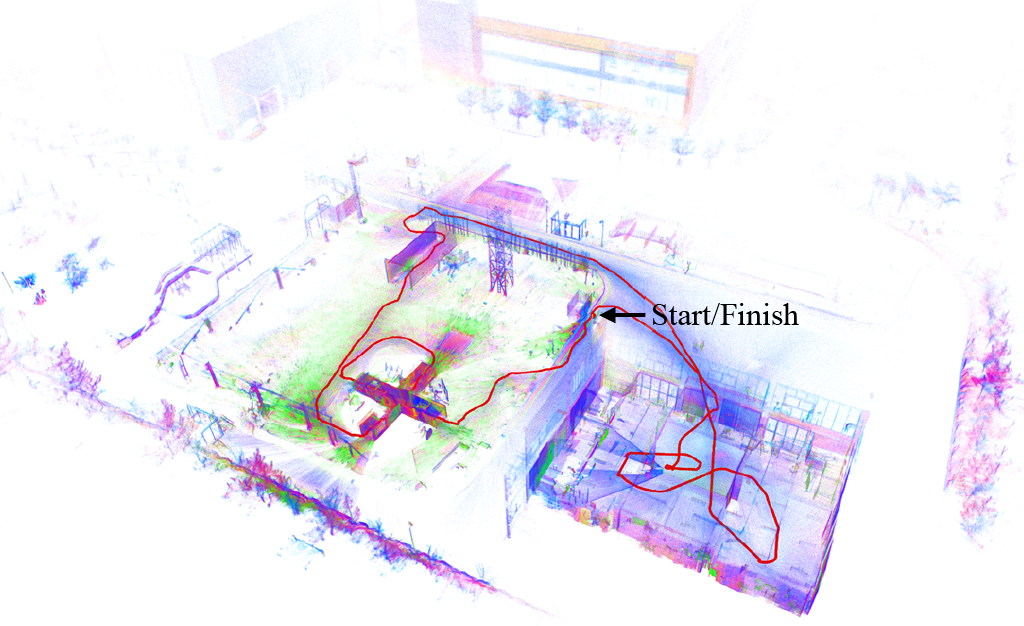}
		\caption{\dataset{Tracked-02}}
	\end{subfigure}
	\hfill
	\begin{subfigure}[t]{0.325\textwidth}
		\centering
		\includegraphics[width=\linewidth]{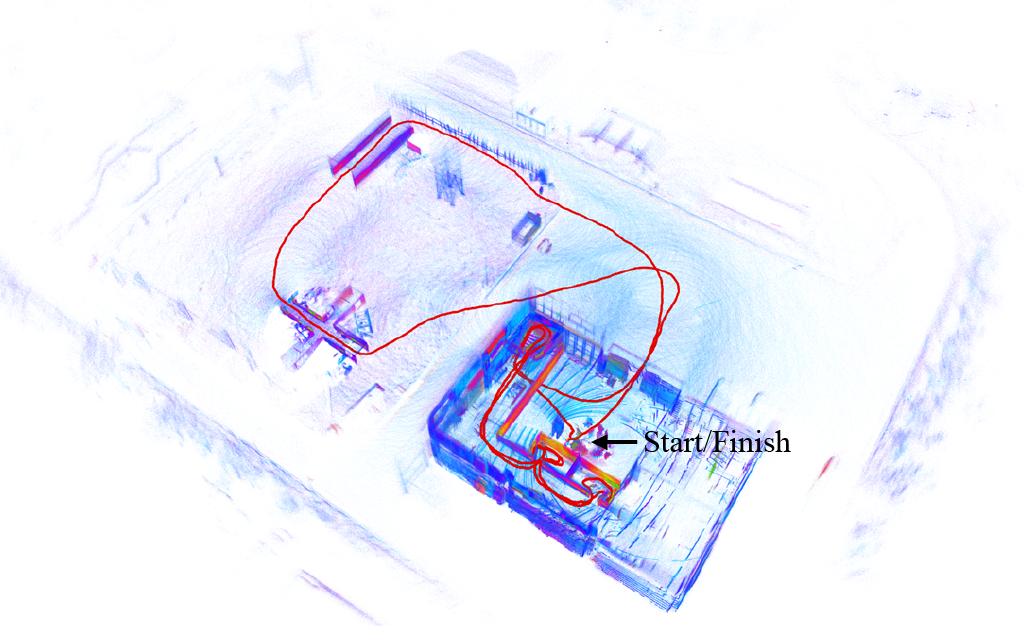}
		\caption{\dataset{Handheld-A-01}}
	\end{subfigure}

	\vspace{-0.2cm}
	\begin{subfigure}[t]{0.325\textwidth}
		\centering
		\includegraphics[width=\linewidth]{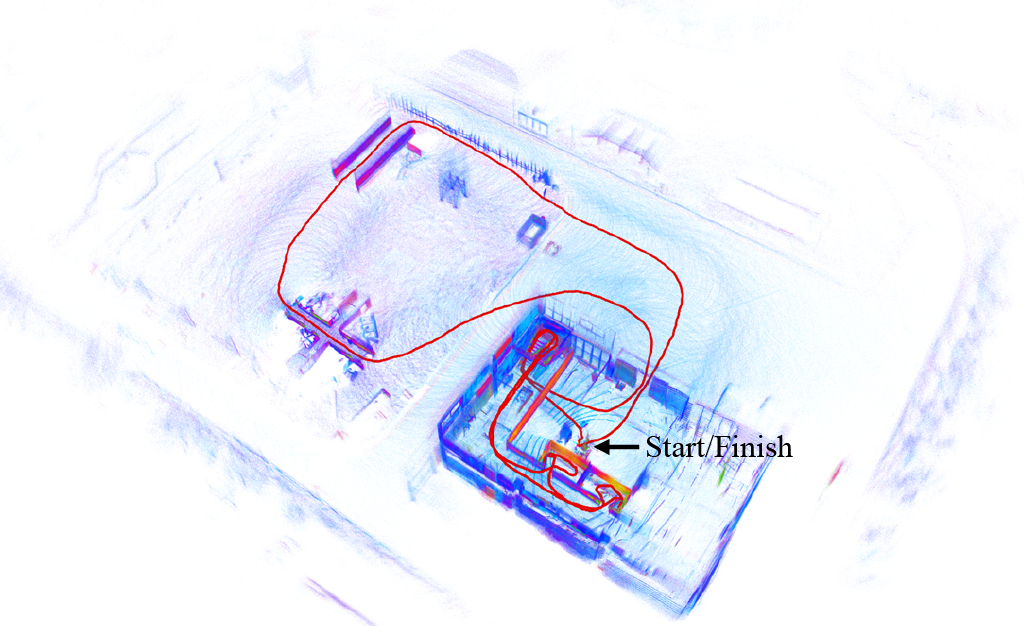}
		\caption{\dataset{Handheld-A-02}}
	\end{subfigure}
	\hfill
	\begin{subfigure}[t]{0.325\textwidth}
		\centering
		\includegraphics[width=\linewidth]{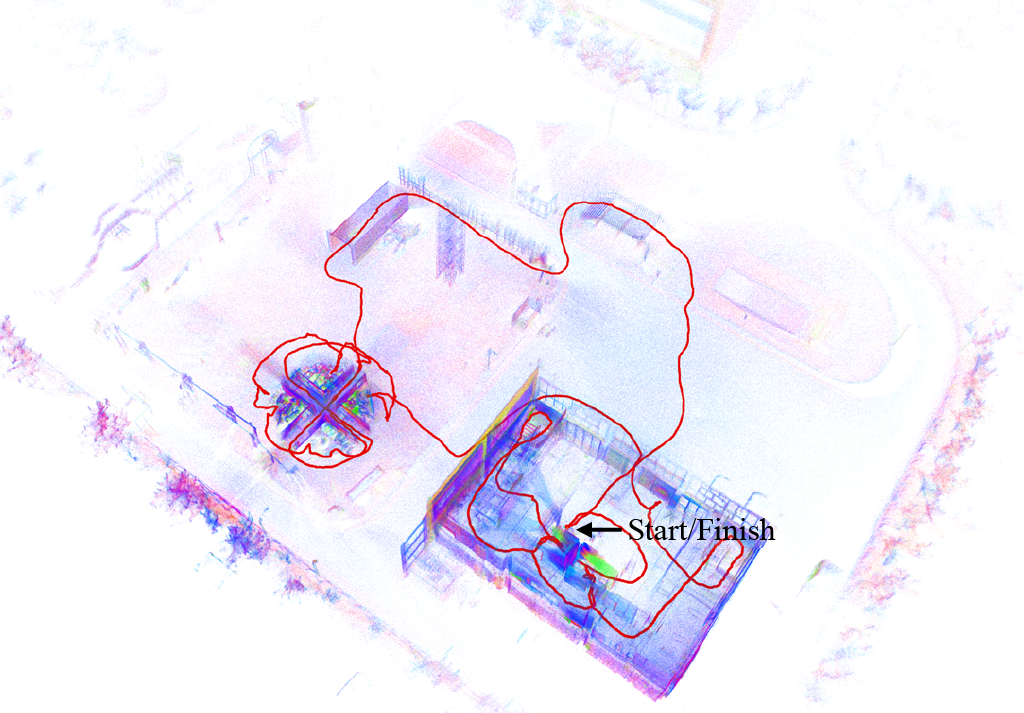}
		\caption{\dataset{Handheld-B-01}}
	\end{subfigure}
	\hfill
	\begin{subfigure}[t]{0.325\textwidth}
		\centering
		\includegraphics[width=\linewidth]{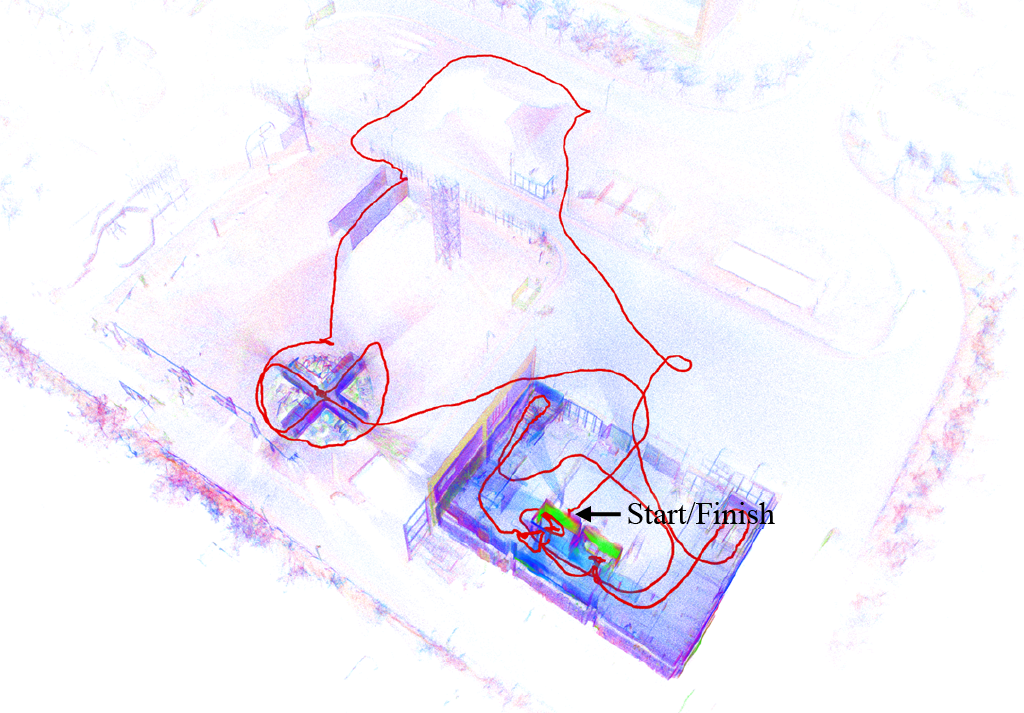}
		\caption{\dataset{Handheld-B-02}}
	\end{subfigure}
	\vspace{-0.1cm}
	\caption{Mapping results and estimated trajectories of GenZ-LIO on the six sequences of the \dataset{NarrowWide} dataset.}
	\label{fig:narrowwide_mapping}
	\vspace{-0.6cm}
\end{figure*}

As shown in Fig.~\ref{fig:narrow_wide_dataset}(b), the \dataset{NarrowWide} dataset is collected using three platforms to cover a wide range of spatial scales, including extremely confined spaces.
A tracked robot platform (Teledyne FLIR PackBot~510), equipped with a Livox MID-70 LiDAR and a VectorNav VN-100 IMU, is used to acquire the \dataset{Tracked-01} and \dataset{Tracked-02} sequences.
In addition, four handheld sequences are recorded to complement the tracked sequences, covering not only portions of the same environment but also more confined spaces that are difficult for the tracked robot to access due to its size.
Specifically, \dataset{Handheld-A-01} and \dataset{Handheld-A-02} are collected using a Velodyne VLP-16 LiDAR and a VectorNav VN-100 IMU, while \dataset{Handheld-B-01} and \dataset{Handheld-B-02} are collected using a Livox AVIA LiDAR with its built-in BMI088 IMU.
For all platforms, a camera is mounted solely for visualization and is not utilized by the odometry pipeline.
The sensor configuration of each platform is summarized in Table~\ref{tab:narrowwide_platform}.
In addition, details and traversal routes for each sequence in the \dataset{NarrowWide} dataset are provided in Table~\ref{tab:narrowwide_sequence} and Fig.~\ref{fig:narrowwide_mapping}, respectively.

For ground truth (GT) generation, we adopt a map-based trajectory estimation procedure, motivated by the GT system used in the \dataset{Newer College}~\cite{ramezani2020iros} dataset.
Specifically, a high-resolution prior map of the environment is first constructed using a survey-grade 3D imaging laser scanner (Leica BLK360).
Globally consistent 6-DoF GT trajectories are then obtained by adapting PALoc~\cite{hu2024tmech}, which localizes each sequence against the prior map.
For trajectory segments where the map-based localization became unstable, additional scan-to-map refinement is performed to improve trajectory consistency.
The \dataset{NarrowWide} dataset is available~at \textcolor{magenta}{\href{https://github.com/cocel-postech/NarrowWide}{https://github.com/cocel-postech/NarrowWide}}.

\begin{table}[!t]
	\captionsetup{}
	\caption{Sensor configurations of the three platforms used to collect the \dataset{NarrowWide} dataset. The platforms correspond to those shown in Fig.~\ref{fig:narrow_wide_dataset}(b).}
	\label{tab:narrowwide_platform}
	\centering
	\setlength{\tabcolsep}{3.0pt}
	\vspace{-0.2cm}
	{
		\fontsize{7.5}{9}\selectfont
		\begin{tabular*}{\columnwidth}{@{\extracolsep{\fill}}lccc@{}}
			\toprule[0.8pt]
			Platform & Tracked robot & Handheld A & Handheld B \\
			\midrule
			LiDAR & Livox MID-70 & Velodyne VLP-16 & Livox AVIA \\
			\;\;\;\;FoV (H\,$\times$\,V) & $70.4^\circ$ circular & $360^\circ \times 30^\circ$ & $70.4^\circ \times 77.2^\circ$ \\
			\;\;\;\;Max. range [m] & 260 & 100 & 450 \\
			\;\;\;\;Frequency [Hz] & 10 & 10 & 10 \\
			\midrule
			IMU & VectorNav VN-100 & VectorNav VN-100 & Bosch BMI088 \\
			\;\;\;\;Frequency [Hz] & 200 & 200 & 200 \\
			\midrule
			Camera & FLIR Blackfly S & Intel D435i & FLIR Blackfly S \\
			\;\;\;\;Frequency [Hz] & 30 & 30 & 30 \\
			\bottomrule[0.8pt]
	\end{tabular*}}
\end{table}

\begin{table}[!t]
	\captionsetup{}
	\caption{Characteristics of each sequence in the \dataset{NarrowWide} dataset. The minimum width denotes the width of the narrowest space traversed in each sequence, whereas the widest open areas traversed exceed $100\,\mathrm{m}$ in all sequences. Note that every sequence returns to its starting position, forming a loop.}
	\label{tab:narrowwide_sequence}
	\centering
	\setlength{\tabcolsep}{3.0pt}
	\renewcommand{\arraystretch}{0.9}
	\vspace{-0.18cm}
	{
		\fontsize{7.5}{9}\selectfont
		\begin{tabular*}{\columnwidth}{@{\extracolsep{\fill}}lccccc@{}}
			\toprule[0.8pt]
			\makecell{Sequence} & \makecell{Platform} & \makecell{Dist.\\{[m]}} & \makecell{Duration\\{[s]}} & \makecell{Min. width\\{[m]}} & \makecell{\# of confined--\\open transitions} \\
			\midrule
			\dataset{Tracked-01} & Tracked robot & 265.7 & 621.0 & 1.0 & 8 \\
			\dataset{Tracked-02} & Tracked robot & 269.2 & 578.2 & 1.0 & 8 \\
			\dataset{Handheld-A-01} & Handheld A & 263.0 & 521.9 & 0.3 & 6 \\
			\dataset{Handheld-A-02} & Handheld A & 244.9 & 402.4 & 0.3 & 6 \\
			\dataset{Handheld-B-01} & Handheld B & 408.7 & 636.0 & 0.5 & 10 \\
			\dataset{Handheld-B-02} & Handheld B & 415.9 & 529.1 & 0.5 & 8 \\
			\bottomrule[0.8pt]
	\end{tabular*}}
\end{table}

\subsubsection{SYSTEM CONFIGURATIONS}
All experiments are conducted on a desktop PC equipped with an Intel i7-13700K CPU and $32\,\mathrm{GB}$ RAM.
The default configuration of GenZ-LIO is as follows.
The sliding window size is set to $N_w=5$.
The target number of voxelized points is bounded by $[N_{\min}, N_{\max}] = [1,\!000, 4,\!000]$.
The exponent in the scale-to-setpoint mapping is fixed to $p = 2$.
The spatial-scale threshold is set to $\tau_m = 30.0\,\mathrm{m}$, and the scaling factors used for normalizing the tracking error and its derivative are fixed to $\lambda_p = 0.1$ and $\lambda_d = 0.2$, respectively.
The voxel size is constrained within $[d_{\min}, d_{\max}] = [0.02, 1.0]\,\mathrm{m}$.
The proportional and derivative gains are bounded by $[K_{p,\min}, K_{p,\max}] = [5 \times 10^{-6}, 5 \times 10^{-5}]$ and $[K_{d,\min}, K_{d,\max}] = [5 \times 10^{-8}, 5 \times 10^{-7}]$, respectively.
The root voxel size is fixed to $d_{\mathrm{root}} = 0.5\,\mathrm{m}$.
The numerical tolerance for omitting near-zero scalar point-to-point residual terms is set to $\epsilon_\mathrm{po}=1\times10^{-6}\,\mathrm{m}$.
For hybrid-metric residual fusion, the scaling factor is set to $\lambda_\mathrm{po} = 0.05$.
Sensitivity analysis of the principal parameters across confined, open, and confined--open conditions is provided in Sec.~\ref{sec:parameter_sensitivity_analysis}.

\subsection{BENCHMARK EXPERIMENTS}\label{subsec: exp_benchmark}
% claim
The first experiment evaluates the odometry accuracy and divergence rate of the proposed method on 42 sequences from ten datasets, including the nine public datasets described above and the proposed \dataset{NarrowWide} dataset.
This experiment supports our first claim that the proposed system consistently achieves competitive odometry estimation performance across confined spaces, open areas, and transitions between them.

% setup
In this experiment, we compare GenZ-LIO against SOTA LIO systems, including FAST-LIO2~\cite{xu2022tro}, Faster-LIO~\cite{bai2022ral}, AdaLIO~\cite{lim2023ur}, Point-LIO~\cite{he2023ais}, LIO-EKF~\cite{wu2024icra}, DLIO~\cite{chen2023icra}, iG-LIO~\cite{chen2024ral}, and PV-LIO~\cite{hviktortsoi2023pvlio}.
Notably, PV-LIO is a reimplementation of VoxelMap~\cite{yuan2022ral}, extending it into a LIO framework~\cite{xu2022tro}, and serves as the baseline for GenZ-LIO.

In addition, two ablated variants are evaluated to isolate the effects of the proposed modules.
The first variant incorporates the proposed scale-aware adaptive voxelization into the baseline while retaining only the point-to-plane error metric for residual computation during the state update (referred to as \textit{Baseline w/ adap. vox.} in Table~\ref{tab:benchmark}).
The second variant applies only the proposed hybrid-metric state update on the baseline, without using the scale-aware adaptive voxelization (referred to as \textit{Baseline w/ hybrid-metric} in Table~\ref{tab:benchmark}).
For all compared methods, the voxel size is set to $0.25\,\mathrm{m}$ for confined scenes and $0.4\,\mathrm{m}$ for open scenes.
For confined--open transition sequences, $0.25\,\mathrm{m}$ is used to avoid excessive point reduction in confined segments.
This protocol prevents the compared methods from being disadvantaged by the overly coarse downsampling that may result from directly applying their default voxel sizes in confined regions.
The initial voxel size $d_0$ of GenZ-LIO follows the same criterion.
Sensor- and platform-dependent preprocessing parameters are configured per sequence and applied identically across all methods, whereas algorithm-specific parameters are kept at the default values provided by the corresponding open-source implementations.
As for the evaluation metric, we follow the common reporting practice in SOTA odometry systems~\cite{xu2022tro,bai2022ral,chen2023icra,chen2024ral,zheng2025tro} and report only translational errors to keep the extensive 42-sequence comparison compact.

\begin{table*}[p]
	\caption{Absolute translational errors (ATE), reported as RMSE in meters, for each sequence. Due to space limitations, dataset names are abbreviated as follows: \dataset{SubT-MRS}~\cite{zhao2024cvpr}~(\dataset{SM}), \dataset{SuperLoc}~\cite{zhao2025icra} (\dataset{SL}), \dataset{2021 HILTI}~\cite{helmberger2022ral} (\dataset{H'21}), \dataset{2022 HILTI}~\cite{zhang2022ral} (\dataset{H'22}), \dataset{GEODE}~\cite{chen2026ijrr} (\dataset{GD}), \dataset{M3DGR}~\cite{zhang2025arxiv} (\dataset{M3D}), \dataset{NTU-VIRAL}~\cite{nguyen2022ijrr} (\dataset{NV}), \dataset{ENWIDE}~\cite{pfreundschuh2024icra}~(\dataset{EW}), and \dataset{Oxford Spires}~\cite{tao2025ijrr} (\dataset{OS}). The best result is shown in \textbf{bold}. Note that ``$\times$'' denotes a divergent run, defined as an ATE RMSE exceeding $200\,\mathrm{m}$, whereas ``--'' indicates that the method cannot process the LiDAR sensor used in the sequence. The divergence rate is computed as the percentage of divergent runs among the sequences that the method can process. For each sequence, the method with the lowest ATE is ranked first, and the resulting ranks are averaged across sequences. Divergent cases receive the worst rank, and cases marked with ``--'' are excluded.}
	\centering
	\scriptsize
	\renewcommand\arraystretch{1.17}
	\setlength{\tabcolsep}{3.7pt}
	\begin{threeparttable}
		\begin{tabular}{c| l l| *{11}{c}}
			\toprule[0.8pt]
			Scale & \multicolumn{2}{c|}{Dataset / Sequence}
			& \makecell{FAST-LIO2\\~\cite{xu2022tro}}
			& \makecell{Faster-LIO\\~\cite{bai2022ral}}
			& \makecell{AdaLIO\\~\cite{lim2023ur}}
			& \makecell{Point-LIO\\~\cite{he2023ais}}
			& \makecell{LIO-EKF\\~\cite{wu2024icra}}
			& \makecell{DLIO\\~\cite{chen2023icra}}
			& \makecell{iG-LIO\\~\cite{chen2024ral}}
			& \makecell{PV-LIO~\cite{hviktortsoi2023pvlio}\\(Baseline)}
			& \makecell{Baseline w/\\adap. vox.}
			& \makecell{Baseline w/\\hybrid-metric}
			& \makecell{Ours} \\
			\midrule
			
			%%%%%%%%%%%%%%% Confined %%%%%%%%%%%%%%%%
			\multirow{19}{*}{\rotatebox{90}{Confined}}
			% ---- SubT-MRS (2 rows) ----
			& \multirow{2}{*}{\rotatebox{90}{\datasettable{SM}}}
			& \datasettable{Long Corridor}
			& \cellcolor{g3} 1.77 & \cellcolor{g9} 9.76 & \cellcolor{g2} 1.74 & \cellcolor{g11} 29.10 & \cellcolor{g10} 26.52 & \cellcolor{g8} 2.29 & \cellcolor{g1} \bf{1.55} & \cellcolor{g7} 2.01 & \cellcolor{g4} 1.90 & \cellcolor{g4} 1.90 & \cellcolor{g4} 1.90 \\
			& & \datasettable{Laurel Cavern}
			& \cellcolor{g7} 3.64 & \cellcolor{g8} 4.22 & \cellcolor{g9} 4.63 & \cellcolor{g10} 5.91 & \cellcolor{div} $\times$ & \cellcolor{g6} 0.58 & \cellcolor{g2} 0.36 & \cellcolor{g5} 0.42 & \cellcolor{g3} 0.37 & \cellcolor{g4} 0.39 & \cellcolor{g1} \bf{0.34} \\
			\cmidrule(lr){2-14}
			
			% ---- SuperLoc (3 rows) ----
			& \multirow{3}{*}{\rotatebox{90}{\datasettable{SL}}}
			& \datasettable{Cave\,01}
			& \cellcolor{div} $\times$ & \cellcolor{g8} 1.10 & \cellcolor{div} $\times$ & \cellcolor{g5} 0.18 & \cellcolor{div} $\times$ & \cellcolor{g7} 0.28 & \cellcolor{g1} \bf{0.12} & \cellcolor{g4} 0.15 & \cellcolor{g6} 0.25 & \cellcolor{g3} 0.13 & \cellcolor{g1} \bf{0.12} \\
			& & \datasettable{Cave\,02}
			& \cellcolor{g9} 4.49 & \cellcolor{g7} 0.81 & \cellcolor{g8} 3.98 & \cellcolor{g10} 6.57 & \cellcolor{div} $\times$ & \cellcolor{g6} 0.57 & \cellcolor{g2} 0.43 & \cellcolor{g2} 0.43 & \cellcolor{g4} 0.44 & \cellcolor{g1} \bf{0.42} & \cellcolor{g4} 0.44 \\
			& & \datasettable{Cave\,04}
			& \cellcolor{g7} 1.59 & \cellcolor{div} $\times$ & \cellcolor{g8} 2.89 & \cellcolor{g6} 0.52 & \cellcolor{div} $\times$ & \cellcolor{g9} 6.11 & \cellcolor{g1} \bf{0.17} & \cellcolor{g5} 0.26 & \cellcolor{g2} 0.21 & \cellcolor{g4} 0.22 & \cellcolor{g2} 0.21 \\
			\cmidrule(lr){2-14}
			
			% ---- HILTI'21 (2 rows) ----
			& \multirow{2}{*}{\rotatebox{90}{\datasettable{H'21}}}
			& \datasettable{Basement\,04}
			& \cellcolor{g9} 0.63 & \cellcolor{g5} 0.27 & \cellcolor{g7} 0.47 & \cellcolor{g8} 0.50 & \cellcolor{nosupport} -- & \cellcolor{div} $\times$ & \cellcolor{g5} 0.27 & \cellcolor{g4} 0.13 & \cellcolor{g1} \bf 0.05 & \cellcolor{g3} 0.06 & \cellcolor{g1} \bf 0.05 \\
			& & \datasettable{Drone Arena}
			& \cellcolor{g3} 0.19 & \cellcolor{g1} \bf 0.18 & \cellcolor{g8} 0.23 & \cellcolor{g4} 0.20 & \cellcolor{nosupport} -- & \cellcolor{div} $\times$ & \cellcolor{div} $\times$ & \cellcolor{g4} 0.20 & \cellcolor{g4} 0.20 & \cellcolor{g4} 0.20 & \cellcolor{g1} \bf 0.18 \\
			\cmidrule(lr){2-14}
			
			% ---- HILTI'22 (3 rows) ----
			& \multirow{3}{*}{\rotatebox{90}{\datasettable{H'22}}}
			& \datasettable{Exp\,10}
			& \cellcolor{div} $\times$ & \cellcolor{div} $\times$ & \cellcolor{div} $\times$ & \cellcolor{div} $\times$ & \cellcolor{div} $\times$ & \cellcolor{div} $\times$ & \cellcolor{g5} 1.77 & \cellcolor{g4} 0.61 & \cellcolor{g1} \bf 0.31 & \cellcolor{g3} 0.36 & \cellcolor{g2} 0.33 \\
			& & \datasettable{Exp\,16}
			& \cellcolor{div} $\times$ & \cellcolor{div} $\times$ & \cellcolor{g3} 0.74 & \cellcolor{g4} 0.95 & \cellcolor{div} $\times$ & \cellcolor{div} $\times$ & \cellcolor{div} $\times$ & \cellcolor{div} $\times$ & \cellcolor{g1} \bf 0.13 & \cellcolor{div} $\times$ & \cellcolor{g2} 0.20 \\
			& & \datasettable{Exp\,18}
			& \cellcolor{g5} 0.12 & \cellcolor{g9} 1.81 & \cellcolor{g6} 0.17 & \cellcolor{g8} 0.30 & \cellcolor{g10} 2.05 & \cellcolor{div} $\times$ & \cellcolor{g7} 0.20 & \cellcolor{g3} 0.08 & \cellcolor{g2} 0.06 & \cellcolor{g4} 0.11 & \cellcolor{g1} \bf 0.04 \\
			\cmidrule(lr){2-14}
			
			% ---- GEODE (1 row) ----
			& \multirow{1}{*}{\rotatebox{90}{\datasettable{GD}}}
			& \datasettable{Stairs}
			& \cellcolor{g6} 1.69 & \cellcolor{g7} 3.66 & \cellcolor{g5} 1.45 & \cellcolor{g8} 4.24 & \cellcolor{g11} 11.22 & \cellcolor{g9} 7.29 & \cellcolor{g2} 0.25 & \cellcolor{g10} 8.79 & \cellcolor{g4} 0.40 & \cellcolor{g3} 0.38 & \cellcolor{g1} \bf 0.24 \\
			\cmidrule(lr){2-14}
			
			% ---- M3DGR (2 rows) ----
			& \multirow{2}{*}{\rotatebox{90}{\datasettable{M3D}}}
			& \datasettable{Corridor\,01}
			& \cellcolor{g6} 5.84 & \cellcolor{g7} 6.26 & \cellcolor{g5} 5.24 & \cellcolor{g8} 19.63 & \cellcolor{nosupport} -- & \cellcolor{div} $\times$ & \cellcolor{div} $\times$ & \cellcolor{g4} 0.92 & \cellcolor{g2} 0.72 & \cellcolor{g3} 0.78 & \cellcolor{g1} \bf 0.57 \\
			& & \datasettable{Corridor\,02}
			& \cellcolor{g4} 2.16 & \cellcolor{g6} 5.55 & \cellcolor{g5} 3.55 & \cellcolor{g7} 9.53 & \cellcolor{nosupport} -- & \cellcolor{div} $\times$ & \cellcolor{div} $\times$ & \cellcolor{div} $\times$ & \cellcolor{g3} 0.95 & \cellcolor{g1} \bf 0.70 & \cellcolor{g2} 0.71 \\
			\cmidrule(lr){2-14}
			
			% Divergence Rate
			\addlinespace[2pt]
			& \multicolumn{2}{c|}{Divergence rate [\%]}
			& 23.08 & 23.08 & 15.38 & 7.69 & 66.67 & 53.85 & 30.77 & 15.38 & \bf 0.00 & 7.69 & \bf 0.00 \\
			\cmidrule(lr){2-14}
			
			% Average rank
			\addlinespace[2pt]
			& \multicolumn{2}{c|}{Average rank}
			& 7.08 & 7.69 & 6.77 & 7.69 & 10.78 & 9.08 & 5.15 & 5.62 & 2.85 & 3.69 & \bf 1.77 \\
			\midrule
			
			%%%%%%%%%%%%%%% Open %%%%%%%%%%%%%%%%
			\multirow{21}{*}{\rotatebox{90}{Open}}
			% ---- GEODE (6 rows) ----
			& \multirow{6}{*}{\rotatebox{90}{\datasettable{GD}}}
			& \datasettable{Waterways-Short}
			& \cellcolor{g8} 14.48 & \cellcolor{g6} 3.68 & \cellcolor{g10} 17.93 & \cellcolor{g5} 2.16 & \cellcolor{g1} \bf 0.64 & \cellcolor{g3} 1.25 & \cellcolor{g11} 105.66 & \cellcolor{g9} 15.11 & \cellcolor{g7} 5.22 & \cellcolor{g4} 2.07 & \cellcolor{g2} 0.69 \\
			& & \datasettable{Waterways-Medium}
			& \cellcolor{g6} 49.24 & \cellcolor{g4} 17.83 & \cellcolor{g7} 55.50 & \cellcolor{g8} 94.31 & \cellcolor{g5} 24.87 & \cellcolor{g3} 6.06 & \cellcolor{div} $\times$ & \cellcolor{div} $\times$ & \cellcolor{div} $\times$ & \cellcolor{g2} 5.76 & \cellcolor{g1} \bf 4.07 \\
			& & \datasettable{Waterways-Long}
			& \cellcolor{g7} 146.70 & \cellcolor{div} $\times$ & \cellcolor{g6} 143.06 & \cellcolor{g5} 94.80 & \cellcolor{g3} 14.49 & \cellcolor{g1} \bf 14.08 & \cellcolor{div} $\times$ & \cellcolor{div} $\times$ & \cellcolor{div} $\times$ & \cellcolor{g4} 24.19 & \cellcolor{g2} 14.28 \\
			& & \datasettable{Offroad-02}
			& \cellcolor{g5} 0.37 & \cellcolor{g10} 7.06 & \cellcolor{g8} 0.42 & \cellcolor{g9} 0.86 & \cellcolor{g5} 0.37 & \cellcolor{div} $\times$ & \cellcolor{g5} 0.37 & \cellcolor{g4} 0.33 & \cellcolor{g1} \bf 0.30 & \cellcolor{g1} \bf 0.30 & \cellcolor{g1} \bf 0.30 \\
			& & \datasettable{Offroad-04}
			& \cellcolor{g6} 0.38 & \cellcolor{div} $\times$ & \cellcolor{g8} 0.45 & \cellcolor{g9} 1.85 & \cellcolor{g7} 0.43 & \cellcolor{div} $\times$ & \cellcolor{g5} 0.35 & \cellcolor{g2} 0.29 & \cellcolor{g2} 0.29 & \cellcolor{g2} 0.29 & \cellcolor{g1} \bf 0.28 \\
			& & \datasettable{Offroad-07}
			& \cellcolor{g4} 0.33 & \cellcolor{g10} 1.43 & \cellcolor{g7} 0.39 & \cellcolor{g9} 0.69 & \cellcolor{g8} 0.46 & \cellcolor{div} $\times$ & \cellcolor{g5} 0.35 & \cellcolor{g6} 0.37 & \cellcolor{g1} \bf 0.28 & \cellcolor{g3} 0.32 & \cellcolor{g2} 0.30 \\
			\cmidrule(lr){2-14}
			
			% ---- NTU-VIRAL (3 rows) ----
			& \multirow{3}{*}{\rotatebox{90}{\datasettable{NV}}}
			& \datasettable{SPMS\,01}
			& \cellcolor{g8} 2.73 & \cellcolor{g10} 3.26 & \cellcolor{g7} 1.82 & \cellcolor{g5} 0.87 & \cellcolor{g4} 0.64 & \cellcolor{g6} 0.96 & \cellcolor{g11} 3.44 & \cellcolor{g9} 2.89 & \cellcolor{g1} \bf 0.31 & \cellcolor{g1} \bf 0.31 & \cellcolor{g1} \bf 0.31 \\
			& & \datasettable{SPMS\,02}
			& \cellcolor{g9} 2.42 & \cellcolor{g8} 2.09 & \cellcolor{g7} 1.80 & \cellcolor{g3} 0.37 & \cellcolor{g10} 4.44 & \cellcolor{g6} 0.65 & \cellcolor{div} $\times$ & \cellcolor{g5} 0.55 & \cellcolor{g2} 0.35 & \cellcolor{g3} 0.37 & \cellcolor{g1} \bf 0.34 \\
			& & \datasettable{SPMS\,03}
			& \cellcolor{g8} 1.59 & \cellcolor{g10} 5.24 & \cellcolor{g7} 1.03 & \cellcolor{g4} 0.37 & \cellcolor{g6} 0.52 & \cellcolor{g5} 0.39 & \cellcolor{div} $\times$ & \cellcolor{g9} 2.31 & \cellcolor{g1} \bf 0.25 & \cellcolor{g1} \bf 0.25 & \cellcolor{g1} \bf 0.25 \\
			\cmidrule(lr){2-14}
			
			% ---- ENWIDE (4 rows) ----
			& \multirow{4}{*}{\rotatebox{90}{\datasettable{EW}}}
			& \datasettable{Katzensee\,S}
			& \cellcolor{g8} 0.51 & \cellcolor{g5} 0.28 & \cellcolor{g6} 0.29 & \cellcolor{g7} 0.38 & \cellcolor{div} $\times$ & \cellcolor{g9} 0.54 & \cellcolor{g10} 0.58 & \cellcolor{g3} 0.20 & \cellcolor{g1} \bf 0.18 & \cellcolor{g3} 0.20 & \cellcolor{g1} \bf 0.18 \\
			& & \datasettable{Katzensee\,D}
			& \cellcolor{g9} 0.66 & \cellcolor{g7} 0.48 & \cellcolor{g6} 0.46 & \cellcolor{g5} 0.38 & \cellcolor{g10} 7.44 & \cellcolor{div} $\times$ & \cellcolor{g8} 0.63 & \cellcolor{g1} \bf 0.24 & \cellcolor{g4} 0.25 & \cellcolor{g1} \bf 0.24 & \cellcolor{g1} \bf 0.24 \\
			& & \datasettable{Intersection\,S}
			& \cellcolor{g5} 0.27 & \cellcolor{g6} 0.28 & \cellcolor{g8} 8.79 & \cellcolor{g10} 36.08 & \cellcolor{div} $\times$ & \cellcolor{g7} 0.34 & \cellcolor{g9} 23.74 & \cellcolor{g4} 0.19 & \cellcolor{g1} \bf 0.16 & \cellcolor{g2} 0.17 & \cellcolor{g2} 0.17 \\
			& & \datasettable{Intersection\,D}
			& \cellcolor{g8} 26.89 & \cellcolor{g9} 56.50 & \cellcolor{g5} 2.53 & \cellcolor{g10} 63.28 & \cellcolor{div} $\times$ & \cellcolor{g6} 6.36 & \cellcolor{g7} 8.47 & \cellcolor{g3} 0.26 & \cellcolor{g1} \bf 0.25 & \cellcolor{g1} \bf 0.25 & \cellcolor{g3} 0.26 \\
			\cmidrule(lr){2-14}
			
			% ---- M3DGR (2 rows) ----
			& \multirow{2}{*}{\rotatebox{90}{\datasettable{M3D}}}
			& \datasettable{GNSS-denial\,01}
			& \cellcolor{g7} 5.80 & \cellcolor{g4} 0.14 & \cellcolor{g8} 6.44 & \cellcolor{g9} 11.78 & \cellcolor{nosupport} -- & \cellcolor{g6} 4.63 & \cellcolor{div} $\times$ & \cellcolor{g5} 0.17 & \cellcolor{g2} 0.07 & \cellcolor{g3} 0.08 & \cellcolor{g1} \bf 0.05 \\
			& & \datasettable{GNSS-denial\,02}
			& \cellcolor{g7} 0.60 & \cellcolor{g5} 0.50 & \cellcolor{g8} 1.80 & \cellcolor{g1} \bf 0.48 & \cellcolor{nosupport} -- & \cellcolor{g9} 2.72 & \cellcolor{div} $\times$ & \cellcolor{g3} 0.49 & \cellcolor{g6} 0.53 & \cellcolor{g3} 0.49 & \cellcolor{g1} \bf 0.48 \\
			\cmidrule(lr){2-14}
			
			% Divergence Rate
			\addlinespace[2pt]
			& \multicolumn{2}{c|}{Divergence rate [\%]}
			& \bf 0.00 & 13.33 & \bf 0.00 & \bf 0.00 & 23.08 & 26.67 & 40.00 & 13.33 & 13.33 & \bf 0.00 & \bf 0.00 \\
			\cmidrule(lr){2-14}
			
			% Average rank
			\addlinespace[2pt]
			& \multicolumn{2}{c|}{Average rank}
			& 7.00 & 7.73 & 7.20 & 6.60 & 7.08 & 7.00 & 9.00 & 5.67 & 3.47 & 2.27 & \bf 1.40 \\
			\midrule
			
			%%%%%%%%%%%%%%% Confined-Open %%%%%%%%%%%%%%%%
			\multirow{20}{*}{\rotatebox{90}{Confined\,--\,Open}}
			% ---- SubT-MRS (1 row) ----
			& \multirow{1}{*}{\rotatebox{90}{\datasettable{SM}}}
			& \datasettable{Multi Floor}
			& \cellcolor{div} $\times$ & \cellcolor{g7} 2.84 & \cellcolor{g5} 0.64 & \cellcolor{g6} 1.71 & \cellcolor{g9} 13.41 & \cellcolor{g8} 3.47 & \cellcolor{g1} \bf 0.25 & \cellcolor{div} $\times$ & \cellcolor{g3} 0.28 & \cellcolor{g4} 0.37 & \cellcolor{g2} 0.26 \\
			\cmidrule(lr){2-14}
			
			% ---- SuperLoc (1 row) ----
			& \multirow{1}{*}{\rotatebox{90}{\datasettable{SL}}}
			& \datasettable{Corridor\,02}
			& \cellcolor{g8} 1.89 & \cellcolor{g10} 2.88 & \cellcolor{g9} 1.98 & \cellcolor{g7} 1.74 & \cellcolor{g11} 65.59 & \cellcolor{g5} 1.39 & \cellcolor{g6} 1.48 & \cellcolor{g3} 0.73 & \cellcolor{g4} 0.77 & \cellcolor{g1} \bf 0.69 & \cellcolor{g1} \bf 0.69 \\
			\cmidrule(lr){2-14}
			
			% ---- Oxford Spires (6 rows) ----
			& \multirow{6}{*}{\rotatebox{90}{\datasettable{OS}}}
			& \datasettable{christ-church-01}
			& \cellcolor{g7} 0.79 & \cellcolor{g6} 0.71 & \cellcolor{g7} 0.79 & \cellcolor{g10} 1.04 & \cellcolor{g11} 9.81 & \cellcolor{g2} 0.25 & \cellcolor{g1} \bf 0.05 & \cellcolor{g9} 0.81 & \cellcolor{g4} 0.49 & \cellcolor{g5} 0.57 & \cellcolor{g3} 0.47 \\
			& & \datasettable{christ-church-02}
			& \cellcolor{g7} 0.37 & \cellcolor{g6} 0.32 & \cellcolor{g7} 0.37 & \cellcolor{g10} 0.55 & \cellcolor{g11} 4.61 & \cellcolor{g2} 0.13 & \cellcolor{g1} \bf 0.12 & \cellcolor{g9} 0.39 & \cellcolor{g4} 0.22 & \cellcolor{g5} 0.24 & \cellcolor{g3} 0.21 \\
			& & \datasettable{christ-church-05}
			& \cellcolor{g7} 0.32 & \cellcolor{g7} 0.32 & \cellcolor{g7} 0.32 & \cellcolor{g10} 0.44 & \cellcolor{g11} 2.04 & \cellcolor{g2} 0.15 & \cellcolor{g1} \bf 0.14 & \cellcolor{g6} 0.30 & \cellcolor{g3} 0.23 & \cellcolor{g5} 0.27 & \cellcolor{g4} 0.24 \\
			& & \datasettable{blenheim-palace-01}
			& \cellcolor{g3} 0.13 & \cellcolor{g6} 0.16 & \cellcolor{g4} 0.15 & \cellcolor{g7} 0.19 & \cellcolor{g10} 14.54 & \cellcolor{g9} 0.32 & \cellcolor{div} $\times$ & \cellcolor{g8} 0.29 & \cellcolor{g1} \bf 0.10 & \cellcolor{g4} 0.15 & \cellcolor{g1} \bf 0.10 \\
			& & \datasettable{blenheim-palace-02}
			& \cellcolor{g4} 0.24 & \cellcolor{div} $\times$ & \cellcolor{g4} 0.24 & \cellcolor{g8} 0.44 & \cellcolor{g9} 0.94 & \cellcolor{g7} 0.33 & \cellcolor{div} $\times$ & \cellcolor{g6} 0.26 & \cellcolor{g1} \bf 0.06 & \cellcolor{g3} 0.08 & \cellcolor{g2} 0.07 \\
			& & \datasettable{blenheim-palace-05}
			& \cellcolor{g7} 0.37 & \cellcolor{g6} 0.32 & \cellcolor{g7} 0.37 & \cellcolor{div} $\times$ & \cellcolor{g10} 3.01 & \cellcolor{g9} 0.45 & \cellcolor{g4} 0.12 & \cellcolor{g5} 0.26 & \cellcolor{g2} 0.11 & \cellcolor{g2} 0.11 & \cellcolor{g1} \bf 0.10 \\
			\cmidrule(lr){2-14}
			
			% ---- NarrowWide (6 rows) ----
			& \multirow{6}{*}{\rotatebox{90}{\datasettable{NarrowWide}}}
			& \datasettable{Tracked-01}
			& \cellcolor{g5} 0.23 & \cellcolor{g1} \bf 0.11 & \cellcolor{g3} 0.17 & \cellcolor{g8} 1.48 & \cellcolor{nosupport} -- & \cellcolor{div} $\times$ & \cellcolor{g7} 0.28 & \cellcolor{g9} 3.71 & \cellcolor{g4} 0.21 & \cellcolor{g6} 0.24 & \cellcolor{g2} 0.16 \\
			& & \datasettable{Tracked-02}
			& \cellcolor{div} $\times$ & \cellcolor{g1} \bf 0.11 & \cellcolor{div} $\times$ & \cellcolor{g6} 0.57 & \cellcolor{nosupport} -- & \cellcolor{div} $\times$ & \cellcolor{g3} 0.18 & \cellcolor{div} $\times$ & \cellcolor{g4} 0.20 & \cellcolor{g5} 0.23 & \cellcolor{g2} 0.12 \\
			& & \datasettable{Handheld-A-01}
			& \cellcolor{div} $\times$ & \cellcolor{div} $\times$ & \cellcolor{g5} 0.47 & \cellcolor{g3} 0.20 & \cellcolor{div} $\times$ & \cellcolor{div} $\times$ & \cellcolor{g6} 2.26 & \cellcolor{div} $\times$ & \cellcolor{g4} 0.22 & \cellcolor{g1} \bf 0.18 & \cellcolor{g2} 0.19 \\
			& & \datasettable{Handheld-A-02}
			& \cellcolor{div} $\times$ & \cellcolor{div} $\times$ & \cellcolor{g3} 0.25 & \cellcolor{g5} 0.32 & \cellcolor{div} $\times$ & \cellcolor{div} $\times$ & \cellcolor{g6} 0.58 & \cellcolor{div} $\times$ & \cellcolor{g4} 0.28 & \cellcolor{g2} 0.22 & \cellcolor{g1} \bf 0.15 \\
			& & \datasettable{Handheld-B-01}
			& \cellcolor{g5} 1.45 & \cellcolor{g3} 0.32 & \cellcolor{g6} 6.61 & \cellcolor{div} $\times$ & \cellcolor{nosupport} -- & \cellcolor{div} $\times$ & \cellcolor{div} $\times$ & \cellcolor{div} $\times$ & \cellcolor{g2} 0.24 & \cellcolor{g4} 0.38 & \cellcolor{g1} \bf 0.15 \\
			& & \datasettable{Handheld-B-02}
			& \cellcolor{g5} 3.17 & \cellcolor{div} $\times$ & \cellcolor{g3} 2.03 & \cellcolor{div} $\times$ & \cellcolor{nosupport} -- & \cellcolor{div} $\times$ & \cellcolor{div} $\times$ & \cellcolor{div} $\times$ & \cellcolor{g2} 0.67 & \cellcolor{g4} 2.07 & \cellcolor{g1} \bf 0.17 \\
			\cmidrule(lr){2-14}
			
			% Divergence Rate
			\addlinespace[2pt]
			& \multicolumn{2}{c|}{Divergence rate [\%]}
			& 28.57 & 28.57 & 7.14 & 21.43 & 20.00 & 42.86 & 28.57 & 42.86 & \bf 0.00 & \bf 0.00 & \bf 0.00\\
			\cmidrule(lr){2-14}
			
			% Average rank
			\addlinespace[2pt]
			& \multicolumn{2}{c|}{Average rank}
			& 7.21 & 6.86 & 5.71 & 7.93 & 10.40 & 7.57 & 5.57 & 8.43 & 3.00 & 3.64 & \bf 1.86 \\
			\midrule
			& \multicolumn{2}{c|}{Total divergence rate [\%]}
			& 16.67 & 21.43 & 7.14 & 9.52 & 34.38 & 40.48 & 33.33 & 23.81 & 4.76 & 2.38 & \bf 0.00 \\
			\midrule
			& \multicolumn{2}{c|}{Total average rank}
			& 7.10 & 7.43 & 6.57 & 7.38 & 9.16 & 7.83 & 6.67 & 6.57 & 3.12 & 3.17 & \bf 1.67 \\
			\bottomrule[0.8pt]
		\end{tabular}
	\end{threeparttable}
	\vspace{-0.1cm}
	\label{tab:benchmark}
\end{table*}

\begin{figure*}[t!]
	\captionsetup[subfigure]{skip=1pt}
	\centering
	\makebox[\textwidth][c]{%
	\begin{subfigure}[t]{0.3275\textwidth}
		\vspace{0pt}
		\centering
		\includegraphics[width=0.935\linewidth]{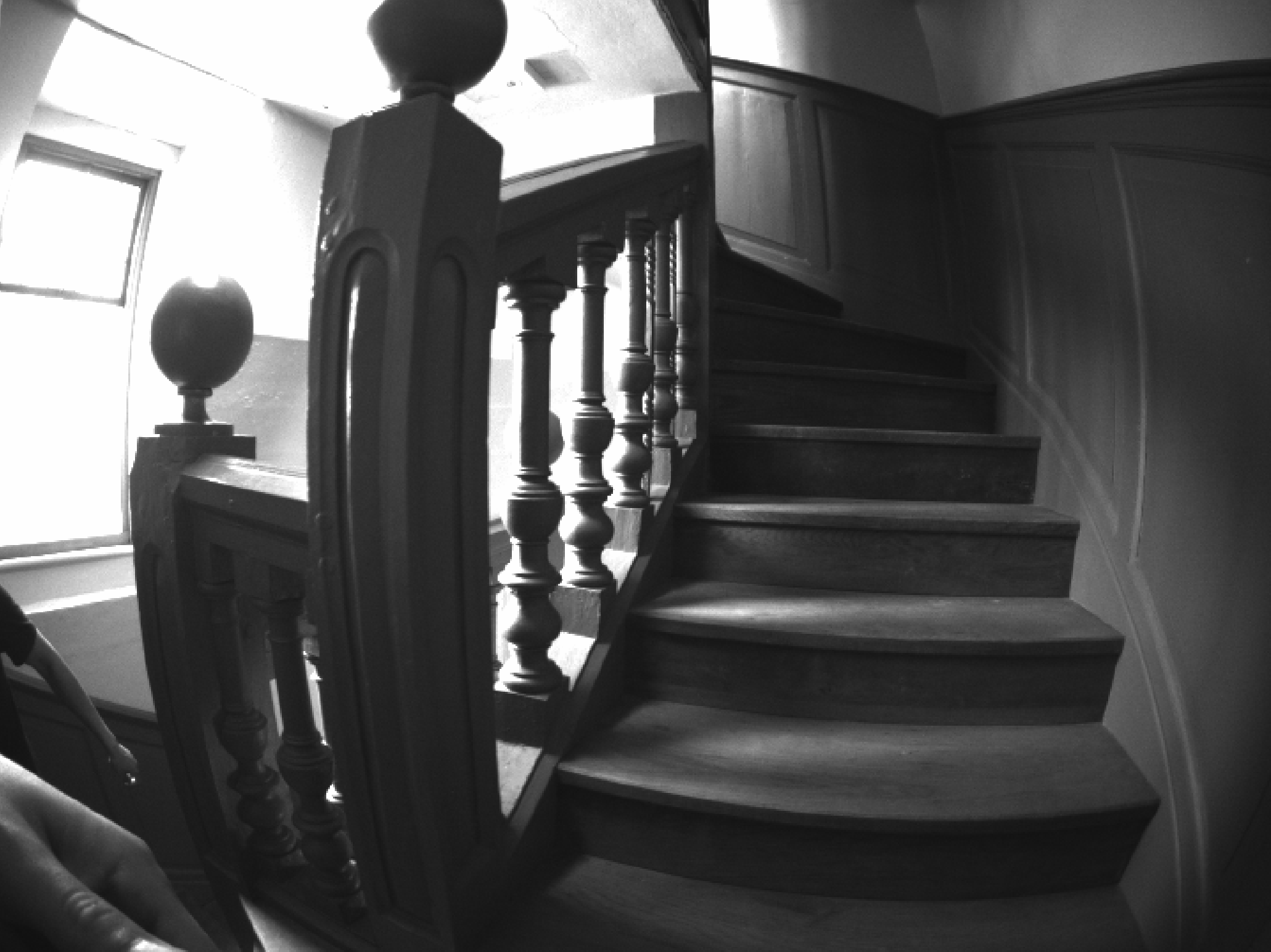}\\[-1pt]
		\includegraphics[width=0.9355\linewidth]{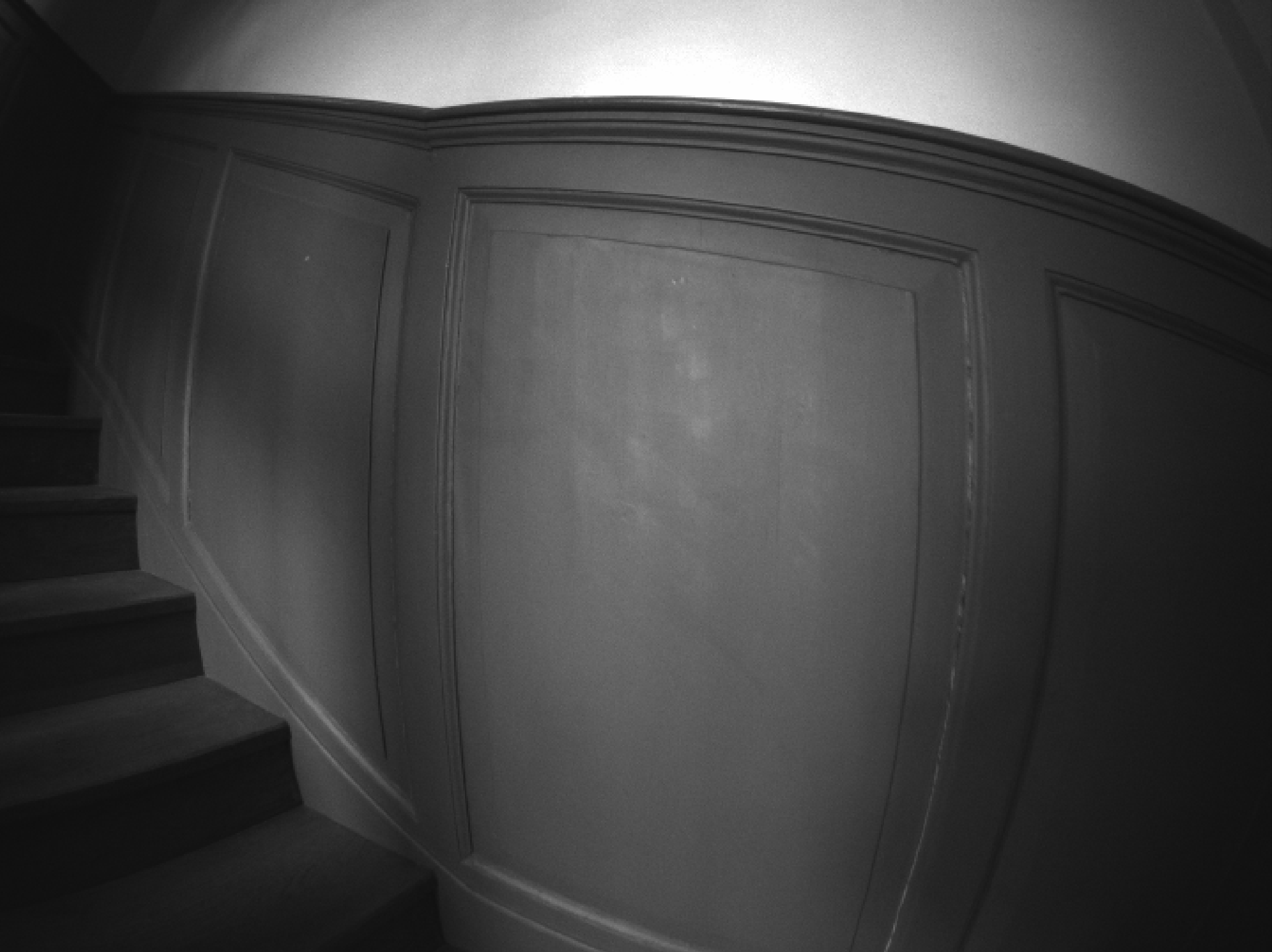}
		\par\vspace{2pt}
		\caption{Confined staircase}
		\label{fig:exp16_onboard_image}
	\end{subfigure}
	\hspace{0.008\textwidth}
	\begin{subfigure}[t]{0.3235\textwidth}
		\vspace{0pt}
		\centering
		\includegraphics[width=\linewidth]{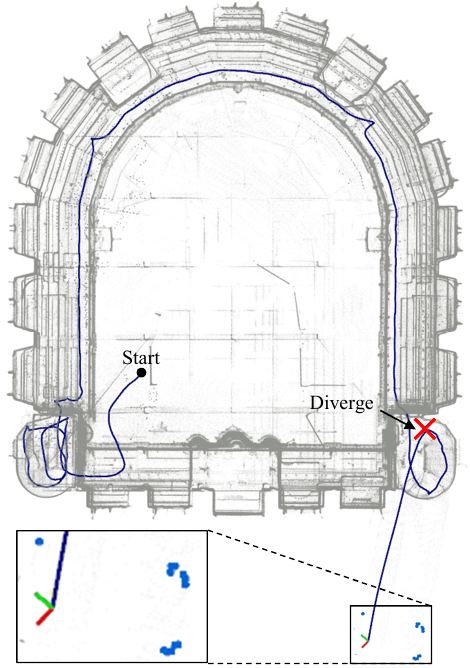}
		\caption{PV-LIO~\cite{hviktortsoi2023pvlio} (Baseline)}
		\label{fig:exp16_baseline}
	\end{subfigure}
	\hspace{0.008\textwidth}
	\begin{subfigure}[t]{0.3233\textwidth}
		\vspace{0pt}
		\centering
		\includegraphics[width=\linewidth]{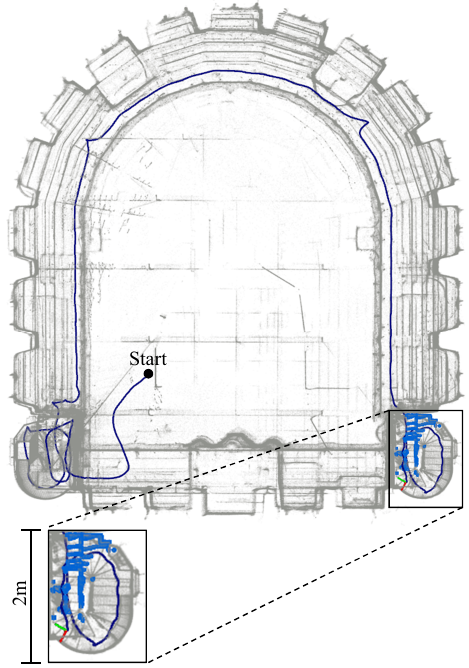}
		\caption{Baseline with adaptive voxelization}
		\label{fig:exp16_baseline_with_adap_vox}
	\end{subfigure}}
	\vspace{-0.15cm}
	\caption{Qualitative comparison on the \dataset{Exp\,16} sequence of the \dataset{2022 HILTI}~\cite{zhang2022ral} dataset. (a)~Onboard views captured in a confined staircase. (b)~Divergence of the baseline~\cite{hviktortsoi2023pvlio} in the confined staircase shown in~(a). (c)~Stable odometry estimation by the baseline with adaptive voxelization in the same staircase. Note that the coordinate frame corresponds to the body frame of the mobile platform. The accumulated map is shown in gray, while the current voxelized scan is shown in blue.}
	\label{fig:exp16_qualitative_comparison}
	\vspace{-0.4cm}
\end{figure*}

\begin{figure*}[t!]
	\captionsetup[subfigure]{skip=1pt}
	\centering
	\makebox[\textwidth][c]{%
	\begin{subfigure}[t]{0.309\textwidth}
		\vspace{0pt}
		\centering
		\includegraphics[width=\linewidth,trim=63 32 63 0,clip]{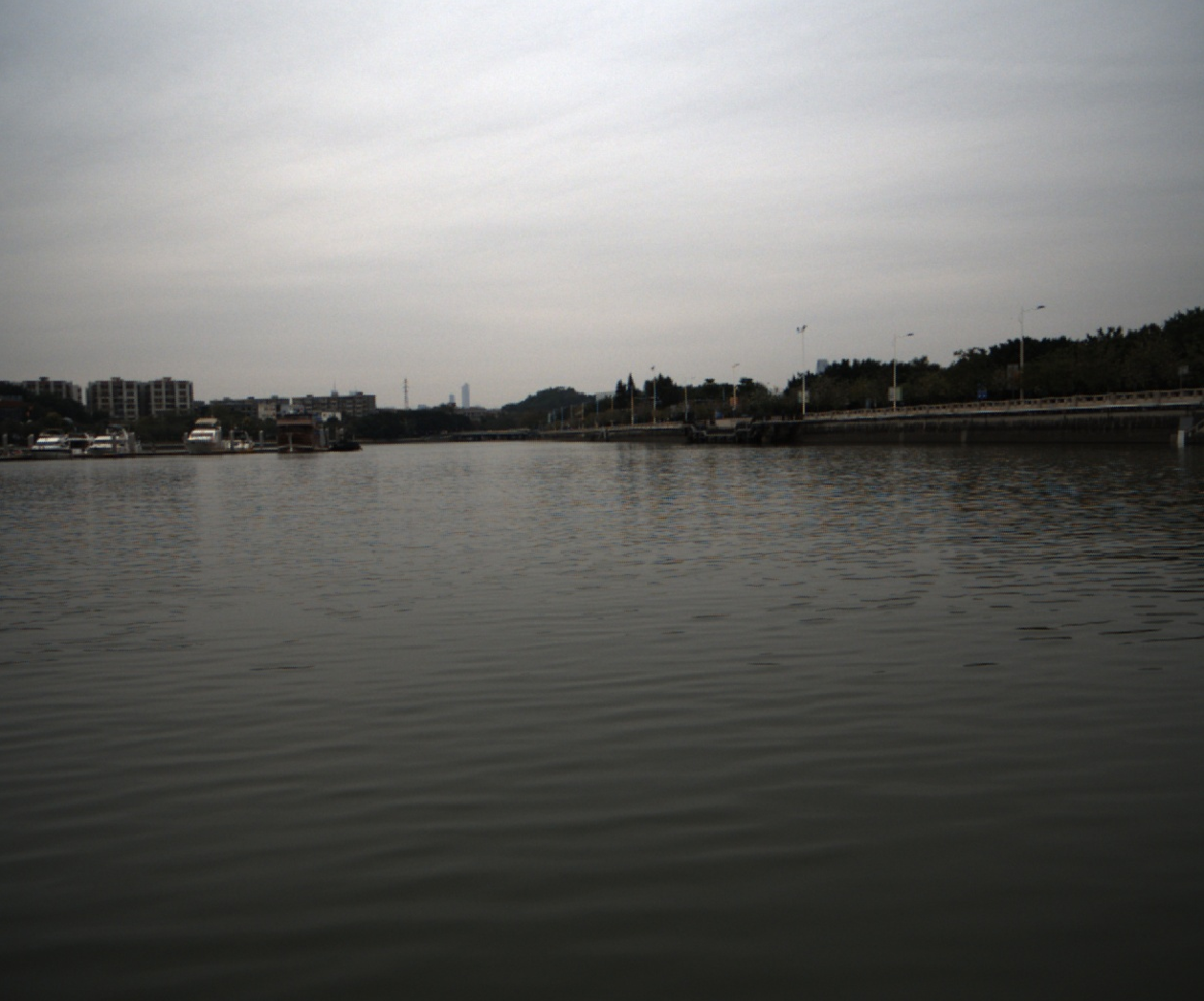}
		\caption{Open waterway}
		\label{fig:waterways_short_onboard_image}
	\end{subfigure}
	\hspace{0.0075\textwidth}
	\begin{subfigure}[t]{0.325\textwidth}
		\vspace{0pt}
		\centering
		\includegraphics[width=\linewidth]{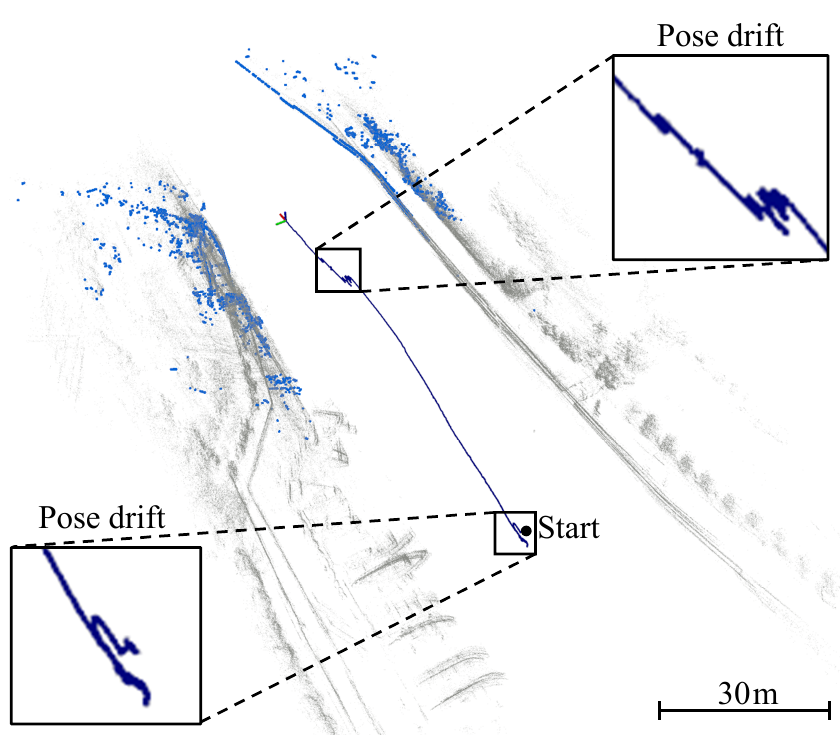}
		\caption{PV-LIO~\cite{hviktortsoi2023pvlio} (Baseline)}
		\label{fig:waterways_short_baseline}
	\end{subfigure}
	\hspace{0.0075\textwidth}
	\begin{subfigure}[t]{0.321\textwidth}
		\vspace{0pt}
		\centering
		\includegraphics[width=\linewidth]{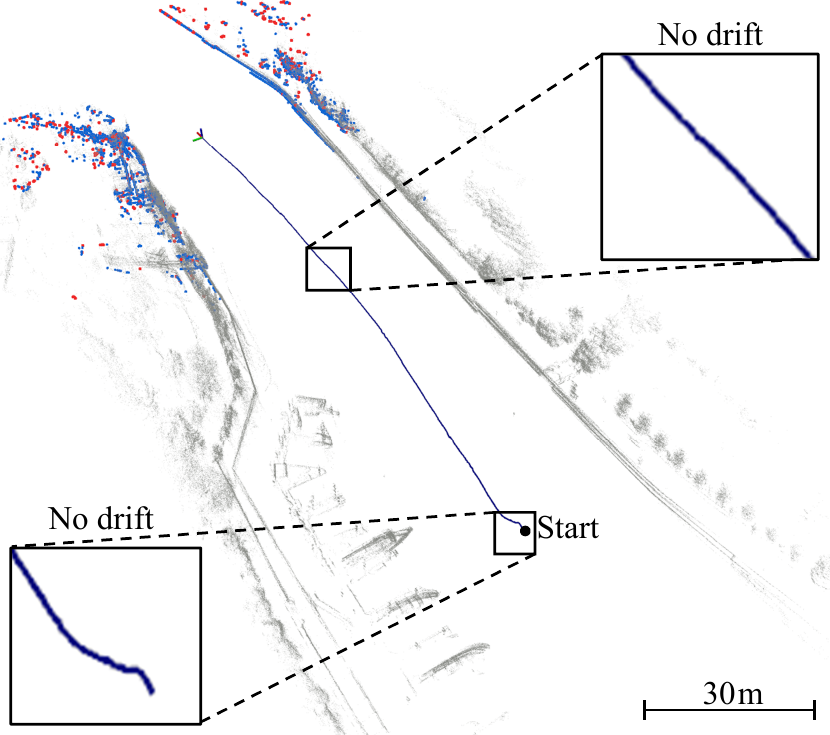}
		\caption{Baseline with hybrid-metric state update}
		\label{fig:waterways_short_baseline_with_hybrid_metric}
	\end{subfigure}}
	\vspace{-0.15cm}
	\caption{Qualitative comparison on the \dataset{Waterways-Short} sequence of the \dataset{GEODE}~\cite{chen2026ijrr} dataset. (a)~Onboard view captured in an open waterway. (b)~Pose drift of the baseline~\cite{hviktortsoi2023pvlio} in the open area shown in~(a). (c)~Stable odometry estimation by the baseline with the hybrid-metric state update in the same open area. Note that the coordinate frame corresponds to the body frame of the mobile platform. The accumulated map is shown in gray, while the point-to-plane constraints are shown in blue and the point-to-point constraints in red.}
	\label{fig:waterways_short_qualitative_comparison}
	\vspace{-0.4cm}
\end{figure*}

\begin{figure*}[t!]
	\captionsetup[subfigure]{skip=1pt}
	\centering
	\begin{subfigure}[t]{0.22\textwidth}
		\centering
		\includegraphics[width=\linewidth]{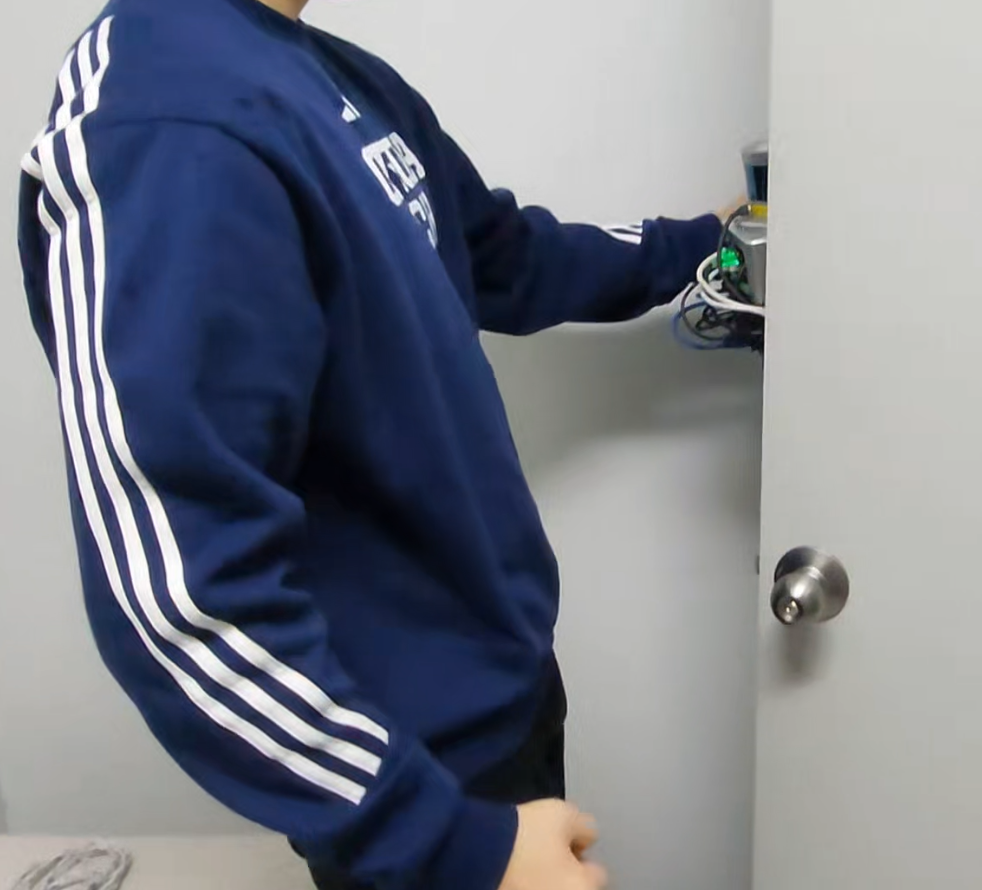}
		\caption{Confined corner}
		\label{fig:handheld_a_01_confined_corner}
	\end{subfigure}
	\hfill
	\begin{subfigure}[t]{0.375\textwidth}
		\centering
		\includegraphics[width=\linewidth]{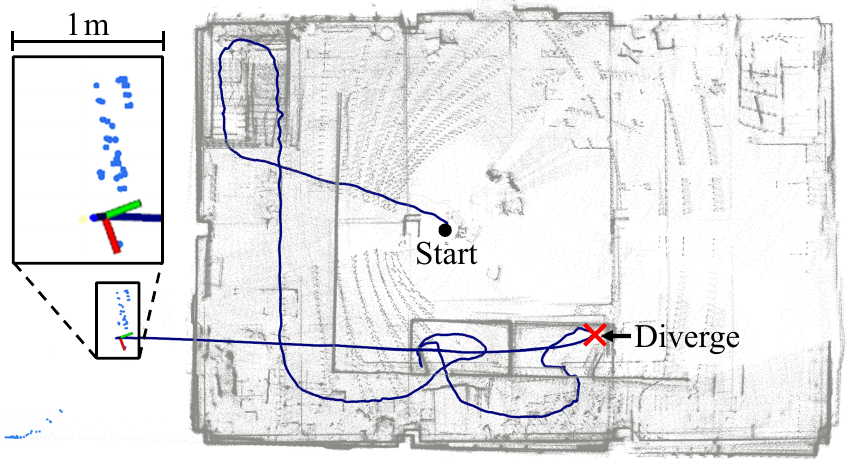}
		\caption{PV-LIO~\cite{hviktortsoi2023pvlio} (Baseline)}
		\label{fig:handheld_a_01_pv_lio}
	\end{subfigure}
	\hfill
	\begin{subfigure}[t]{0.375\textwidth}
		\centering
		\includegraphics[width=\linewidth]{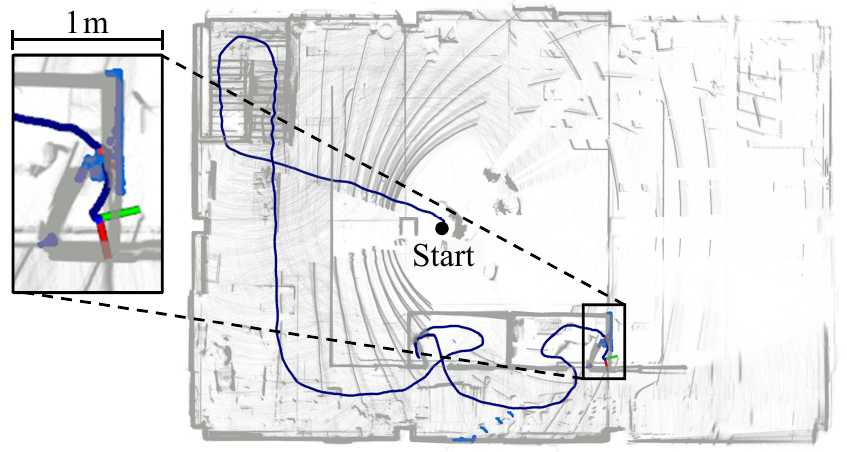}
		\caption{Ours}
		\label{fig:handheld_a_01_genz_lio}
	\end{subfigure}
	\vspace{-0.1cm}
	\caption{Qualitative comparison on the \dataset{Handheld-A-01} sequence of the \dataset{NarrowWide} dataset. (a)~Data acquisition using a handheld device in an extremely confined corner less than $0.5\,\mathrm{m}$ wide. (b)~Divergence of the baseline~\cite{hviktortsoi2023pvlio} in the confined corner shown in~(a). (c)~Stable odometry estimation by GenZ-LIO in the same corner. This scene corresponds to region~C in Fig.~\ref{fig:adap_vox_plot}. Note that the coordinate frame corresponds to the body frame of the handheld device. The accumulated map is shown in gray, while the point-to-plane constraints are shown in blue and the point-to-point constraints in red.}
	\label{fig:handheld_a_01_pv_genz_comparison}
	\vspace{-0.55cm}
\end{figure*}

% Results
%% 1. Overall performance
The benchmark results are summarized in Table~\ref{tab:benchmark}.
GenZ-LIO completed all sequences without divergence, resulting in a total divergence rate of 0.00\%.
It also achieved the lowest total average rank of 1.67 among the compared methods.
These results suggest that GenZ-LIO maintains stable odometry estimation while preserving competitive accuracy across a wide range of spatial scales.
To better understand these results, we next investigate how each proposed module contributes to the overall behavior of GenZ-LIO in representative failure-prone scenarios.

%% 2. Effect of scale-aware adaptive voxelization
The effect of the proposed scale-aware adaptive voxelization was most evident in sequences that included extremely confined scenes.
In the \dataset{Exp\,16} sequence of the \dataset{2022 HILTI}~\cite{zhang2022ral} dataset, most methods diverged in the confined staircases, where a fixed voxel size could cause the voxelized point count to drop sharply, resulting in insufficient geometric information for reliable state estimation.
In contrast, \emph{baseline with adaptive voxelization} remained stable by adjusting the voxel size to drive the voxelized point count toward a scale-informed setpoint; see Fig.~\ref{fig:exp16_qualitative_comparison} and Sec.~\ref{subsec: exp_ablation_gain_scheduling}.
Similarly, AdaLIO~\cite{lim2023ur}, which also employs an adaptive voxelization strategy, avoided divergence on this sequence.
Among the methods without adaptive voxelization, only Point-LIO~\cite{he2023ais} remained stable, likely due to its point-wise state update that partially compensated for the loss of geometric characteristics in highly confined regions.

%% 3. Effect of hybrid-metric state update
The benefit of the proposed hybrid-metric state update was prominent in open outdoor environments with weak planar structure.
In the \dataset{Waterways-Short}, \dataset{Waterways-Medium}, and \dataset{Waterways-Long} sequences of the \dataset{GEODE}~\cite{chen2026ijrr} dataset, LiDAR returns were absent from the water surface, leading to a scarcity of reliable planar constraints along degenerate directions.
In such scenarios, methods relying primarily on point-to-plane residuals often suffered from degraded accuracy or divergence, as reflected by the large errors or divergent cases.
In contrast, \emph{baseline with hybrid-metric state update} mitigated large errors induced by weak observability by leveraging point-to-point residuals to compensate for the lack of planar structure; see Fig.~\ref{fig:waterways_short_qualitative_comparison} and Sec.~\ref{subsec: exp_hybrid_metric}.
Consistent with this observation, LIO-EKF~\cite{wu2024icra}, which utilizes point-to-point residuals, also showed improved robustness in environments with limited planar structure.

%% 4. Combined effect on NarrowWide
The combined effect of the two proposed modules was further observed on the \dataset{NarrowWide} dataset, which contains extreme scene-scale variations.
In the \dataset{Handheld-A-01} and \dataset{Handheld-A-02} sequences, the baseline method diverged due to a sharp reduction in voxelized point count in highly confined spaces; see Fig.~\ref{fig:handheld_a_01_pv_genz_comparison}.
In contrast, the adaptive voxelization module adjusted the voxel size in these regions, thereby helping preserve sufficient geometric constraints.
In addition, the hybrid-metric state update improved robustness by exploiting point-to-point residuals when reliable point-to-plane constraints were unavailable in noisy or weakly structured regions.
When both modules were jointly integrated, the full GenZ-LIO system consistently avoided divergence and achieved stable odometry estimation across all \dataset{NarrowWide} sequences, demonstrating the benefit of combining the two modules in challenging scale-varying environments.
A supplementary evaluation in general operating scenarios is provided in Sec.~\ref{sec:general_operating_scenarios} for a broader assessment of GenZ-LIO's practical applicability.

\begin{figure*}[t!]
	\centering
	\includegraphics[width=1.0\linewidth]{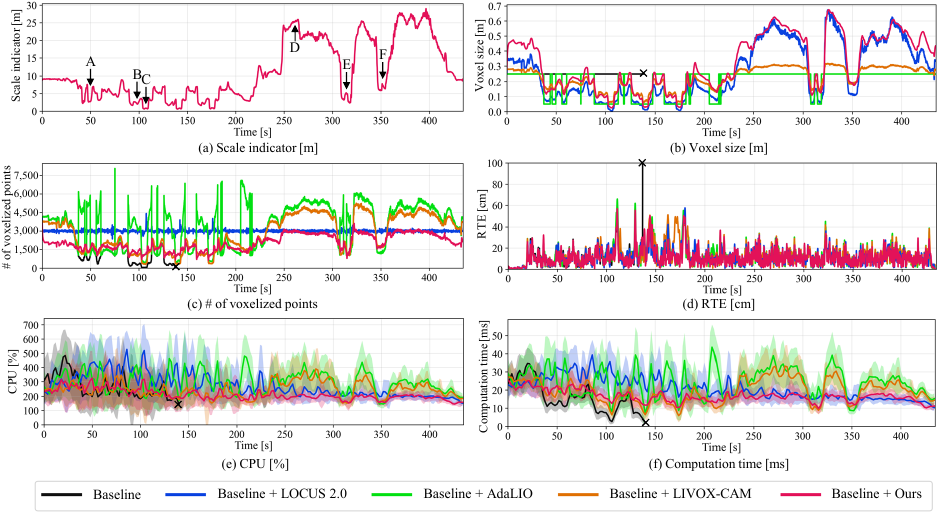}
	\vspace{-0.1cm}
	\caption{Comparison of adaptive voxelization strategies~\cite{reinke2022ral, lim2023ur, cheng2025ral} applied to a baseline system~\cite{hviktortsoi2023pvlio} on the \dataset{Handheld-A-01} sequence of the \dataset{NarrowWide} dataset. (a)~Temporal evolution of the proposed scale indicator $\bar m_t$. The markers A–F correspond to the regions A–F in Fig.~\ref{fig:Figure1}, respectively. (b)~Voxel sizes adjusted by each adaptive strategy. (c)~Number of points in the voxelized scan obtained using the adjusted voxel size. (d)~Relative translational error of the estimated poses. (e)~CPU usage, visualized using a sliding-window temporal average with a shaded band indicating one standard deviation to reduce short-term fluctuations. (f)~Computation time per frame, visualized using the same smoothing scheme for clarity. The baseline system diverged during the sequence and is therefore plotted only up to the divergence point, which is indicated by a $\times$ marker. These results are further summarized in Table~\ref{tab:adap_vox_comparison}.}
	\label{fig:adap_vox_plot}
	\vspace{-0.4cm}
\end{figure*}

\subsection{COMPARISON WITH ADAPTIVE VOXELIZATION STRATEGIES}\label{subsec: exp_adap_vox_comparison}
% claim
The second experiment compares the proposed scale-aware adaptive voxelization with existing adaptive voxelization strategies in a confined--open scenario.
This experiment supports our second claim that the scale-aware adaptive voxelization enables robust state estimation while efficiently regulating computational resources by accounting for varying spatial~scales.

% setup
For a fair comparison, all adaptive voxelization strategies~\cite{reinke2022ral, lim2023ur, cheng2025ral} are evaluated within the same baseline LIO framework, PV-LIO~\cite{hviktortsoi2023pvlio}, with the adaptive voxelization module being the only varying component.
The evaluation is conducted on the \dataset{Handheld-A-01} sequence of the \dataset{NarrowWide} dataset, which spans a wide range of spatial scales, from extremely confined corners to open spaces.

For LOCUS\,2.0~\cite{reinke2022ral} and LIVOX-CAM~\cite{cheng2025ral}, which aim to track a fixed target number of voxelized points, we set a fixed setpoint of $N_{\mathrm{desired}}^{\mathrm{fixed}} = 3,\!000$ to retain adequate geometric information in open outdoor scenes.
For AdaLIO~\cite{lim2023ur}, which adopts a threshold-based switching strategy, we set the coarse voxel size, fine voxel size, and point count threshold to $d_{\mathrm{coarse}}^{\mathrm{fixed}} = 0.25\,\mathrm{m}$, $d_{\mathrm{fine}}^{\mathrm{fixed}} = 0.05\,\mathrm{m}$, and $\tau_N^{\mathrm{fixed}} = 1,\!000$, respectively.
For the proposed method, the voxel size is adaptively adjusted to track the setpoint $N_{\mathrm{desired},t}$ that varies based on the scale indicator $\bar m_t$, where the lower and upper bounds of $N_{\mathrm{desired},t}$ are set to $N_{\min} = 1,\!000$ and $N_{\max} = 3,\!000$,~respectively.

Performance is assessed in terms of both computational efficiency and odometry accuracy.
Computational efficiency is evaluated using CPU usage and per-frame computation time, while odometry accuracy is measured by relative translational error (RTE) and ATE.
CPU usage is reported as the system-wide processor utilization during runtime, expressed as a percentage, where 100\% corresponds to the full utilization of a single CPU core.
CPU usage measurements are obtained using a cross-platform system monitoring library~\cite{psutil}.

% overall performance
As summarized in Table~\ref{tab:adap_vox_comparison}, the proposed method achieved the best odometry accuracy while also exhibiting the lowest CPU usage and computation time among the compared adaptive voxelization strategies.
% pv-lio
As illustrated in Figs.~\ref{fig:handheld_a_01_pv_genz_comparison} and~\ref{fig:adap_vox_plot}, the baseline PV-LIO~\cite{hviktortsoi2023pvlio}, which maintained a fixed voxel size of $d_0 = 0.25\,\mathrm{m}$, suffered a substantial reduction in the number of voxelized points in a confined corner around $140\,\mathrm{s}$ and consequently diverged due to insufficient geometric constraints for reliable state estimation.
% locus
In contrast, when adaptive voxelization strategies were applied to the baseline system, divergence was not observed in the confined regions.
LOCUS\,2.0~\cite{reinke2022ral} generally maintained the predefined fixed target number of voxelized points, as illustrated in Fig.~\ref{fig:adap_vox_plot}(c).
Consequently, the number of voxelized points remained relatively high even in confined environments.
This behavior resulted in relatively higher CPU usage and computation time than the other methods, as shown in Figs.~\ref{fig:adap_vox_plot}(e) and~(f).

% adalio
Next, AdaLIO~\cite{lim2023ur} switched the voxel size to a fine value when the number of voxelized points obtained with the coarse voxel size fell below the predefined threshold $\tau_N^{\mathrm{fixed}} = 1,\!000$, as illustrated in Figs.~\ref{fig:adap_vox_plot}(b) and~(c).
This switching behavior prevented the voxelized point count from becoming excessively low in confined environments.
However, when the point count obtained with the coarse voxel size was close to the threshold~$\tau_N^{\mathrm{fixed}}$, frequent switching caused noticeable fluctuations in the voxelized point count, highlighting the need for continuous voxel size adjustment.

% livox-cam
LIVOX-CAM~\cite{cheng2025ral} adopts a voxel size adjustment strategy similar to that of LOCUS\,2.0, aiming to regulate the voxel size such that the number of voxelized points converges to a fixed target $N_\mathrm{desired}^\mathrm{fixed}$.
Nevertheless, as shown in Fig.~\ref{fig:adap_vox_plot}(c), the voxelized point count did not consistently remain at the predefined target of $N_{\mathrm{desired}}^{\mathrm{fixed}} = 3,\!000$.
In open areas such as region~D, the voxelized point count remained high, resulting in relatively higher CPU usage and computation time, as illustrated in Figs.~\ref{fig:adap_vox_plot}(e) and~(f).
A detailed discussion of this tracking behavior of LIVOX-CAM is addressed in Sec.~\ref{subsec: exp_ablation_gain_scheduling}.

% ours
Unlike these methods, the proposed scale-aware adaptive voxelization estimates the spatial scale and adjusts the voxel size to regulate the number of voxelized points toward a scale-informed setpoint.
As shown in Fig.~\ref{fig:adap_vox_plot}(c), in confined and narrow spaces such as regions~B and~C (also refer to Fig.~\ref{fig:Figure1} for better understanding), the proposed method operated with fewer voxelized points, reducing computational burden while maintaining stable odometry estimation.
In open areas such as region~D, the voxelized point count was maintained at a higher level while still being bounded by the upper limit $N_{\max} = 3,\!000$.
These results show that accounting for spatial scale enables the proposed method to maintain stable odometry estimation while using computational resources more efficiently across varying environments.

\begin{table}[!t]
	\caption{Comparison of computational efficiency and odometry accuracy for different adaptive voxelization strategies evaluated on the \dataset{Handheld-A-01} sequence of the \dataset{NarrowWide} dataset.
		CPU usage and computation time are reported using the mean and the 95th percentile (p95).
		RTE and ATE denote the relative and absolute translational errors, respectively, both reported as RMSE.
		All adaptive voxelization methods~\cite{reinke2022ral,lim2023ur,cheng2025ral} are evaluated within the same baseline framework~\cite{hviktortsoi2023pvlio}.
		The baseline values shown in gray report CPU usage and computation time only up to the divergence point and are therefore excluded from direct comparison.
		The symbol ``$\times$'' indicates divergence, and the best performance is highlighted in \textbf{bold}. The temporal behavior of each method over the sequence is further illustrated in Fig.~\ref{fig:adap_vox_plot}.}
	\label{tab:adap_vox_comparison}
	\centering
	\setlength{\tabcolsep}{2.1pt}
	\renewcommand\arraystretch{0.8}
	\begin{tabular}{lccccccc}
		\toprule[0.8pt]
		Method
		& \multicolumn{2}{c}{CPU\,[\%]}
		& \multicolumn{2}{c}{Comp. time\,[ms]}
		& \multicolumn{1}{c}{RTE\,[cm]}
		& \multicolumn{1}{c}{ATE\,[m]} \\
		\cmidrule(lr){2-3}
		\cmidrule(lr){4-5}
		\cmidrule(lr){6-6}
		\cmidrule(lr){7-7}
		& Mean & p95 & Mean & p95 & RMSE & RMSE \\
		\midrule
		\textcolor{gray}{Baseline~\cite{hviktortsoi2023pvlio}}
		& \textcolor{gray}{270.12} & \textcolor{gray}{485.02} & \textcolor{gray}{16.83} & \textcolor{gray}{36.70} & \textcolor{gray}{$\times$} & \textcolor{gray}{$\times$} \\
		\midrule
		+\,LOCUS\,2.0~\cite{reinke2022ral}
		& 273.92 & 477.59
		& 20.98 & 42.65
		& 1.49 & 0.29 \\
		+\,AdaLIO~\cite{lim2023ur}
		& 294.57 & 467.87
		& 24.14 & 50.35
		& 1.56 & 0.34 \\
		+\,LIVOX-CAM~\cite{cheng2025ral}
		& 245.93 & 372.98
		& 19.60 & 34.59
		& 1.60 & 0.30 \\
		+\,Ours
		& \textbf{212.23} & \textbf{314.16}
		& \textbf{16.64} & \textbf{26.24}
		& \textbf{1.42} & \textbf{0.16} \\
		\bottomrule[0.8pt]
	\end{tabular}
	\vspace{-0.3cm}
\end{table}

\subsection{ABLATION STUDY ON SENSITIVITY-INFORMED GAIN SCHEDULING}\label{subsec: exp_ablation_gain_scheduling}
% claim
The third experiment conducts an ablation study on the proposed sensitivity-informed gain scheduling to analyze its effect on tracking the desired number of voxelized points.
This experiment supports our third claim that the sensitivity-informed gain scheduling improves transient response, resulting in faster convergence and reduced overshoot and oscillations during voxel size control.

The evaluation is conducted on the \dataset{Exp\,16} sequence of the \dataset{2022 HILTI}~\cite{zhang2022ral} dataset, which exhibits abrupt and severe spatial-scale variations and thus poses a challenging test case for voxel size control.
To evaluate setpoint tracking performance, we employ two standard control performance indices: the integral of absolute error (IAE) and overshoot.
The IAE, which measures the cumulative tracking error over the entire sequence, is defined over $T$ discrete timesteps as $\sum_{t=1}^{T} \left| N_t - N_{\mathrm{desired},t} \right| \Delta t_\mathrm{scan}$, where $N_t$ denotes the number of points in the voxelized scan $\mathcal{V}_t$ obtained with the updated voxel size $d_t$.
Overshoot is defined as $\max\!\left(0,\max\limits_{1 \leq t \leq T} \frac{N_t - N_{\mathrm{desired},t}}{N_{\mathrm{desired},t}}\right)$, which quantifies the maximum relative extent to which the response exceeds the desired setpoint.
Together, these metrics capture overall tracking accuracy and transient oscillatory behavior during voxel size control.
Additionally, we evaluate the RMSE of the ATE to assess odometry estimation accuracy.
By jointly considering IAE, overshoot, and ATE, this ablation study examines not only how effectively each method regulates voxelized point count, but also how such regulation affects downstream odometry performance.

\begin{figure*}[t!]
	\centering
	\includegraphics[width=1.0\linewidth]{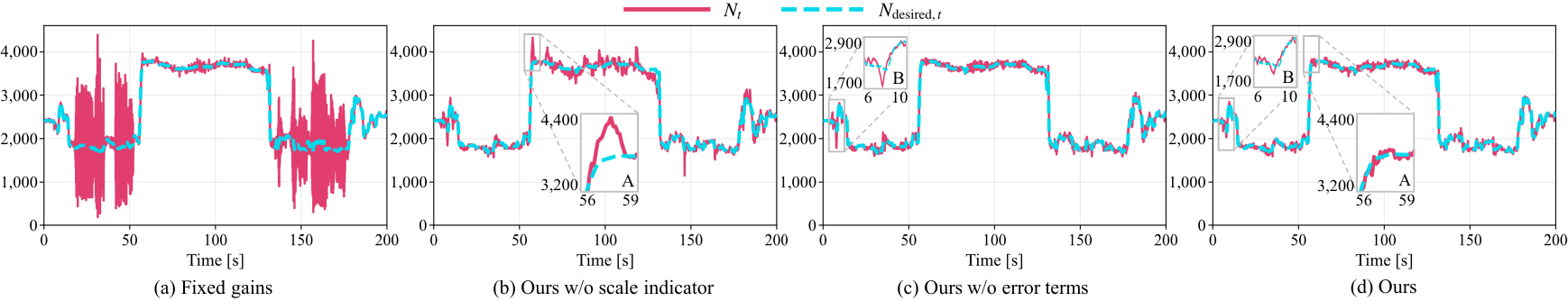}
	\vspace{-0.25cm}
	\caption{Setpoint tracking performance comparison on the \dataset{Exp\,16} sequence of the \dataset{2022 HILTI}~\cite{zhang2022ral} dataset. The plots show the temporal evolution of the scale-informed setpoint~$N_{\mathrm{desired},t}$ and the corresponding voxelized point count~$N_t$ for the following ablations: (a)~PD controller with fixed gains, (b)~ours without scale indicator~$\bar m_t$, (c)~ours without the error terms~$|e_t|$ and $|\Delta e_t|$ in gain scheduling, and (d)~ours with sensitivity-informed gain scheduling. Region~A in~(b) and region~B in~(c) correspond to the zoomed regions in~(d), highlighting differences in local tracking behavior.}
	\label{fig:gain_scheduling_ablation}
	\vspace{-0.4cm}
\end{figure*}

We compare six control strategies: the \emph{linear scaling strategy} proposed by LOCUS\,2.0~\cite{reinke2022ral} (Fig.~\ref{fig:adap_vox_diagram_comparison}(a)); the \emph{volume-based scaling strategy} proposed by LIVOX-CAM~\cite{cheng2025ral} (Fig.~\ref{fig:adap_vox_diagram_comparison}(c));
a PD controller with fixed gains~(referred to as \emph{Fixed gains} in Table~\ref{tab:gain_scheduling_ablation}),
where the proportional and derivative gains are set to the midpoints of their respective bounds, i.e., $K_{p,t} = \frac{K_{p,\min} + K_{p,\max}}{2}$ and $K_{d,t} = \frac{K_{d,\min} + K_{d,\max}}{2}$;
our method without the scale indicator~(referred to as \emph{Ours w/o scale indicator} in Table~\ref{tab:gain_scheduling_ablation}); our method without the magnitudes of the tracking error and its derivative (referred to as \emph{Ours w/o error terms} in Table~\ref{tab:gain_scheduling_ablation});
and the full sensitivity-informed gain scheduling proposed in this work, as illustrated in Fig.~\ref{fig:adap_vox_diagram_comparison}(d).
All methods are intentionally configured to use the same scale indicator $\bar m_t$ and the same scale-to-setpoint mapping for generating $N_{\mathrm{desired},t}$, and are evaluated within the same baseline odometry framework, PV-LIO~\cite{hviktortsoi2023pvlio}.
That is, once $N_{\mathrm{desired},t}$ is determined, only the voxel size control strategy is varied, allowing the ablation to isolate the effect of different control mechanisms.

\begin{table}[!t]
	\caption{Ablation study results on voxel size control strategies for tracking a scale-informed setpoint $N_{\mathrm{desired},t}$, evaluated on the \dataset{Exp\,16} sequence of the \dataset{2022 HILTI}~\cite{zhang2022ral} dataset. IAE denotes the integral of absolute error; see Sec.~\ref{subsec: exp_ablation_gain_scheduling}. All control methods are evaluated within the same baseline odometry framework. The best performance is highlighted in \textbf{bold}.}
	\label{tab:gain_scheduling_ablation}
	\centering
	\setlength{\tabcolsep}{4.8pt}
	\renewcommand\arraystretch{0.8}
	\begin{tabular}{lccc}
		\toprule[0.8pt]
		Method & IAE [$\times 10^3$] $\downarrow$ & Overshoot $\downarrow$ & ATE [m] $\downarrow$ \\
		\midrule
		Linear scaling~\cite{reinke2022ral} & 14.03 & 0.24 & 0.21 \\
		Volume-based scaling~\cite{cheng2025ral} & 270.71 & 0.87 & 0.34 \\
		Fixed gains & 53.74 & 1.63 & 0.24 \\
		Ours w/o scale indicator & 13.32 & 0.16 & 0.19 \\
		Ours w/o error terms & 7.95 & 0.15 & 0.21 \\
		Ours & $\textbf{6.04}$ & \textbf{0.09} & \textbf{0.13} \\
		\bottomrule[0.8pt]
	\end{tabular}
	\vspace{-0.3cm}
\end{table}

The results of all methods are summarized in Table~\ref{tab:gain_scheduling_ablation}.
Compared with the scaling-based control strategies~\cite{reinke2022ral, cheng2025ral}, our method (i.e., sensitivity-informed gain scheduling) achieved substantially lower IAE, overshoot, and ATE, indicating more accurate and stable setpoint tracking.
The particularly large IAE of the volume-based scaling strategy~\cite{cheng2025ral} can be attributed to its voxel size update mechanism.
Specifically, volume-based scaling computes the next voxel size from a fixed reference voxel size $d_\mathrm{temp}^\mathrm{fixed}$ that is not recursively updated over time.
As a result, when the discrepancy between this fixed reference and the proper voxel size becomes large, the update can under- or over-adjust the voxel size, causing persistent tracking error to accumulate over the sequence.

Compared with the PD controller with fixed gains, the benefit of gain scheduling is also clear.
As shown in Fig.~\ref{fig:gain_scheduling_ablation}(a), when the spatial scale becomes small, the voxelized point count becomes highly sensitive to voxel size changes, so fixed gains can easily induce oscillatory behavior.
Among the ablated variants, removing the scale indicator led to noticeable overshoot when the spatial scale changed rapidly, as shown in Fig.~\ref{fig:gain_scheduling_ablation}(b), whereas removing the error-related terms weakened corrective action when large deviations from the setpoint occurred, as shown in Fig.~\ref{fig:gain_scheduling_ablation}(c).
By contrast, the proposed method jointly adjusts the gains according to both the spatial scale and the tracking condition, thereby reducing overshoot while improving convergence.

The observed setpoint tracking behaviors are also reflected in odometry accuracy.
Methods with unstable tracking and larger oscillations tended to yield higher ATE, whereas the proposed method reduced such oscillations through more stable control and thereby achieved the lowest ATE.
These results indicate that stable setpoint tracking is beneficial not only for voxel size control but also for accurate odometry~estimation.

\subsection{EFFECT OF HYBRID-METRIC STATE UPDATE IN ENVIRONMENTS WITH LIMITED PLANAR STRUCTURE}\label{subsec: exp_hybrid_metric}
% claim
The fourth experiment evaluates the proposed hybrid-metric state update in scenes with limited reliable planar structure.
This experiment supports our fourth claim that complementing point-to-plane residuals~\cite{rusinkiewicz2001IntConfThreeDDigitalImagingAndModeling} with point-to-point residuals~\cite{besl1992tpami} improves odometry robustness in such scenes.

We consider two scenarios in which planar constraints become insufficient along several directions: the \dataset{Waterways-Short} sequence of the \dataset{GEODE}~\cite{chen2026ijrr} dataset and the \dataset{Handheld-B-02} sequence of the \dataset{NarrowWide} dataset.
To examine the numerical conditioning of the LiDAR update, we compute the condition numbers of the translational and rotational blocks of $({\mathbf{H}_{\widetilde{\mathbf{x}}}^{\ell}})^\intercal \mathbf{H}_{\widetilde{\mathbf{x}}}^{\ell}$, where $\mathbf{H}_{\widetilde{\mathbf{x}}}^{\ell}$ is the stacked Jacobian matrix defined in~(\ref{eq: stacked lidar residual jacobian covariance}).
These block-wise condition numbers are widely used as degeneracy indicators in methods that analyze or handle LiDAR degeneracy~\cite{zhang2016icra,ebadi2021JIntelligentAndRoboticSystem,tagliabue2021lion,lee2024ral,lee2025ral,hu2025arxiv}.
A~larger value indicates that the corresponding translational or rotational constraints are unevenly distributed, leaving some directions weakly constrained and making the LiDAR update more susceptible to pose drift or scan-matching slip along those directions.
Thus, lower condition numbers suggest improved numerical conditioning of the LiDAR update~\cite{cheney1998numerical}.

\begin{figure}[t!]
	\centering
	
	\makebox[\linewidth][c]{%
		\begin{subfigure}[t]{0.4\linewidth}
			\centering
			\includegraphics[width=\linewidth]{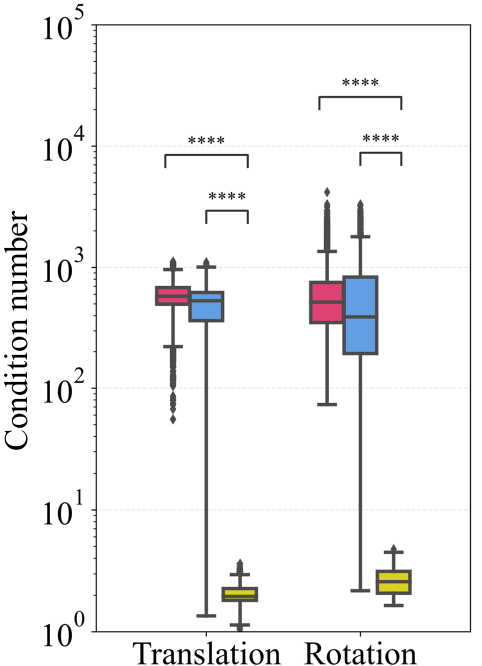}
			\caption{Seq.~\dataset{Waterways-Short}}
		\end{subfigure}
		\hspace{1mm}
		\begin{subfigure}[t]{0.4\linewidth}
			\centering
			\includegraphics[width=\linewidth]{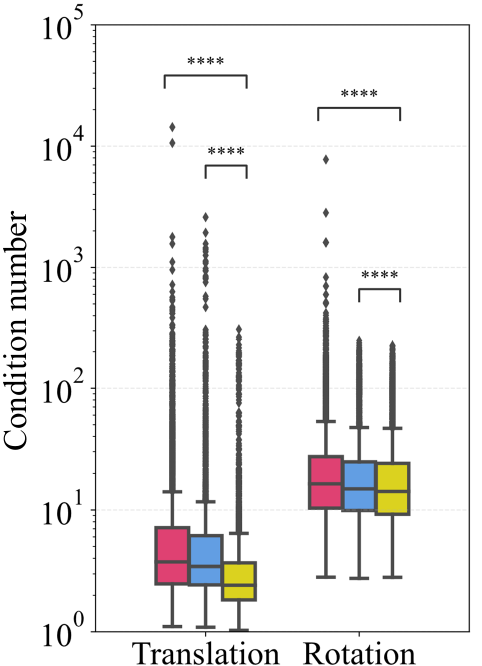}
			\caption{Seq.~\dataset{Handheld-B-02}}
		\end{subfigure}
	}
	
	\vspace{0.1mm}
	
	% Legend
	\begin{subfigure}[t]{\linewidth}
		\centering
		\includegraphics[width=0.63\linewidth]{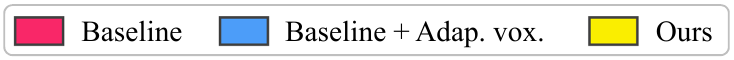}
	\end{subfigure}
	\vspace{-0.15cm}
	\caption{
		Box plots of the condition number on (a) the \dataset{Waterways-Short} sequence of the \dataset{GEODE}~\cite{chen2026ijrr} dataset and (b) the \dataset{Handheld-B-02} sequence of the \dataset{NarrowWide} dataset, evaluated by applying the proposed modules to the baseline system~\cite{hviktortsoi2023pvlio}.
		A lower condition number indicates improved numerical stability of the system~\cite{cheney1998numerical}.
		The $\text{****}$ annotations indicate paired $t$-test results with $p<10^{-4}$.
	}
	\label{fig:plot_condition_number}
	\vspace{-0.6cm}
\end{figure}

\begin{figure}[t!]
	\centering
	\includegraphics[width=1.0\linewidth]{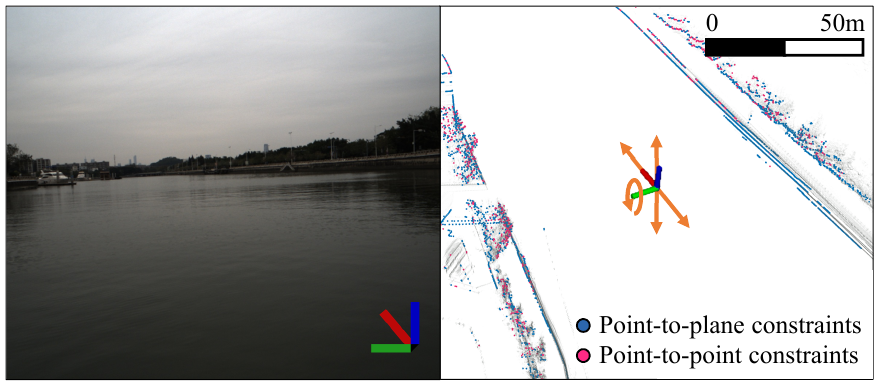}
	\vspace{-0.2cm}
	\caption{Mapping result of GenZ-LIO on the \dataset{Waterways-Short} sequence of the \dataset{GEODE}~\cite{chen2026ijrr} dataset. Translational and rotational directions that are weakly constrained due to insufficient geometric constraints, and are therefore susceptible to LiDAR degeneracy, are indicated by orange arrows. The visualized coordinate frame corresponds to the robot body frame, and the camera image is included solely for improved scene understanding.}
	\vspace{-0.2cm}
	\label{fig:qualitative_waterways}
\end{figure}

As shown in Fig.~\ref{fig:plot_condition_number}, the full system yielded the lowest condition numbers in both scenarios, suggesting improved numerical conditioning of the LiDAR update.
On the \dataset{Waterways-Short} sequence, the baseline exhibited large condition numbers and severe drift, whereas applying only the scale-aware adaptive voxelization reduced these values but still left several directions weakly constrained.
By contrast, the full system further lowered the condition numbers and achieved the lowest ATE of $0.69\,\mathrm{m}$.
This result is consistent with the scene characteristics in Fig.~\ref{fig:qualitative_waterways}, where the waterway environment provides weak planar constraints for the $x$- and $z$-axis translations and the pitch rotation.
In this setting, the proposed hybrid-metric state update provides additional point-to-point constraints for correspondences in non-planar regions, thereby strengthening geometric constraints along weakly constrained directions.

A similar trend was observed on the \dataset{Handheld-B-02} sequence.
Although the condition numbers were lower than those of \dataset{Waterways-Short}, the baseline still exhibited poorly conditioned updates and eventually diverged.
Applying only the scale-aware adaptive voxelization improved stability and lowered the condition numbers, yielding an ATE of $0.67\,\mathrm{m}$ on the \dataset{Handheld-B-02} sequence in Table~\ref{tab:benchmark}, whereas the full system achieved the lowest condition numbers and an
ATE of $0.17\,\mathrm{m}$.
These results indicate that the hybrid-metric state update is most effective when combined with scale-aware adaptive voxelization: the latter maintains a sufficient number of voxelized points, while the former improves robustness under directionally insufficient planar constraints.

\begin{figure}[t!]
	\centering
	\begin{subfigure}[t]{\linewidth}
		\centering
		\includegraphics[width=\linewidth]{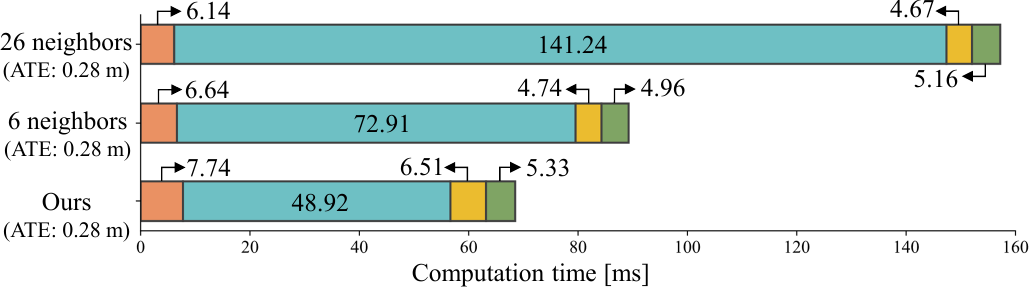}
		\caption{GenZ-LIO}
	\end{subfigure}

	\vspace{-0.2cm}

	\begin{subfigure}[t]{\linewidth}
		\centering
		\includegraphics[width=\linewidth]{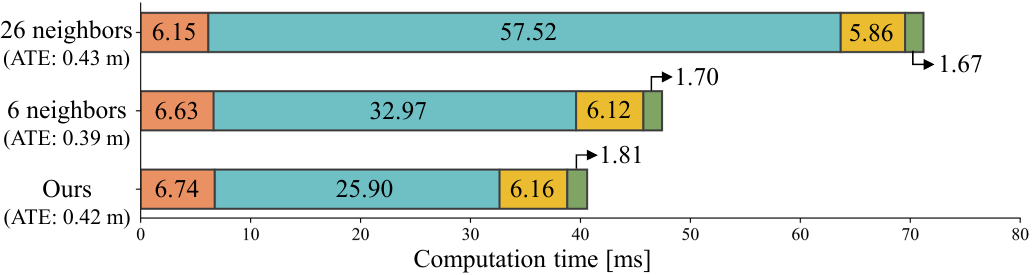}
		\caption{LIO-EKF~\cite{wu2024icra}}
	\end{subfigure}

	\vspace{-0.2cm}

	\begin{subfigure}[t]{\linewidth}
		\centering
		\includegraphics[width=0.9\linewidth]{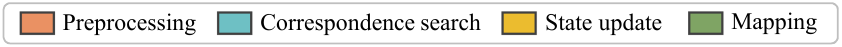}
	\end{subfigure}

	\vspace{0.2cm}

	\caption{Average computation time and ATE (RMSE) on the \dataset{Offroad-04} sequence of the \dataset{GEODE}~\cite{chen2026ijrr} dataset under different correspondence search strategies for the following systems: (a) GenZ-LIO and (b)~LIO-EKF~\cite{wu2024icra}.}
	\label{fig:plot_voxel_pruned_search}
	\vspace{-0.5cm}
\end{figure}

\subsection{ABLATION STUDY ON VOXEL-PRUNED CORRESPONDENCE SEARCH}\label{subsec: exp_ablation_voxel_pruned_search}
% claim
The fifth experiment performs an ablation study on the proposed voxel-pruned correspondence search to quantify its impact on computation time.
This experiment supports our fifth claim that the voxel-pruned correspondence search algorithm substantially reduces computation time by pruning redundant traversal of neighboring voxels.

As discussed in Sec.~\ref{subsec: exp_hybrid_metric}, the hybrid-metric state update improves robustness by incorporating point-to-point constraints when planar information is insufficient.
However, point-to-point correspondence search can introduce substantial computational overhead, since each candidate voxel may contain tens to hundreds of accumulated points and nearest-neighbor distance evaluations must be repeated across multiple voxels.

To examine how effectively the proposed pruning strategy alleviates this cost, we compare three correspondence search strategies on the \dataset{Offroad-04} sequence of the \dataset{GEODE}~\cite{chen2026ijrr} dataset using two systems: GenZ-LIO, which combines point-to-plane and point-to-point error metrics, and LIO-EKF~\cite{wu2024icra}, which relies only on the point-to-point error metric.
The compared strategies are: (i)~searching the root voxel and all 26 neighboring voxels~\cite{vizzo2023ral, wu2024icra}, (ii)~searching the root voxel and only its 6 surface-sharing neighboring voxels~\cite{he2025ral}, and (iii)~the proposed voxel-pruned correspondence search.
In addition to computation time, we also compare the RMSE of the ATE to examine whether reducing the search space affects odometry accuracy.

\begin{table*}[t]
	\captionsetup{}
	\caption{Comparison between GenZ-LIO and related methods.}
	\label{tab:methodological_comparison_paper}
	\centering
	\small
	\renewcommand{\arraystretch}{1.10}
	\setlength{\tabcolsep}{4pt}
	{
		\begin{tabular}{>{\raggedright\arraybackslash}p{0.17\textwidth}| >{\raggedright\arraybackslash}p{0.13\textwidth} >{\raggedright\arraybackslash}p{0.34\textwidth} >{\raggedright\arraybackslash}p{0.28\textwidth}}
			\toprule
			\textbf{Stage} & \textbf{Method} & \textbf{Adaptation criterion} & \textbf{Control strategy} \\
			\midrule
			\multirow[c]{4}{=}[-9pt]{Adaptive voxelization}
			& LOCUS\,2.0~\cite{reinke2022ral}
			& Fixed target number of voxelized points $N_\mathrm{desired}^\mathrm{fixed}$
			& Linear scaling \\
			& AdaLIO~\cite{lim2023ur}
			& Fixed threshold for the number of voxelized points $\tau_N^\mathrm{fixed}$
			& Threshold-based switching between two predefined voxel sizes \\
			& LIVOX-CAM~\cite{cheng2025ral}
			& Fixed target number of voxelized points $N_\mathrm{desired}^\mathrm{fixed}$
			& Volume-based scaling \\
			& GenZ-LIO
			& Scale-informed target number of voxelized points $N_{\mathrm{desired},t}$
			& PD control with gain scheduling \\
			\toprule
			\textbf{Stage} & \textbf{Method} & \textbf{Residual type} & \textbf{Residual weighting} \\
			\midrule
			\multirow[c]{4}{=}[-9pt]{State update}
			& PV-LIO~\cite{hviktortsoi2023pvlio}
			& Point-to-plane~\cite{rusinkiewicz2001IntConfThreeDDigitalImagingAndModeling} residual
			& LiDAR measurement noise model~\cite{yuan2022ral} \\
			& LIO-EKF~\cite{wu2024icra}
			& Point-to-point~\cite{besl1992tpami} residual
			& Fixed isotropic observation covariance \\
			& GenZ-ICP~\cite{lee2024ral}
			& Point-to-plane~\cite{rusinkiewicz2001IntConfThreeDDigitalImagingAndModeling} and point-to-point~\cite{besl1992tpami} residuals
			& Ratio of point-to-plane and point-to-point residuals \\
			& GenZ-LIO
			& Point-to-plane~\cite{rusinkiewicz2001IntConfThreeDDigitalImagingAndModeling} and point-to-point~\cite{besl1992tpami} residuals
			& LiDAR measurement noise model~\cite{yuan2022ral} and discretization-error variance \\
			\bottomrule
	\end{tabular}}
\end{table*}

As shown in Fig.~\ref{fig:plot_voxel_pruned_search}, the proposed voxel-pruned correspondence search achieved the lowest correspondence search time in both GenZ-LIO and LIO-EKF, while maintaining accuracy comparable to the other strategies.
Searching all neighboring voxels incurred the highest computation time, whereas restricting the search to only surface-sharing neighbors reduced the cost but still left unnecessary voxel traversals.
By contrast, the proposed strategy achieved the largest reduction in correspondence search time in both frameworks, which supports the effectiveness of selecting only query-adjacent neighboring voxels and pruning accesses that cannot yield a closer correspondence.

The ATEs in Fig.~\ref{fig:plot_voxel_pruned_search} also show that these reductions in computation time did not meaningfully degrade odometry accuracy: for GenZ-LIO, the ATE remained unchanged across the compared strategies, while for LIO-EKF, only minor variations were observed.
Taken together, these results show that the proposed voxel-pruned correspondence search improves computational efficiency while preserving the quality of correspondence matching and downstream odometry estimation.

\section{FURTHER ANALYSIS AND VALIDATION}
\label{sec:further_analysis_and_validation}
{
\captionsetup[figure]{}
\captionsetup[table]{}
\captionsetup[subfigure]{}
Building on the experiments supporting the key claims of GenZ-LIO, this section provides further analysis and validation of its methodological distinctions, underlying design choices, parameter sensitivity, and applicability across broader operating conditions.
It first compares the proposed components with closely related methods in Sec.~\ref{sec:methodological_comparison} and then examines the design rationale for the median-range scale indicator, scale-dependent point-count setpoint, geometric-mean gain interpolation, and point-to-point covariance formulation in Secs.~\ref{sec:median_scale_indicator}--\ref{sec:ptpo_covariance_design}.
Finally, parameter sensitivity and performance in general operating scenarios are evaluated in Secs.~\ref{sec:parameter_sensitivity_analysis} and~\ref{sec:general_operating_scenarios}, respectively.

\subsection{DIRECT COMPARISON WITH RELATED METHODS}
\label{sec:methodological_comparison}
Table~\ref{tab:methodological_comparison_paper} compares the two core components of GenZ-LIO, namely scale-aware adaptive voxelization and the hybrid-metric state update, with closely related methods at the component level.
At the adaptive voxelization stage, the proposed scale-aware adaptive voxelization differs from LOCUS\,2.0~\cite{reinke2022ral}, AdaLIO~\cite{lim2023ur}, and LIVOX-CAM~\cite{cheng2025ral} in both its adaptation criterion and control strategy.
LOCUS\,2.0 and LIVOX-CAM regulate the voxel size using a fixed target number of voxelized points through linear and volume-based scaling, respectively.
AdaLIO instead switches between two predefined voxel sizes according to a fixed point-count threshold.
In contrast, GenZ-LIO varies the target point count itself according to the spatial scale represented by the current scan and uses PD control with sensitivity-informed gain scheduling to track this scale-informed setpoint.
Thus, the proposed adaptation accounts for spatial-scale variation in both the operating target and the control response.

At the state-update stage, the proposed hybrid-metric state update differs from PV-LIO~\cite{hviktortsoi2023pvlio}, LIO-EKF~\cite{wu2024icra}, and GenZ-ICP~\cite{lee2024ral} in both its residual formulation and weighting strategy.
PV-LIO uses point-to-plane residuals weighted by a LiDAR measurement noise model, whereas LIO-EKF uses point-to-point residuals with a fixed isotropic observation covariance.
GenZ-ICP combines both residual types but determines their residual-type weights from the ratio of point-to-plane and point-to-point residuals.
GenZ-LIO instead formulates both residual types within an ESIKF and assigns correspondence-wise weights based on the LiDAR measurement noise model and the discretization-error variance.
This formulation therefore combines the complementary geometric constraints through their estimated reliability rather than through a residual-type-level heuristic.

\begin{figure*}[t!]
	\captionsetup{}
	\captionsetup[subfigure]{skip=1pt}
	\centering
	\makebox[\textwidth][c]{%
		\begin{subfigure}[t]{0.267\textwidth}
			\centering
			\includegraphics[width=\linewidth]{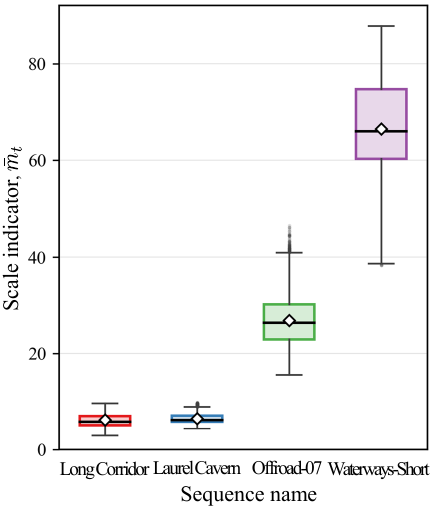}
			\caption{}
			\label{fig: setpoint_ablation_a}
		\end{subfigure}
		\hspace{0.05cm}
		\hfill
		\begin{subfigure}[t]{0.365\textwidth}
			\centering
			\includegraphics[width=\linewidth]{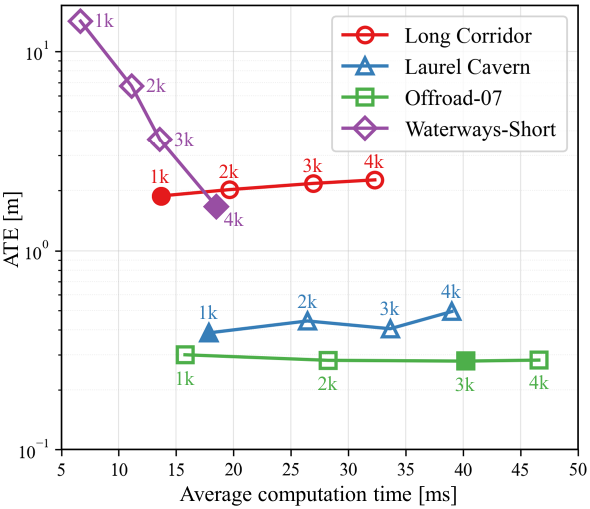}
			\caption{}
			\label{fig: setpoint_ablation_b}
		\end{subfigure}
		\hfill
		\begin{subfigure}[t]{0.353\textwidth}
			\centering
			\includegraphics[width=\linewidth]{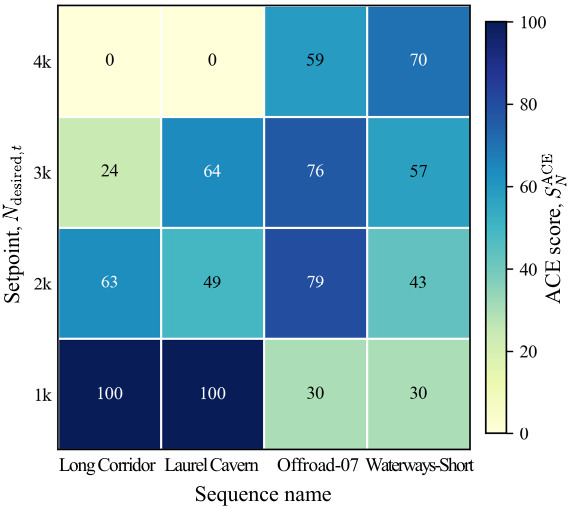}
			\caption{}
			\label{fig: setpoint_ablation_c}
		\end{subfigure}
	}
	\vspace{-0.25cm}
	\caption{Fixed point-count setpoint analysis under different spatial scales. (a) Distribution of the scale indicator $\bar m_t$ for each evaluated sequence, where the white diamond and black line indicate the mean and median, respectively. (b) ATE and average computation time obtained by fixing the target point count to $N_{\mathrm{desired},t}\in\{1,\!000,2,\!000,3,\!000,4,\!000\}$. The filled markers identify the setpoints with the lowest ATE. (c)~Sequence-wise ACE score $S_N^{\mathrm{ACE}}$, which jointly summarizes ATE and computation time; higher values indicate a better accuracy--efficiency trade-off within the same sequence.}
	\label{fig: setpoint_ablation}
	\vspace{-0.7cm}
\end{figure*}

\subsection{DESIGN CONSIDERATIONS FOR THE MEDIAN-RANGE SCALE INDICATOR}
\label{sec:median_scale_indicator}
GenZ-LIO employs uniform voxel grid sampling~\cite{schnabel2006vgtc,rusu2011icra} and regulates the voxel size such that the resulting voxelized point count tracks the setpoint~$N_{\mathrm{desired},t}$ determined from the spatial scale of the current LiDAR scan.
Rather than representing a precise information-optimal point count for each scan, $N_{\mathrm{desired},t}$ serves as a coarse operating target that reduces redundant local observations and unnecessary computation in confined scenes while retaining sufficient geometric support in open scenes.
Because voxelized point clouds retain some geometric redundancy, fine-grained setpoint selection is unnecessary and does not necessarily improve odometry accuracy.
Accordingly, the scale indicator only needs to capture a stable trend in spatial-scale variation rather than fully characterize complex spatial structures.

To capture this trend with minimal computational overhead, common range statistics can serve as candidates for a lightweight scale indicator.
Among common range statistics, the mean is sensitive to a small number of distant returns, while the trimmed mean requires an additional trimming ratio.
The mode of continuous range values requires predefined histogram bin widths and boundaries and may vary with these settings.
In contrast, the median requires no statistic-specific parameter and is robust to outliers~\cite{rousseeuw2003robust,huber2011robust}.
Based on these properties, the median is adopted as a lightweight statistic for representing variations in spatial extent.

\subsection{PRACTICAL RATIONALE FOR SCALE-DEPENDENT POINT-COUNT SETPOINTS}
\label{sec:scale_dependent_setpoint}
This subsection provides an empirical rationale for assigning different point-count setpoints to scenes with different spatial scales.
The scale-informed setpoint in Sec.~\ref{subsec:scale-informed setpoint} is not intended to be an information-optimal point count for every individual scan.
Instead, it provides a practical mechanism for increasing geometric support in open scenes while avoiding unnecessary computation in confined scenes.

We evaluated fixed point-count setpoints on four sequences collected in confined and open environments.
The confined scenes are the \dataset{Long Corridor} and \dataset{Laurel Cavern} sequences from the \dataset{SubT-MRS}~\cite{zhao2024cvpr} dataset.
The open scenes are the \dataset{Offroad-07} and \dataset{Waterways-Short} sequences from the \dataset{GEODE}~\cite{chen2026ijrr} dataset.
For each sequence, we used PV-LIO~\cite{hviktortsoi2023pvlio} as the baseline LIO system and applied the adaptive voxelization module described in Sec.~\ref{sec:scale-aware adaptive voxelization}.
To isolate the effect of the target point count, we replaced the scale-dependent setpoint in~(\ref{eq: scale-informed setpoint}) with a fixed value $N_{\mathrm{desired},t}=N$, where $N\in\mathcal{K}$ and $\mathcal{K}=\{1,\!000,\,2,\!000,\,3,\!000,\,4,\!000\}$.
All other components were kept unchanged.

For each fixed setpoint, we measured the RMSE of the ATE and the average computation time per scan.
We also computed the distribution of the scale indicator $\bar m_t$ for each sequence.
To jointly summarize trajectory accuracy and computational efficiency, we define an accuracy--computational efficiency (ACE) score.
Let $A_N$ and $T_N$ denote the ATE and the average computation time obtained with setpoint $N$, respectively.
For each sequence, we first convert both metrics into normalized costs over the candidate set $\mathcal{K}$.
Since ATE spans a wide numerical range over the evaluated settings, we normalize ATE on a logarithmic scale as
\begin{equation}
\hat{A}_N=
\frac{\log A_N-\min_{M\in\mathcal{K}}\log A_M}
{\max_{M\in\mathcal{K}}\log A_M-\min_{M\in\mathcal{K}}\log A_M},
\end{equation}
and normalize the computation time as
\begin{equation}
\hat{T}_N=
\frac{T_N-\min_{M\in\mathcal{K}}T_M}
{\max_{M\in\mathcal{K}}T_M-\min_{M\in\mathcal{K}}T_M},
\end{equation}
where $\hat{A}_N$ and $\hat{T}_N$ are the normalized accuracy and runtime costs, respectively, and smaller values indicate better performance within the same sequence.
The ACE score, denoted by $S_N^{\mathrm{ACE}}$, is then defined as follows:
\begin{equation}\label{eq: ace_score}
S_N^{\mathrm{ACE}}
=
100
\left[
1-
\left(
\alpha \hat{A}_N
+
(1-\alpha)\hat{T}_N
\right)
\right],
\end{equation}
where the weight $\alpha\in[0,1]$ controls the relative importance of accuracy and runtime.
We set $\alpha=0.7$ to emphasize trajectory accuracy while still penalizing excessive computation.
Because the normalization is performed independently for each sequence, the ACE score should be interpreted as a relative ranking of setpoints within the same sequence, not as an absolute performance measure across different sequences.

The results are summarized in Fig.~\ref{fig: setpoint_ablation}.
First, Fig.~\ref{fig: setpoint_ablation}(a) shows that the scale indicator remains low in the confined sequences and increases in the open sequences.
The mean scale indicators are $6.18\,\mathrm{m}$ and $6.48\,\mathrm{m}$ for \dataset{Long Corridor} and \dataset{Laurel Cavern}, respectively, whereas they increase to $26.73\,\mathrm{m}$ for \dataset{Offroad-07} and $66.47\,\mathrm{m}$ for \dataset{Waterways-Short}.
This separation allows us to examine how the preferred point-count setpoint changes from confined to open scenes.

As shown in Fig.~\ref{fig: setpoint_ablation}(b), increasing the fixed setpoint does not uniformly improve odometry accuracy.
In the two confined sequences, the smallest setpoint of $1,\!000$ points yields the lowest ATE.
Increasing the setpoint to $4,\!000$ points raises the ATE from $1.88\,\mathrm{m}$ to $2.26\,\mathrm{m}$ in \dataset{Long Corridor} and from $0.39\,\mathrm{m}$ to $0.49\,\mathrm{m}$ in \dataset{Laurel Cavern}, while the average computation time more than doubles.
These results suggest that retaining more points in confined scenes mainly increases redundant observations and computation, with little additional benefit to the geometric constraints used for state estimation.

The preferred setpoint shifts as the spatial scale increases.
For the intermediate-scale \dataset{Offroad-07} sequence, the lowest ATE is obtained with $3,\!000$ points, but the ACE score in Fig.~\ref{fig: setpoint_ablation}(c) favors $2,\!000$ points because it provides nearly the same accuracy with lower runtime.
For the large-scale \dataset{Waterways-Short} sequence, larger setpoints are more beneficial: increasing the setpoint from $1,\!000$ to $4,\!000$ points reduces the ATE from $14.19\,\mathrm{m}$ to $1.67\,\mathrm{m}$, and the ACE score increases from $30$ to $70$ despite the runtime increase.
These trends suggest that open scenes can benefit from more voxelized points to retain sufficient geometric constraints, whereas confined scenes can be handled with fewer points.

Overall, the preferred fixed setpoint shifts from $1,\!000$ points in small-scale scenes, to $2,\!000$--$3,\!000$ points in an intermediate-scale scene, and to $4,\!000$ points in the largest-scale scene.
This observation supports the scale-dependent setpoint design in~(\ref{eq: scale-informed setpoint}), where the target point count increases with the spatial scale and the lower and upper bounds prevent unnecessary computation and unbounded point growth.

\subsection{DESIGN RATIONALE FOR GEOMETRIC-MEAN GAIN INTERPOLATION}
\label{sec:geometric_mean_gain_interpolation}
This subsection further explains the use of the geometric mean in the sensitivity-informed gain scheduling described in Sec.~\ref{subsec:sensitivity-informed gain scheduling}.
The normalized spatial-scale term $\phi_t\in[0,1]$ represents the spatial scale and the associated sensitivity of the voxelized point count to voxel-size changes.
The normalized error-related term is denoted generically by $\psi_t\in[0,1]$, where $\psi_t=\psi_{p,t}$ for proportional-gain scheduling and $\psi_t=\psi_{d,t}$ for derivative-gain scheduling.

To jointly incorporate the two normalized terms with minimal computational overhead, we consider three standard averaging operators: the arithmetic, geometric, and harmonic means.
The corresponding interpolation factors are defined, respectively, as
\begin{equation}
\Gamma_t^{\mathrm{A}}=\frac{\phi_t+\psi_t}{2},
\qquad
\Gamma_t^{\mathrm{G}}=\sqrt{\phi_t\psi_t},
\qquad
\Gamma_t^{\mathrm{H}}=\frac{2\phi_t\psi_t}{\phi_t+\psi_t},
\end{equation}
where $\Gamma_t^{\mathrm{H}}$ is defined as zero when $\phi_t=\psi_t=0$.
For nonnegative normalized inputs, these interpolation factors satisfy $\Gamma_t^{\mathrm{H}}\leq\Gamma_t^{\mathrm{G}}\leq\Gamma_t^{\mathrm{A}}$.

The distinction among these averaging strategies becomes particularly clear when the two normalized inputs are imbalanced.
For example, consider $\phi_t=0.1$ and $\psi_t=0.9$, corresponding to a confined scene with high point-count sensitivity and a tracking condition favoring a stronger control response, respectively.
The resulting interpolation factors are $\Gamma_t^{\mathrm{H}}=0.18$, $\Gamma_t^{\mathrm{G}}=0.30$, and $\Gamma_t^{\mathrm{A}}=0.50$.
The arithmetic mean remains relatively large due to the larger input and may produce an aggressive gain despite the small scale-related term indicating that a cautious response is preferable.
Conversely, the harmonic mean remains close to the smaller input and may excessively suppress the stronger response indicated by the larger input.
The geometric mean lies between these two responses, providing a compromise that is more responsive than the harmonic mean while remaining more conservative than the arithmetic mean.

A weighted arithmetic combination could adjust this trade-off, but it would introduce an additional weighting parameter and associated tuning.
In contrast, the geometric mean is symmetric and parameter-free.
We therefore adopt it as a practical compromise between corrective responsiveness and oscillation suppression rather than as a globally optimal aggregation rule for every operating condition.

\subsection{ANALYSIS OF POINT-TO-POINT COVARIANCE DESIGN}
\label{sec:ptpo_covariance_design}
This subsection discusses the rationale behind the two design elements that constitute the combined point-to-point covariance in~(\ref{eq: final point-to-point covariance}), namely the covariance scaling factor $\lambda_\mathrm{po}$ and the discretization-error variance $R_\mathrm{disc}$.
The scaling factor $\lambda_\mathrm{po}$ adjusts the numerical scale gap between the point-to-point and point-to-plane covariances in the hybrid-metric state update, whereas the discretization-error variance $R_\mathrm{disc}$ reflects the additional uncertainty that can arise from the voxel-based map representation and the nearest-neighbor association.

\subsubsection{COVARIANCE SCALING FACTOR}
\label{sec:lambda_po}
$R_\mathrm{pl}^{\ell}$ in~(\ref{eq:R_pl}) and $R_\mathrm{po}^{\mathrm{norm},\ell}$ in~(\ref{eq:R_po}) are the point-to-plane~\cite{rusinkiewicz2001IntConfThreeDDigitalImagingAndModeling} and normalized point-to-point~\cite{besl1992tpami} covariances, respectively.
Because these two scalar covariance terms depend on different covariance sources, a numerical scale gap exists between them.
Specifically, $R_\mathrm{pl}^{\ell}$ depends on the covariances of the plane parameters and the query point, as given in~(\ref{eq:R_pl}), whereas $R_\mathrm{po}^{\mathrm{norm},\ell}$ is obtained by projecting the covariances of the query point and its closest target point onto the unit residual direction, as given in~(\ref{eq:R_po}).
Therefore, even under the same LiDAR noise model~\cite{yuan2022ral}, the difference in the underlying covariance sources leads to the scale gap between $R_\mathrm{pl}^{\ell}$ and $R_\mathrm{po}^{\mathrm{norm},\ell}$ observed in Fig.~\ref{fig:covariance_scale_comparison}.

In the Kalman update, each residual is weighted by the inverse of its covariance.
Accordingly, leaving this scale gap uncompensated allows an excessively large $R_\mathrm{po}^{\mathrm{norm},\ell}$ relative to $R_\mathrm{pl}^{\ell}$ to substantially lower the weight assigned to the point-to-point residuals.
As a result, the point-to-point constraints that could help stabilize the state update in non-planar or weakly structured regions may be largely suppressed.

To examine whether such a scale adjustment is necessary and, if so, which value of $\lambda_\mathrm{po}$ is appropriate, we evaluated six values of $\lambda_\mathrm{po}$ on the \dataset{Stairs} and \dataset{Waterways-Short} sequences of the \dataset{GEODE}~\cite{chen2026ijrr} dataset and the \dataset{Tracked-01} sequence of our \dataset{NarrowWide} dataset.
These three sequences represent confined, open, and confined--open transition environments, respectively.
Here, $\lambda_\mathrm{po}\!=\!1$ corresponds to the case where no down-scaling is applied to the point-to-point residual covariance in~(\ref{eq: final point-to-point covariance}).

As shown in Table~\ref{tab:lambda_po_ablation}, $\lambda_\mathrm{po}\!=\!1$ yields the least favorable average rank among the evaluated values.
This suggests that, without any scale reduction of the point-to-point covariance, the point-to-point constraints can be weighted too weakly, which in turn can degrade odometry accuracy.
In contrast, $\lambda_\mathrm{po}\!=\!0.05$ achieves the best average rank, showing the lowest ATE on all three evaluated sequences.
We therefore introduce the scaling factor and set $\lambda_\mathrm{po}\!=\!0.05$ as the default value in this work.

\begin{figure}[!t]
	\centering
	\includegraphics[width=1.0\linewidth]{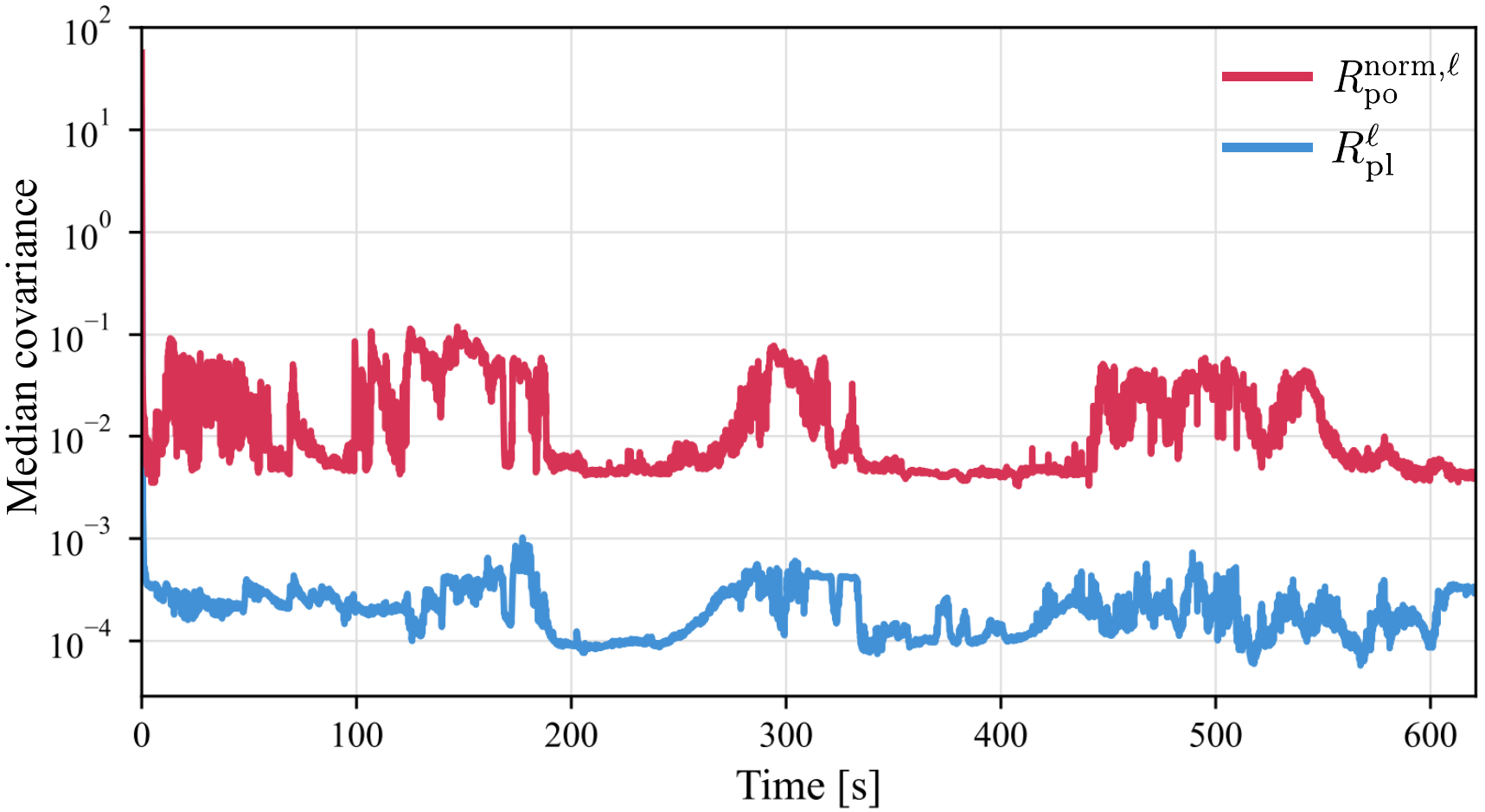}
	\caption{Comparison of the frame-wise medians of the point-to-plane covariance $R_\mathrm{pl}^{\ell}$ and the normalized point-to-point covariance $R_\mathrm{po}^{\mathrm{norm},\ell}$ over time on the \dataset{Tracked-01} sequence of our \dataset{NarrowWide} dataset. The two covariance terms show a persistent numerical scale gap throughout the sequence.}
	\label{fig:covariance_scale_comparison}
	\vspace{-0.35cm}
\end{figure}

\subsubsection{DISCRETIZATION-ERROR VARIANCE}
\label{sec:R_disc}
Here, we discuss the need for the discretization-error variance $R_\mathrm{disc}$ from a mathematical perspective and then validate it experimentally.
As defined in~(\ref{eq:R_po}), the normalized point-to-point covariance $R_\mathrm{po}^{\mathrm{norm},\ell}$ is obtained by projecting $\mathbf{R}_\mathrm{po}^{\ell}$ onto the unit residual direction.
For notational brevity, let $\mathbf{u}_\mathrm{po}^{\ell}=\mathbf{z}_\mathrm{po}^{\ell}/\lVert\mathbf{z}_\mathrm{po}^{\ell}\rVert$.
Then, $R_\mathrm{po}^{\mathrm{norm},\ell}$ can be written as follows:
\begin{equation} \label{eq: normalized ptpo cov using u_po}
R_\mathrm{po}^{\mathrm{norm},\ell}
=
\left({\mathbf{u}_\mathrm{po}^{\ell}}\right)^\intercal
\mathbf{R}_\mathrm{po}^{\ell}
\mathbf{u}_\mathrm{po}^{\ell}.
\end{equation}

Since $\mathbf{R}_\mathrm{po}^{\ell}$ is a symmetric positive semi-definite matrix, it admits the following eigendecomposition:
\begin{equation}
\mathbf{R}_\mathrm{po}^{\ell}
=
\mathbf{E}^{\ell}\mathbf{\Lambda}^{\ell}{\mathbf{E}^{\ell}}^\intercal,
\end{equation}
where $\mathbf{\Lambda}^{\ell}=\mathrm{diag}(\rho_1^\ell,\rho_2^\ell,\rho_3^\ell)$ with $0\leq\rho_1^\ell\leq\rho_2^\ell\leq\rho_3^\ell$, and $\mathbf{E}^{\ell}=[\mathbf{e}_1^\ell,\mathbf{e}_2^\ell,\mathbf{e}_3^\ell]$ is the orthonormal eigenvector matrix.
Substituting this eigendecomposition into the quadratic form in~(\ref{eq: normalized ptpo cov using u_po}) yields:
\begin{align}
R_\mathrm{po}^{\mathrm{norm},\ell}
&=
\left({\mathbf{u}_\mathrm{po}^{\ell}}\right)^\intercal
\mathbf{E}^{\ell}\mathbf{\Lambda}^{\ell}{\mathbf{E}^{\ell}}^\intercal
\mathbf{u}_\mathrm{po}^{\ell} \nonumber\\
&=
\left({\mathbf{E}^{\ell}}^\intercal\mathbf{u}_\mathrm{po}^{\ell}\right)^\intercal
\mathbf{\Lambda}^{\ell}
\left({\mathbf{E}^{\ell}}^\intercal\mathbf{u}_\mathrm{po}^{\ell}\right),
\end{align}
where we define $\boldsymbol{\gamma}^{\ell}={\mathbf{E}^{\ell}}^\intercal\mathbf{u}_\mathrm{po}^{\ell}$, whose $i$-th element $\gamma_i^\ell={\mathbf{e}_i^\ell}^\intercal\mathbf{u}_\mathrm{po}^{\ell}$ denotes the component of the unit residual direction along the $i$-th eigenvector.
Since $\mathbf{E}^{\ell}$ is orthonormal and $\lVert\mathbf{u}_\mathrm{po}^{\ell}\rVert=1$, we have $\lVert\boldsymbol{\gamma}^{\ell}\rVert=1$, which is equivalent to $\sum_{i=1}^{3}(\gamma_i^\ell)^2=1$.
Therefore, $R_\mathrm{po}^{\mathrm{norm},\ell}$ can be interpreted as a weighted average of the eigenvalues of $\mathbf{R}_\mathrm{po}^{\ell}$, where the weights are determined by the residual direction, as follows:
\begin{equation} \label{eq: weighted average of eigenvalues}
R_\mathrm{po}^{\mathrm{norm},\ell}
=
{\boldsymbol{\gamma}^{\ell}}^\intercal
\mathbf{\Lambda}^{\ell}
\boldsymbol{\gamma}^{\ell}
=
\sum_{i=1}^{3}
\rho_i^\ell(\gamma_i^\ell)^2 .
\end{equation}

\begin{table}[!t]

	\caption{Sensitivity analysis of the point-to-point covariance scaling factor $\lambda_\mathrm{po}$ on the \dataset{Stairs} and \dataset{Waterways-Short} sequences of \dataset{GEODE}~\cite{chen2026ijrr} and the \dataset{Tracked-01} sequence of our \dataset{NarrowWide} dataset. ATE is reported as RMSE in meters. The lowest ATE and the lowest average rank are shown in \textbf{bold}.}
	\vspace{-0.1cm}
	\label{tab:lambda_po_ablation}
	\centering
	\footnotesize
	\setlength{\tabcolsep}{6.7pt}
	\renewcommand\arraystretch{0.75}
	\begin{tabular}{c|ccc|c}
		\toprule[0.8pt]
		$\lambda_\mathrm{po}$ & \dataset{Stairs} & \makecell{\dataset{Waterways-Short}} & \dataset{Tracked-01} & \makecell{Average rank} \\
		\midrule
		0.01 & 0.25 & 1.73 & 0.31 & 4.00 \\
		0.05 & \textbf{0.24} & \textbf{0.69} & \textbf{0.16} & \textbf{1.00} \\
		0.1 & 0.36 & 0.91 & 0.17 & 2.67 \\
		0.2 & 0.31 & 0.95 & 0.18 & 3.00 \\
		0.5 & 0.39 & 2.42 & 0.22 & 5.00 \\
		1 & 0.38 & 4.10 & 0.28 & 5.33 \\
		\bottomrule[0.8pt]
	\end{tabular}
	\vspace{-0.3cm}
\end{table}

However, this direction-dependent projection can make the projected covariance excessively small, so that it no longer reflects the additional uncertainty inherent in the nearest-neighbor association.
For instance, if the residual direction is nearly aligned with the eigenvector $\mathbf{e}_1^\ell$ corresponding to the smallest eigenvalue $\rho_1^\ell$, which is the case where the weighted average in~(\ref{eq: weighted average of eigenvalues}) attains its minimum, then $(\gamma_1^\ell)^2\approx1$, $(\gamma_2^\ell)^2\approx0$, and $(\gamma_3^\ell)^2\approx0$, which gives:
\begin{equation}
R_\mathrm{po}^{\mathrm{norm},\ell}
\approx
\rho_1^\ell .
\end{equation}
In this case, if $\rho_1^\ell\approx0$, the projected covariance $R_\mathrm{po}^{\mathrm{norm},\ell}$ can become extremely small.
As a result, an excessively large inverse-covariance weight can be assigned to the corresponding point-to-point residual in the state update, even though that nearest-neighbor association may be affected by map discretization or local ambiguity.

\begin{figure}[!t]
	\centering
	\begin{subfigure}[t]{1.0\columnwidth}
		\centering
		\includegraphics[width=\linewidth]{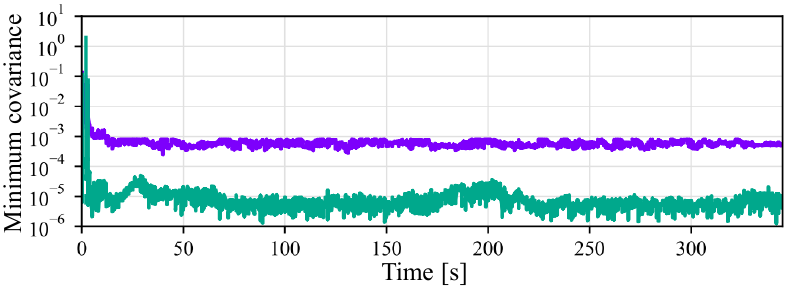}
		\captionsetup{font=scriptsize,skip=0pt}
		\caption{\dataset{Stairs}}
	\end{subfigure}
	
	\vspace{-0.28cm}
	
	\begin{subfigure}[t]{1.0\columnwidth}
		\centering
		\includegraphics[width=\linewidth]{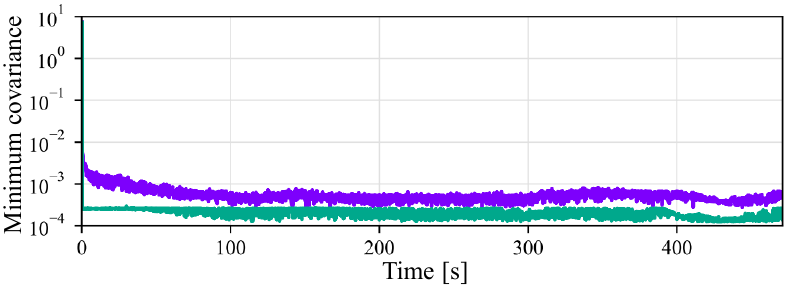}
		\captionsetup{font=scriptsize,skip=0pt}
		\caption{\dataset{Waterways-Short}}
	\end{subfigure}
	
	\vspace{-0.28cm}
	
	\begin{subfigure}[t]{1.0\columnwidth}
		\centering
		\includegraphics[width=\linewidth]{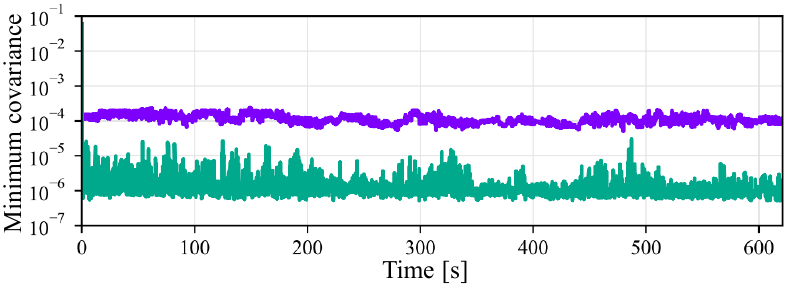}
		\captionsetup{font=scriptsize,skip=0pt}
		\caption{\dataset{Tracked-01}}
	\end{subfigure}
	
	\vspace{-0.1cm}
	\includegraphics[width=0.88\linewidth]{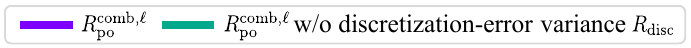}
	\vspace{0.05cm}
	\caption{Temporal evolution of the frame-wise minimum combined point-to-point covariance $R_\mathrm{po}^{\mathrm{comb},\ell}$ with and without the discretization-error variance $R_\mathrm{disc}$.
		The minimum in each frame is computed over all point-to-point correspondences generated in that frame.
		\dataset{Stairs} and \dataset{Waterways-Short} are sequences of the \dataset{GEODE}~\cite{chen2026ijrr} dataset, and \dataset{Tracked-01} is a sequence of our \dataset{NarrowWide} dataset.}
	\label{fig:disc_covariance_ablation}
	\vspace{-0.35cm}
\end{figure}

To mitigate this issue, the proposed formulation adds the discretization-error variance $R_\mathrm{disc}$ in~(\ref{eq: discretization error}) to the projected covariance $R_\mathrm{po}^{\mathrm{norm},\ell}$.
The variance $R_\mathrm{disc}$ increases as the candidate search region becomes broader or the surrounding points become sparser, thereby reflecting the additional uncertainty arising from the nearest-neighbor association and the voxel-map discretization.
Consequently, the combined point-to-point covariance admits the data-dependent lower bound $R_\mathrm{po}^{\mathrm{comb},\ell}\geq\lambda_\mathrm{po}R_\mathrm{disc}$.
That is, $R_\mathrm{disc}$ acts as a covariance floor that prevents an excessively large inverse-covariance weight from being assigned to an individual point-to-point residual.

To evaluate this effect, we compare the full formulation with an ablated variant in which $R_\mathrm{disc}$ is removed, while applying the same default value $\lambda_\mathrm{po}=0.05$ determined above.
The full formulation uses $R_\mathrm{po}^{\mathrm{comb},\ell}=\lambda_\mathrm{po}(R_\mathrm{po}^{\mathrm{norm},\ell}+R_\mathrm{disc})$, whereas the ablated variant uses $R_\mathrm{po}^{\mathrm{comb},\ell}=\lambda_\mathrm{po}R_\mathrm{po}^{\mathrm{norm},\ell}$.
We evaluated the two variants on \dataset{Stairs}, \dataset{Waterways-Short}, and \dataset{Tracked-01}, which represent confined, open, and confined--open transition scenarios, respectively.
Fig.~\ref{fig:disc_covariance_ablation} compares the resulting combined point-to-point covariances, and Table~\ref{tab:r_disc_ablation} summarizes the corresponding odometry accuracy.

As shown in Fig.~\ref{fig:disc_covariance_ablation}, removing $R_\mathrm{disc}$ makes the frame-wise minimum of $R_\mathrm{po}^{\mathrm{comb},\ell}$ smaller overall and increases its temporal fluctuation.
In the ablated formulation, $R_\mathrm{po}^{\mathrm{comb},\ell}=\lambda_\mathrm{po}R_\mathrm{po}^{\mathrm{norm},\ell}$; hence, the combined covariance can also approach zero once the projected covariance of some correspondences becomes very small.
In contrast, in the full formulation, $R_\mathrm{disc}$ acts as a data-dependent covariance floor that mitigates such abrupt drops and fluctuations of the minimum covariance, preventing an excessively large inverse-covariance weight from being assigned to an individual point-to-point residual.

The accuracy results in Table~\ref{tab:r_disc_ablation} follow a trend consistent with the covariance behavior observed in Fig.~\ref{fig:disc_covariance_ablation}.
Removing $R_\mathrm{disc}$ degraded odometry accuracy on all three evaluated sequences, with the largest increase in ATE observed on \dataset{Tracked-01}, from $0.16\,\mathrm{m}$ to $3.35\,\mathrm{m}$.
This corresponds to the observation that the temporal fluctuation of the minimum combined covariance without $R_\mathrm{disc}$ is the largest on \dataset{Tracked-01} among the three sequences.

The sequence-wise differences in the degradation can be further explained by the covariance levels observed in Fig.~\ref{fig:disc_covariance_ablation}.
On \dataset{Stairs} and \dataset{Tracked-01}, which involve confined spaces, many points are measured at close range, so the measurement noise is small and the minimum combined covariance without $R_\mathrm{disc}$ drops to the range of $10^{-6}$--$10^{-5}$.
Accordingly, removing $R_\mathrm{disc}$ causes pronounced accuracy degradation on these two sequences, since such excessively small covariances assign excessively large inverse-covariance weights to a few point-to-point residuals, allowing potentially inaccurate associations to dominate the state update.
In contrast, \dataset{Waterways-Short} is a highly open environment in which measured ranges take relatively large values.
The resulting measurement noise is therefore large, and the minimum combined covariance without $R_\mathrm{disc}$ remains around $10^{-4}$.
Consequently, the accuracy degradation on this sequence is comparatively small, since the inverse-covariance weights of the point-to-point residuals remain moderate even without the covariance floor provided by $R_\mathrm{disc}$.

\begin{table}[!t]

	\caption{Ablation study on the combined point-to-point covariance with and without the discretization-error variance $R_\mathrm{disc}$.
		\dataset{Stairs} and \dataset{Waterways-Short} are sequences of the \dataset{GEODE}~\cite{chen2026ijrr} dataset, and \dataset{Tracked-01} is a sequence of our \dataset{NarrowWide} dataset.
		ATE RMSE is reported in meters and the best result on each sequence is shown in \textbf{bold}.}
	\label{tab:r_disc_ablation}
	\centering
	\footnotesize
	\setlength{\tabcolsep}{5.5pt}
	\renewcommand\arraystretch{0.8}
	\begin{tabular}{lccc}
		\toprule[0.8pt]
		Method & \dataset{Stairs} & \dataset{Waterways-Short} & \dataset{Tracked-01} \\
		\midrule
		Full $R_\mathrm{po}^{\mathrm{comb},\ell}$ (Proposed) & \textbf{0.24} & \textbf{0.69} & \textbf{0.16} \\
		$R_\mathrm{po}^{\mathrm{comb},\ell}$ w/o $R_\mathrm{disc}$ & 0.76 & 1.09 & 3.35 \\
		\bottomrule[0.8pt]
	\end{tabular}
	\vspace{-0.3cm}
\end{table}

Taken together, these two results indicate that the role of $R_\mathrm{disc}$ in mitigating excessively small and unstable combined covariances is important for maintaining stable odometry accuracy.
This ablation study therefore supports the design choice of adding $R_\mathrm{disc}$ as a data-dependent covariance floor.

\subsection{PARAMETER SENSITIVITY ANALYSIS}
\label{sec:parameter_sensitivity_analysis}
This subsection examines the sensitivity of GenZ-LIO to its principal parameters under different spatial-scale conditions.
We perform a one-at-a-time analysis on \dataset{Exp\,16} from \dataset{2022 HILTI}~\cite{zhang2022ral}, \dataset{Offroad-02} from \dataset{GEODE}~\cite{chen2026ijrr}, and \dataset{Handheld-A-01} from our \dataset{NarrowWide} dataset, representing confined, open, and confined--open scenarios, respectively.
When evaluating each parameter group, all remaining parameters are fixed at their common default values without sequence-specific tuning.

The condition $p>1$ in~(\ref{eq:rho}) is imposed to ensure a smooth transition to the saturated setpoint.
For $0\leq\bar m_t<\tau_m$, the derivative of the interpolation function is
\begin{equation}
\rho'(\bar m_t)
=\frac{p}{\tau_m}
\left(1-\frac{\bar m_t}{\tau_m}\right)^{p-1}.
\end{equation}
When $p>1$, the left derivative converges to zero at $\bar m_t=\tau_m$, matching the zero derivative in the saturated region.
For $p=1$, the pre-saturation mapping is linear, but its slope changes discontinuously from $1/\tau_m$ to zero at the boundary.
For $0<p<1$, the slope increases without bound as $\bar m_t$ approaches $\tau_m$.
Although $p=1$ and $p=0.5$ lie outside the intended smooth-saturation design, they remain mathematically valid and are therefore included as stress-test cases.

For the gain-bound analysis, let $\mathbf{K}_g=[K_{p,\min},K_{p,\max},K_{d,\min},K_{d,\max}]$.
The four bounds are jointly scaled as
\begin{equation}
\mathbf{K}_g=\alpha_g[10^{-6},10^{-5},10^{-8},10^{-7}],
\end{equation}
where $\alpha_g\in\{1,3,5,7,10\}$ and the default is $\alpha_g=5$.
This joint scaling preserves the min--max ratios and the relative magnitudes of the proportional and derivative gains.

\begin{table}[t]

	\caption{One-at-a-time parameter sensitivity analysis under different spatial-scale conditions. ATE is reported as RMSE in meters. Except for the parameter group indicated in the first column, all parameters are fixed at their default values. ``$\ast$'' indicates the default value of each parameter group, and the best result is shown in \textbf{bold}.}
	\label{tab:parameter_sensitivity_analysis}
	\centering
	\setlength{\tabcolsep}{2.7pt}
	\renewcommand{\arraystretch}{0.70}
	{\footnotesize
		\begin{tabular*}{1.0\columnwidth}{@{}>{\centering\arraybackslash}m{0.15\columnwidth}@{\hspace{2pt}}l@{\extracolsep{\fill}}ccc@{}}
			\toprule[0.8pt]
			\multicolumn{2}{c}{Parameter / Value} & \dataset{Exp\,16} & \dataset{Offroad-02} & \dataset{Handheld-A-01} \\
			\midrule
			\multirow{4}{*}{$\tau_m$}
			& 20.0 & 0.307 & 0.305 & 0.221 \\
			& 30.0$^{\ast}$ & 0.196 & \textbf{0.298} & \textbf{0.192} \\
			& 40.0 & \textbf{0.151} & 0.304 & 0.205 \\
			& 50.0 & 0.367 & 0.301 & 0.217 \\
			\midrule
			\multirow{4}{*}{$p$}
			& 0.5 & 0.242 & 0.300 & 0.198 \\
			& 1.0 & 0.224 & 0.302 & 0.235 \\
			& 2.0$^{\ast}$ & 0.196 & \textbf{0.298} & \textbf{0.192} \\
			& 3.0 & \textbf{0.135} & 0.303 & 0.210 \\
			\midrule
			\multirow{4}{*}{$N_{\min}$}
			& 500 & \textbf{0.150} & 0.301 & 0.198 \\
			& 1,000$^{\ast}$ & 0.196 & \textbf{0.298} & \textbf{0.192} \\
			& 2,000 & 0.177 & 0.305 & 0.226 \\
			& 3,000 & 0.313 & 0.302 & 0.201 \\
			\midrule
			\multirow{4}{*}{$N_{\max}$}
			& 2,000 & 0.339 & 0.300 & \textbf{0.183} \\
			& 3,000 & \textbf{0.128} & 0.303 & 0.223 \\
			& 4,000$^{\ast}$ & 0.196 & \textbf{0.298} & 0.192 \\
			& 5,000 & 0.139 & 0.304 & 0.208 \\
			\midrule
			\multirow{6}{*}{$\lambda_\mathrm{po}$}
			& 0.01 & 0.285 & \textbf{0.298} & 13.376 \\
			& 0.05$^{\ast}$ & 0.196 & \textbf{0.298} & \textbf{0.192} \\
			& 0.1 & \textbf{0.145} & 0.305 & 0.206 \\
			& 0.2 & 0.173 & 0.306 & 0.224 \\
			& 0.5 & 0.169 & 0.306 & 0.326 \\
			& 1.0 & 0.343 & 0.307 & 0.700 \\
			\midrule
			\multirow{5}{*}{$\alpha_g$}
			& 1 & 0.313 & 0.304 & 0.193 \\
			& 3 & \textbf{0.148} & 0.302 & 0.201 \\
			& 5$^{\ast}$ & 0.196 & \textbf{0.298} & \textbf{0.192} \\
			& 7 & 0.202 & 0.302 & 0.221 \\
			& 10 & 0.280 & 0.302 & 0.208 \\
			\bottomrule[0.8pt]
	\end{tabular*}}
\end{table}

As shown in Table~\ref{tab:parameter_sensitivity_analysis}, the sensitivity varies with the operating environment.
On the open \dataset{Offroad-02} sequence, the ATE varies only from $0.298$ to $0.307\,\mathrm{m}$ across all evaluated settings, indicating low sensitivity over the tested ranges.
The confined \dataset{Exp\,16} and confined--open \dataset{Handheld-A-01} sequences exhibit larger variations; nevertheless, the common default values maintain competitive overall performance even though they are not individually optimal on every sequence.

The stress-test cases with $p\leq1$ complete all three sequences without divergence.
However, relative to the default $p=2$, both $p=0.5$ and $p=1$ yield higher ATEs on \dataset{Exp\,16} and \dataset{Handheld-A-01}, while the performance on \dataset{Offroad-02} remains nearly unchanged.
This result does not imply that $p\leq1$ necessarily causes instability; rather, the lower ATE obtained with the default $p=2$ is consistent with the intended smooth-saturation design and suggests that $p>1$ can be beneficial in these scenarios.

\begin{table}[t]

	\caption{Absolute translational errors on additional structured and unstructured sequences without pronounced spatial-scale variations, reported as RMSE in meters. The best results are shown in \textbf{bold}. Note that ``$\times$'' denotes divergence and receives the worst rank.}
	\label{tab:general_operating_scenario_benchmark}
	\centering
	\setlength{\tabcolsep}{1.7pt}
	\renewcommand{\arraystretch}{0.9}
	{\footnotesize
		\begin{tabular}{@{}>{\raggedright\arraybackslash}m{0.23\columnwidth}*{6}{>{\centering\arraybackslash}m{0.098\columnwidth}}>{\centering\arraybackslash}m{0.08\columnwidth}@{}}
			\toprule[0.8pt]
			\multirow{2}{*}{Method}
			& \multicolumn{3}{c}{\dataset{NTU-VIRAL}~\cite{nguyen2022ijrr}}
			& \multicolumn{3}{c}{\dataset{GrandTour}~\cite{frey2026grandtour}}
			& \multirow{2}{*}{\shortstack{Avg.\\rank}} \\
			\cmidrule(lr){2-4}\cmidrule(lr){5-7}
			& \dataset{eee\_01} & \dataset{sbs\_01} & \dataset{tnp\_01}
			& \dataset{KÄB-2} & \dataset{HÖB-2} & \dataset{ALB-2} & \\
			\midrule
			FAST-LIO2~\cite{xu2022tro}
			& 0.89 & 0.35 & 0.30 & 0.07 & \textbf{0.03} & 0.07 & 6.00 \\
			Faster-LIO~\cite{bai2022ral}
			& 0.86 & 0.36 & 0.24 & \textbf{0.06} & \textbf{0.03} & \textbf{0.06} & 4.83 \\
			AdaLIO~\cite{lim2023ur}
			& 0.91 & 0.29 & 0.20 & 0.07 & \textbf{0.03} & 0.07 & 4.67 \\
			Point-LIO~\cite{he2023ais}
			& 0.25 & 0.26 & 0.24 & 0.08 & 0.06 & 0.09 & 6.67 \\
			LIO-EKF~\cite{wu2024icra}
			& 0.56 & 0.28 & 0.25 & 0.07 & 0.68 & 0.07 & 6.17 \\
			DLIO~\cite{chen2023icra}
			& \textbf{0.23} & \textbf{0.25} & 1.78 & 0.38 & $\times$ & 0.87 & 7.67 \\
			iG-LIO~\cite{chen2024ral}
			& 0.58 & 0.30 & 0.21 & 0.07 & \textbf{0.03} & \textbf{0.06} & 4.17 \\
				\begin{tabular}[c]{@{}l@{}}PV-LIO~\cite{hviktortsoi2023pvlio}\\[-2pt](Baseline)\end{tabular}
			& \textbf{0.23} & 0.26 & 0.27 & 0.09 & \textbf{0.03} & 0.12 & 5.50 \\
			\begin{tabular}[c]{@{}l@{}}Baseline w/\\[-2pt]adap. vox.\end{tabular}
			& \textbf{0.23} & 0.26 & \textbf{0.18} & 0.08 & \textbf{0.03} & 0.07 & 2.67 \\
			\begin{tabular}[c]{@{}l@{}}Baseline w/\\[-2pt]hybrid-metric\end{tabular}
			& \textbf{0.23} & 0.26 & 0.21 & 0.07 & \textbf{0.03} & 0.08 & 3.00 \\
			Ours
			& \textbf{0.23} & 0.26 & 0.19 & 0.07 & \textbf{0.03} & 0.07 & \textbf{1.83} \\
			\bottomrule[0.8pt]
	\end{tabular}}
\end{table}

The most pronounced sensitivity is observed for $\lambda_\mathrm{po}$.
In particular, $\lambda_\mathrm{po}=0.01$ excessively reduces the point-to-point covariance and consequently assigns excessively large inverse-covariance weights to the point-to-point residuals, increasing the ATE on \dataset{Handheld-A-01} to $13.376\,\mathrm{m}$.
Conversely, a large value such as $\lambda_\mathrm{po}=1$ overly suppresses the point-to-point contribution and degrades accuracy on this sequence.
The default $\lambda_\mathrm{po}=0.05$ yields the lowest ATE on \dataset{Handheld-A-01} while maintaining competitive performance on the other sequences; a more detailed analysis of this covariance design is provided in Sec.~\ref{sec:ptpo_covariance_design}.

All evaluated configurations complete the three sequences without divergence.
Overall, these results support practical robustness within reasonable parameter ranges using a common default configuration, but do not imply parameter invariance or universal optimality.

\subsection{EVALUATION IN GENERAL OPERATING SCENARIOS}
\label{sec:general_operating_scenarios}
The primary benchmark in Sec.~\ref{subsec: exp_benchmark} focuses on sequences with challenging spatial-scale conditions, including extremely confined spaces, wide open areas, and confined--open transitions.
To evaluate GenZ-LIO against SOTA LIO methods in general operating scenarios, we additionally conduct experiments on sequences without pronounced spatial-scale variations.
We select the \dataset{eee\_01}, \dataset{sbs\_01}, and \dataset{tnp\_01} sequences from the \dataset{NTU-VIRAL}~\cite{nguyen2022ijrr} dataset and the \dataset{Käferberg-Forest\,2} (\dataset{KÄB-2}), \dataset{Hoenggerberg-Forest\,2} (\dataset{HÖB-2}), and \dataset{Albisgütli-Forest\,2}~(\dataset{ALB-2}) sequences from the \dataset{GrandTour}~\cite{frey2026grandtour} dataset.
The former and latter represent structured and unstructured environments, respectively.
For the sequences of the \dataset{GrandTour} dataset, which provide data from multiple LiDAR sensors, we use measurements from the Livox MID-360 LiDAR and its built-in IMU.
All compared methods use the default parameter setups provided by their open-source implementations, without sequence-specific tuning.

As shown in Table~\ref{tab:general_operating_scenario_benchmark}, GenZ-LIO remained stable on all additional structured and unstructured sequences and attained the lowest average rank among all compared methods.
Relative to the baseline method (i.e., PV-LIO~\cite{hviktortsoi2023pvlio}), GenZ-LIO achieved equal or improved accuracy on every sequence, indicating that the proposed components do not degrade baseline performance in general scenarios without pronounced spatial-scale challenges.
These results demonstrate that, although GenZ-LIO is designed to improve robustness under extreme spatial-scale conditions, it maintains competitive performance relative to SOTA LIO systems in general operating scenarios.
}

\section{LIMITATIONS}
\label{sec:limitations}
The proposed scale indicator $\bar m_t$ is a lightweight proxy for the spatial extent of a LiDAR scan and does not explicitly characterize the geometric structure of the surroundings.
Consequently, different environments with similar range distributions may yield similar target point counts, which are not guaranteed to provide the optimal accuracy--efficiency trade-off for every geometric configuration.

GenZ-LIO also assumes that the LiDAR observations contain geometric features usable for scan-to-map registration.
Although adaptive voxelization and the hybrid-metric state update effectively exploit the available measurements, they cannot recover geometric information absent in severely degenerate or featureless environments.
Addressing extremely severe degradation may require the integration of complementary cues, such as LiDAR intensity~\cite{pfreundschuh2024icra}, high-resolution map representations~\cite{pfreundschuh2026arxiv}, or visual measurements~\cite{zheng2025tro}.

\section{CONCLUSION}
In this paper, we presented \textit{GenZ-LIO}, a LIO framework designed to operate robustly while maintaining computational efficiency across environments with substantially different spatial scales.
The proposed method introduces scale-aware adaptive voxelization, which estimates the spatial scale and adjusts the voxel size via a PD controller with sensitivity-informed gain scheduling to drive the voxelized point count toward the scale-informed setpoint.
In addition, the hybrid-metric state update is formulated within an ESIKF to exploit complementary geometric constraints under both structured and unstructured conditions.
The voxel-pruned correspondence search further reduces redundant computations introduced by point-to-point matching.

Extensive evaluations on public benchmarks and the proposed \dataset{NarrowWide} dataset analyze LIO performance under spatial-scale variations across diverse field scenarios.
Across the evaluated sequences, GenZ-LIO maintained stable odometry estimation without divergence under the tested field conditions.
Moreover, ablation studies validate the effect of each proposed module on robustness and computational efficiency across varying spatial scales.
The proposed design provides a practical basis for developing more reliable LiDAR-based odometry systems for field deployments involving frequent changes in spatial scale.
Future work will extend the proposed framework to more diverse sensor modalities, with the aim of improving generalizability across a broader range of environments and operating conditions.

\bibliographystyle{URL-IEEEtrans}
\bibliography{main}
\end{document}